%% file: main.tex
\documentclass[11pt]{article}

\usepackage[margin=0.82in,headheight=15pt]{geometry}
\usepackage[T1]{fontenc}
\usepackage[utf8]{inputenc}
\usepackage{lmodern}
\usepackage{microtype}
\usepackage{amsmath,amssymb}
\usepackage{graphicx}
\usepackage{booktabs,longtable,tabularx,array,multirow}
\usepackage{enumitem}
\usepackage{xcolor}
\usepackage{caption}
\usepackage{subcaption}
\usepackage{float}
\usepackage{placeins}
\usepackage{fancyhdr}
\usepackage{titlesec}
\usepackage{hyperref}
\usepackage{xurl}

\definecolor{GnaniNavy}{HTML}{173B6C}
\definecolor{GnaniTeal}{HTML}{109B8B}
\definecolor{GnaniOrange}{HTML}{E77745}
\definecolor{GnaniLight}{HTML}{EEF4F7}
\definecolor{GnaniGrey}{HTML}{5E6875}
\definecolor{TodoBg}{HTML}{FFF5E8}

\hypersetup{
  colorlinks=true,
  linkcolor=GnaniNavy,
  citecolor=GnaniTeal,
  urlcolor=GnaniTeal,
  pdftitle={Gnani-Evon-v3.3-30B-A3B Technical Report},
  pdfauthor={Sutanshu Saksena Raj, Prasanjith Pasunuti, Parijat Chatterjee, Daksh Maan, Purusharth Malik, Nayanjit Sarkar, Avinash Benki}
}

\setlist[itemize]{leftmargin=1.4em,itemsep=0.2em,topsep=0.25em}
\setlist[enumerate]{leftmargin=1.7em,itemsep=0.2em,topsep=0.25em}
\renewcommand{\arraystretch}{1.16}
\titleformat{\section}{\Large\bfseries\color{GnaniNavy}}{\thesection}{0.65em}{}
\titleformat{\subsection}{\large\bfseries\color{GnaniNavy}}{\thesubsection}{0.65em}{}
\titleformat{\subsubsection}{\normalsize\bfseries\color{GnaniTeal}}{\thesubsubsection}{0.65em}{}

\newcolumntype{Y}{>{\raggedright\arraybackslash}X}
\newcolumntype{L}[1]{>{\raggedright\arraybackslash}p{#1}}
\newcolumntype{R}[1]{>{\raggedleft\arraybackslash}p{#1}}

\newcommand{\model}{Gnani-Evon-v3.3-30B-A3B}
\newcommand{\shortmodel}{Evon v3.3}
\newcommand{\evonbase}{Evon-Base}
\newcommand{\evonit}{Evon-Instruct}
\newcommand{\evonrl}{Evon-RL-Gen}
\newcommand{\evonrlindic}{Evon-RL-Indic}
\newcommand{\sftpone}{SFT-phase-1}
\newcommand{\sftptwo}{SFT-phase-2}
\newcommand{\sftrthree}{SFT-rebuild-3}

\newcommand{\keypoint}[1]{%
  \par\noindent\colorbox{GnaniLight}{%
    \parbox{0.965\linewidth}{#1}}\par}

\begin{document}
\hypersetup{pageanchor=false}

\begin{titlepage}
  \pagecolor{GnaniLight}
  \color{GnaniNavy}
  \vspace*{1.0cm}
  {\color{GnaniTeal}\rule{\textwidth}{2.2pt}}
  \vfill
  {\centering\includegraphics[width=0.36\textwidth]{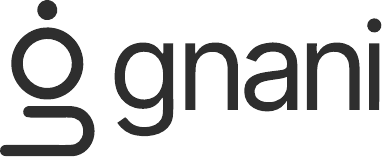}\par}
  \vspace{1.2cm}
  {\fontsize{24}{29}\selectfont\bfseries A Sovereign, Indic-First, Reasoning Model\par}
  \vspace{0.55cm}
  {\large Tokenizer expansion, continual pretraining, supervised fine-tuning, and reinforcement learning. \par}
  \vfill
  {\normalsize
  \textbf{Core Team:} Sutanshu Saksena Raj, Prasanjith Pasunuti, Parijat Chatterjee,
  Daksh Maan, Purusharth Malik, Nayanjit Sarkar \par
  \vspace{0.18cm}
  \textbf{Technical Lead:} Avinash Benki \par
  \vspace{0.18cm}
  {\small Gnani.AI} \par}
  \vspace{0.5cm}
  \begin{tabular}{@{}ll@{}}
    \textbf{Release family} & Released RL checkpoint \\
    \textbf{License} & Apache License 2.0 \\
    \textbf{Report status} & Technical Report \\
  \end{tabular}
  \vfill
  {\small Model artifacts: \url{https://huggingface.co/gnani/gnani-evon-v3.3-30B-A3B}\par}
  \vspace{0.4cm}
  {\color{GnaniTeal}\rule{\textwidth}{2.2pt}}
\end{titlepage}
\nopagecolor
\hypersetup{pageanchor=true}
\pagenumbering{roman}

\begin{abstract}
\model{} is a long-context hybrid Mamba--Transformer mixture-of-experts model for reasoning,
instruction following, and tool use in English and eight Indian languages. Starting from NVIDIA's
Nemotron 3 Nano 30B-A3B checkpoint, it retains the 52-layer backbone and changes the vocabulary,
training data, and post-training objectives. An in-place expansion from 131,072 to 163,072 tokens
reduces mean Indian-language fertility by 38.6\%. The new embedding rows are initialized from their
upstream decompositions and warmed before the backbone is updated. Continual pretraining then adds
broad English--Indic data, high-quality annealing, and direct training at a 131,072-token sequence
length.

The central result appears after post-training. Reinforcement learning across four verifiable task
environments improves mathematics, graduate-level science, and instruction following, but degrades
Indic extractive question answering because the rewards do not require an answer in the requested
language. The share of Indic extraction items answered in English rises from 8.7\% to 91.7\%.
Accuracy conditional on answering in the target script roughly doubles, which is consistent with
preserved reading ability but is measured on a self-selected subset.

Two unreleased studies test repairs. A scoped Indic reward restores the language convention. A
26B-A4B generative judge outperforms a 235B-parameter model on RM-Bench, and distillation restores
81\% of the capabilities displaced by optimizing against it.

Relative to the architecture-identical starting checkpoint, the released model gains 5.2 points on
an Indic aggregate while giving up 1.7 points of English retention. The unreleased Indic-aware branch
gains 14.6 for 5.4. A planned FP8 release reduces the artifact to 32GB with near-zero mean change
across 32 paired benchmarks.
\end{abstract}

\vspace{0.2cm}

\clearpage
\tableofcontents
\clearpage
\pagenumbering{arabic}

\section{Introduction}

Large language models do not process words directly. They first split text into tokens, and the quality of that split depends on which languages were well represented when the tokenizer was built. For many Indian scripts, a tokenizer optimized primarily for English and code can turn one word into several fragments. Training, context capacity, latency, and serving cost are all paid per token, so this fragmentation creates a practical language tax.

\shortmodel{} was built around a simple question: how much Indic capability can be added to an efficient, already capable base model without discarding its English reasoning, code, or deployment advantages? The answer required more than adding translated data~\cite{lowresourceadapt}. Four linked interventions were needed, following the pattern of other sovereign-language adaptation
programmes~\cite{soofi}:

\begin{enumerate}
  \item expand the vocabulary so that Indian scripts are represented efficiently;
  \item adapt the new embedding and output rows without disturbing the inherited model;
  \item continue pretraining on a quality-controlled English--Indic curriculum and teach the model directly at 128K context;
  \item post-train for instruction following and reasoning, measure native-script retention, and study corrective language rewards in a separate experimental branch.
\end{enumerate}

Table~\ref{tab:lineage} gives the development path. In brief, tokenizer adaptation and
continual pretraining produce \evonbase{}; supervised fine-tuning produces \evonit{}; and the
released RL stage produces \evonrl{}. The sibling Indic-aware branch is experimental, not an
ancestor of the release.

\begin{table}[htbp]
\centering
\caption{Production ancestry of the released RL checkpoint and its planned FP8 derivative.}
\label{tab:lineage}
\begin{tabularx}{\textwidth}{@{}L{1.15cm}L{3.9cm}Y@{}}
\toprule
\textbf{Stage} & \textbf{Artifact} & \textbf{What changed} \\
\midrule
0 & NVIDIA base checkpoint & Nemotron 3 Nano 30B-A3B Base BF16, upstream revision \texttt{97ab8012}, with a 131,072-token vocabulary. \\
1 & Tokenizer and initialization & A 32,000-entry continued-BPE extension, portable tokenizer backend, and constituent-mean initialization produce the 163,072-row vocabulary matrices. \\
2 & Embedding warmup & New vocabulary rows are adapted on a 90\% Indic / 10\% English anchor mixture while inherited parameters remain fixed. \\
3 & CPT phase 1 & Broad 8,192-token English--Indic continual pretraining. \\
4 & CPT annealing & High-quality 8,192-token continuation from the selected phase-1 checkpoint. \\
5 & CPT long context & Direct 131,072-token NoPE training through iteration 372. \\
6 & \sftpone{} & First retained supervised curriculum stage. \\
7 & \sftptwo{} & Second retained supervised stage. \\
8 & \sftrthree{} epoch 1 & First pass over the rebuilt \sftrthree{} corpus. Iteration 2,000 is the retained branch point. \\
9 & \sftrthree{} epoch 2 / \evonit{} & Epoch-2 continuation from \sftrthree{} epoch-1 iteration 2,000; its iteration-2,400 export initializes both RL studies. \\
10 & \evonrl{} & Released four-environment GRPO checkpoint. \\
11 & Planned Evon-FP8 & Selectively quantized derivative of \evonrl{}, planned for release. \\
\bottomrule
\end{tabularx}
\end{table}

\subsection{Contributions}

This report makes six main contributions. The lineage they describe runs \evonbase{} $\rightarrow$ \evonit{} $\rightarrow$ \evonrl{}, with \evonrlindic{} branching from \evonit{} as a research arm.

\begin{itemize}
  \item \textbf{An Indic-aware tokenizer upgrade.} The vocabulary is expanded in place rather than replaced, improving eight-language fertility by 38.6\% while preserving inherited token identities.
  \item \textbf{A guarded embedding-adaptation recipe.} Decoupled weight decay can move nominally frozen rows, while an Indic-only warmup can disturb English output calibration even with a frozen backbone.
  \item \textbf{A measured continual-pretraining curriculum.} Per-language uplift informs the data mixture, and direct training establishes the model's 128K context length.
  \item \textbf{A reward-mixture diagnosis.} The released RL run improves reasoning but weakens unrewarded native-script conventions. A separate Indic-aware experiment tests a scoped repair.
  \item \textbf{A reward model beyond verifiable tasks.} An Indic-capable generative judge beats a
  nine-times-larger model on RM-Bench. Distillation then restores 81\% of the capabilities displaced
  by optimizing against that judge while retaining most of its gains.
  \item \textbf{Negative results with operational value.} RL can erase output conventions without
  erasing capability; perplexity can select the wrong retention rate; and tool metrics can hide
  call policies.
\end{itemize}

\section{Starting Point: An Efficient Reference Architecture}

\subsection{Why reuse Nemotron 3 Nano?}

The upstream model is NVIDIA-Nemotron-3-Nano-30B-A3B-Base-BF16~\cite{nemotron3}. It combines Mamba-2 sequence mixers~\cite{mamba2}, sparse attention, and granular mixture-of-experts layers. This matters for two reasons. First, only a small fraction of the total parameters is active for each token. Second, only six attention layers maintain a growing key--value cache, while the Mamba layers carry sequence state compactly.

\keypoint{Sparse experts provide capacity, while routing activates only a small subset for each token.}

The released model contains 31.75B parameters, of which 3.31B are active per token under NVIDIA's
convention. Expanding both untied vocabulary matrices adds 172.03M parameters without changing the
backbone. Table~\ref{tab:architecture} summarizes the design, and Appendix~\ref{app:architectureaccounting}
provides the exact accounting.

\begin{table}[htbp]
\centering
\caption{Architecture inherited from the Nemotron 3 Nano reference configuration.}
\label{tab:architecture}
\begin{tabularx}{0.88\textwidth}{@{}Y L{5.1cm}@{}}
\toprule
\textbf{Hyperparameter} & \textbf{Value} \\
\midrule
Total parameters & 31,749,975,232 (31.75B) \\
Active parameters & 3,313,773,760 (3.31B; NVIDIA convention) \\
Layer count & 52 total: 23 Mamba-2, 23 MoE, 6 GQA \\
Model dimension & 2,688 \\
Attention heads / KV heads & 32 / 2 \\
Attention head dimension & 128 \\
Mamba-2 state dimension / groups & 128 / 8 \\
Mamba-2 heads / head dimension & 64 / 64 \\
Routed / activated / shared expert modules & 128 / 6 / 1 \\
Routed / shared expert hidden dimension & 1,856 / 3,712 ($2\times$ width) \\
MoE activation & squared ReLU (\texttt{relu2}) \\
Router & learned MLP with sigmoid gating \\
Position encoding & none (NoPE) \\
Normalization & RMSNorm \\
Input embedding / output projection & untied \\
Evon vocabulary size & 163,072 \\
Maximum supported sequence length & 131,072 tokens \\
\bottomrule
\end{tabularx}
\end{table}

\subsection{NoPE and the meaning of 128K context}

\shortmodel{} uses no explicit positional embedding, so its 128K support comes from direct training
on 131,072-token sequences rather than positional scaling. That establishes a trained sequence
length, not uniform capability across it. At the CPT checkpoint, HELMET in-context learning rises
from 59.2 at 8K to 81.7 at 128K, while exact retrieval falls from 95.9 to 14.0; retrieval-augmented
generation and long-form question answering remain between those extremes.

Under the same completion prompting, the released checkpoint raises average recall from 57.1 to
84.0. Table~\ref{tab:longctxrl} gives the remaining post-training results. Long-context ability
therefore depends strongly on the task and prompt format, even within the supported length.

\begin{table}[htbp]
\centering\small
\caption{Long-context results under matched completion prompting. \evonrlindic{} is an unreleased
research branch.}
\label{tab:longctxrl}
\begin{tabular}{@{}lrrr@{}}
\toprule
\textbf{Measure} & \textbf{CPT checkpoint} & \textbf{\evonrl{}} & \textbf{\evonrlindic{}} \\
\midrule
Recall average & 57.1 & 84.0 & 81.2 \\
Abstention on unanswerable questions & 98.1 & 99.8 & 99.7 \\
IndicQA F1, HELMET reader slice & 54.3 & 60.5 & 60.1 \\
\bottomrule
\end{tabular}
\end{table}

\section{Tokenizer Expansion}
\label{sec:tokenizer}

\subsection{The language tax}

We measure tokenizer fertility as the number of tokens produced per whitespace-delimited word:

\begin{equation}
  F(\mathcal{D}) =
  \frac{\sum_{x \in \mathcal{D}} N_{\mathrm{tok}}(x)}
       {\sum_{x \in \mathcal{D}} N_{\mathrm{word}}(x)}.
\end{equation}

\keypoint{Lower fertility is better: the same sentence occupies fewer model tokens. This increases usable context and reduces the amount of token-level work needed to process the sentence.}

The upstream vocabulary contains 131,072 entries. We continue its BPE merge process on data from Hindi, Bengali, Gujarati, Kannada, Malayalam, Marathi, Tamil, and Telugu, including romanized text during construction. The selected tokenizer adds 32,000 entries, producing a vocabulary of 163,072.

The released vocabulary contains 162,071 ordinary BPE entries and 1,001 added tokens. Of the added tokens, 19 have named functions and 982 are reserved placeholders. English and existing special tokens retain their upstream identities.

\subsection{Continued BPE and normalization}

Training a completely new tokenizer would break the mapping learned by the base model. Instead, continued BPE preserves inherited pieces and appends merges for sequences that remain fragmented. This follows the in-place expansion pattern studied by Smith et al.~\cite{inplace} and uses adaptive mixture ideas related to AdaptMix~\cite{adaptmix}.

The production Round-09 tokenizer is built on the E3 corpus: Unicode NFKC normalization followed by newline/space cleanup and an Indic-aware regex canonicalization. NFKC folds compatibility variants into a canonical form before BPE learning, while the regex stage aligns better with intended boundaries.

E1 (NFKC only) and E3 (NFKC plus regex) have identical average fertility to four decimal places, 2.1823, in the controlled normalization sweep. Fertility could not arbitrate between them. E3 is the shipped normalization. It changes segmentation and token identities at nearly the same sequence length.

The production lineage uses the Round-09 32K extension and a 163,072-entry vocabulary. A later 64K
extension achieved lower fertility but remained exploratory and was not adopted
(Appendix~\ref{app:mixv5}).

\subsection{Fertility results}

\begin{figure}[htbp]
  \centering
  \includegraphics[width=0.94\textwidth]{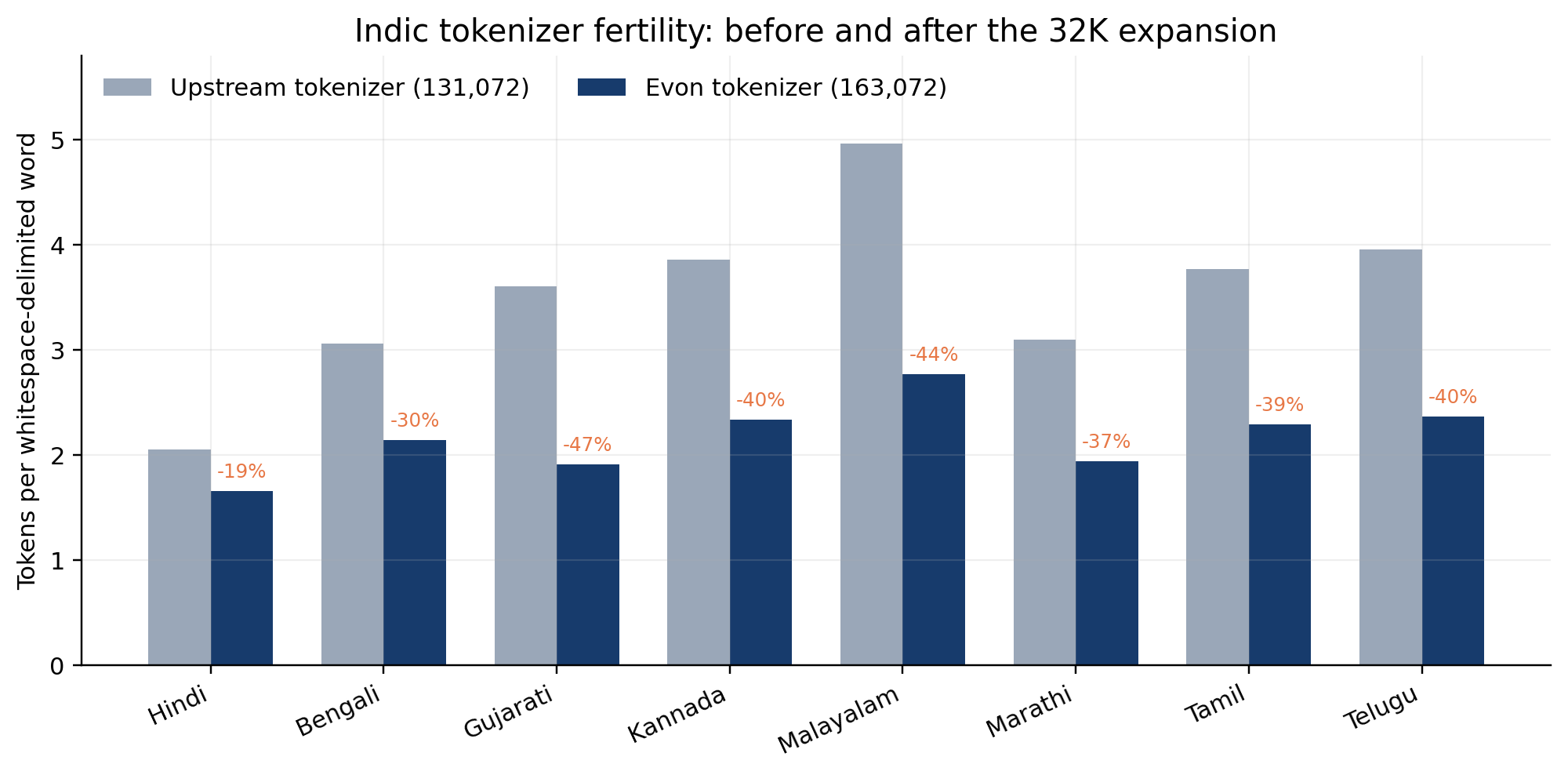}
  \caption{Tokenizer fertility on the held-out Samanantar evaluation slice. Lower is better. Percent labels show change relative to the upstream tokenizer.}
  \label{fig:fertility}
\end{figure}

Figure~\ref{fig:fertility} reports the production tokenizer on the common comparison protocol: Samanantar, 2,000 samples per language, seed 42. The eight-language mean falls from 3.5472 to 2.1773 tokens per word, a 38.6\% reduction. Gujarati improves most (47.1\%), followed by Malayalam (44.3\%).

\begin{table}[htbp]
\centering
\caption{Per-language tokenizer fertility for the 163,072-entry tokenizer carried by the documented model lineage.}
\label{tab:fertility}
\begin{tabular}{@{}lrrr@{}}
\toprule
\textbf{Language} & \textbf{Upstream} & \textbf{Evon v3.3} & \textbf{Change} \\
\midrule
Hindi & 2.0513 & 1.6568 & $-19.2\%$ \\
Bengali & 3.0598 & 2.1416 & $-30.0\%$ \\
Gujarati & 3.6086 & 1.9099 & $-47.1\%$ \\
Kannada & 3.8643 & 2.3377 & $-39.5\%$ \\
Malayalam & 4.9654 & 2.7684 & $-44.3\%$ \\
Marathi & 3.0981 & 1.9387 & $-37.4\%$ \\
Tamil & 3.7705 & 2.2948 & $-39.1\%$ \\
Telugu & 3.9595 & 2.3706 & $-40.1\%$ \\
\midrule
Eight-language mean & 3.5472 & 2.1773 & $-38.6\%$ \\
\bottomrule
\end{tabular}
\end{table}

\subsection{An exploratory 64K vocabulary study}

A larger 64,000-entry extension was trained and measured before the production vocabulary was
fixed. It reaches lower fertility than the shipped tokenizer, at a cost of a second embedding
matrix and a regression on one language, and it was not adopted.
Appendix~\ref{app:mixv5} gives the measurements and the reasoning.

\subsection{Native versus romanized text}

Romanized Indic remains substantially more expensive than native script. Across the new eight-language run, the macro romanized-to-native fertility ratio is 1.84. Seven languages fall in a range of approximately 1.59--1.94 times native fertility. Tamil is the clear outlier at 2.58 times: 2.314 tokens per word in native script versus 5.964 when romanized.

This does not invalidate the production tokenizer's native-script improvement. It identifies its weak flank. Applications expecting heavy romanized Tamil or code-mixed Latin-script traffic should account for the larger token budget and may benefit from a future tokenizer revision.

\begin{table}[htbp]
\centering
\caption{Native-script versus controlled ITRANS romanized fertility for the production tokenizer.}
\label{tab:romanizedfertility}
\begin{tabular}{@{}lrrr@{}}
\toprule
\textbf{Language} & \textbf{Native T/W} & \textbf{Romanized T/W} & \textbf{Ratio} \\
\midrule
Bengali & 2.106 & 3.355 & 1.59 \\
Gujarati & 1.917 & 3.168 & 1.65 \\
Hindi & 1.649 & 2.731 & 1.66 \\
Kannada & 2.343 & 4.219 & 1.80 \\
Malayalam & 2.832 & 4.958 & 1.75 \\
Marathi & 1.944 & 3.767 & 1.94 \\
Tamil & 2.314 & 5.964 & \textbf{2.58} \\
Telugu & 2.363 & 4.064 & 1.72 \\
\midrule
Macro average & 2.183 & 4.028 & 1.84 \\
\bottomrule
\end{tabular}
\end{table}

The native column reproduces the standalone Round-09 fertility result closely (macro 2.183 versus 2.1773, with the largest per-language difference 0.064). The romanized samples are controlled ITRANS transliterations, not naturally occurring Latin-script web text, so the ratios isolate script conversion rather than broader domain differences.

The expansion did not directly target Punjabi or Odia. Their tokenizer efficiency should be measured separately before extending the eight-language fertility conclusions to them.

\subsection{Initializing the new matrices}

The input embedding and output projection are separate matrices (\texttt{tie\_word\_embeddings=false}). Each new BPE piece has an exact decomposition into upstream pieces. Its new row is initialized as the mean of those constituent rows, independently for the input and output matrices:

\begin{equation}
  e_{\mathrm{new}}(t)
  = \frac{1}{|S(t)|}\sum_{s\in S(t)} e_{\mathrm{old}}(s),
  \qquad
  w_{\mathrm{new}}(t)
  = \frac{1}{|S(t)|}\sum_{s\in S(t)} w_{\mathrm{old}}(s).
\end{equation}

\keypoint{A new token starts near the representation of the old tokens that used to spell it. This is a much better starting point than a random vector, while preserving every inherited row exactly.}

\subsection{A tokenizer pathology worth reporting}

One continued-BPE entry represented a degenerate space-plus-newline sequence. It could still be emitted through direct vocabulary lookup even after its merge path was removed. Because its initialized row was pathological, this single high-frequency whitespace entry contaminated many English encodings.

The slot was made unreachable without changing vocabulary size and was later repurposed as the atomic \texttt{<|file\_sep|>} token at ID 144891. This preserves tensor alignment and provides a transparent provenance explanation for a special token located inside the vocabulary-extension range.

The broader lesson is simple: after tokenizer surgery, audit frequent encodings in every retention language. Merge-table reasoning alone is insufficient because production tokenizer backends may emit atomic vocabulary entries independently of the merge path expected by the model author.

\section{Embedding Warmup}

\subsection{Why a separate stage is needed}

Vocabulary expansion modifies only a small part of the model, but those rows participate in every read and every output softmax. The first adaptation stage therefore freezes the inherited backbone and inherited vocabulary rows while updating the new rows in both untied matrices.

The warmup consumes approximately 3B tokens from a 90\% Indic / 10\% English-anchor mixture. The English slice is small enough to keep the objective focused on the new entries, but large enough to preserve the competition between new and inherited output rows.

\begin{table}[htbp]
\centering
\caption{Embedding-only warmup configuration.}
\begin{tabularx}{0.9\textwidth}{@{}L{4.2cm}Y@{}}
\toprule
\textbf{Setting} & \textbf{Value} \\
\midrule
Trainable parameters & New vocabulary rows in the untied input and output matrices \\
Frozen parameters & Backbone and all inherited vocabulary rows \\
Data mixture & 90\% Indic, 10\% English anchor \\
Training tokens & $\sim$3B \\
Peak learning rate & $1\times10^{-4}$ \\
Parallelism & TP=1, EP=8, PP=1 \\
Retention gate & HellaSwag plus bit-exact checks on frozen rows \\
\bottomrule
\end{tabularx}
\end{table}

\subsection{Decoupled weight decay can move frozen rows}

Setting a gradient to zero does not necessarily freeze a parameter under AdamW. The decoupled weight-decay update contains a direct parameter term:

\begin{equation}
  \theta_{t+1}
  = \theta_t - \eta \widehat{m}_t
    - \eta\lambda\theta_t.
\end{equation}

If the gradient-derived term $\widehat{m}_t$ is zero, the final term can still move $\theta_t$. The first warmup implementation therefore allowed inherited rows to drift by about 4\% in relative magnitude over 8,000 iterations, enough to collapse HellaSwag near random performance.

The fix snapshots inherited rows before each optimizer step and restores them immediately afterward. The guard is tensor-parallel aware, and a bit-exact comparison is run as a training invariant.

\subsection{Recovery trajectory}

\begin{figure}[htbp]
  \centering
  \includegraphics[width=0.9\textwidth]{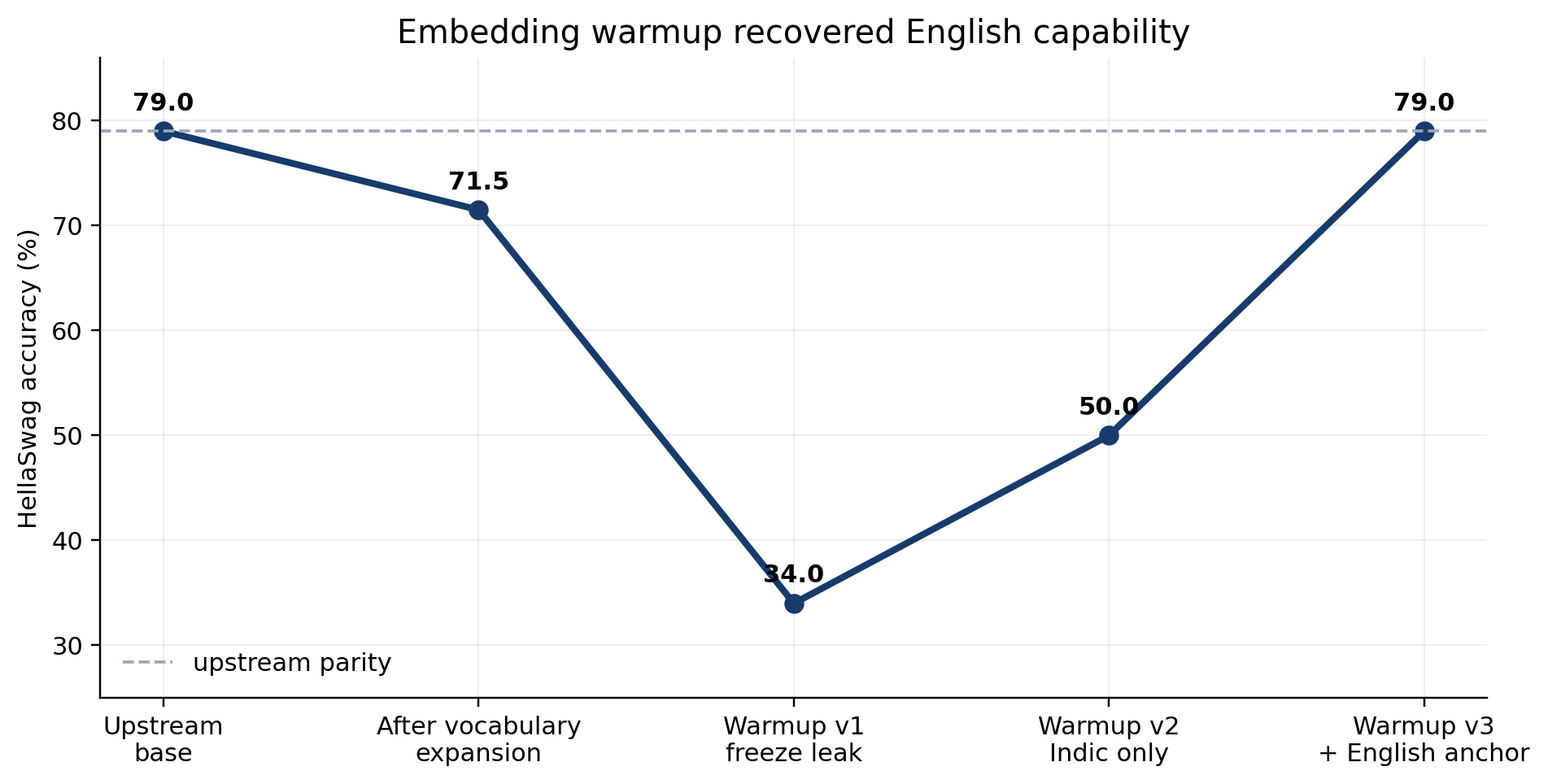}
  \caption{English-quality recovery across tokenizer initialization and embedding-only warmup.}
  \label{fig:warmup}
\end{figure}

Figure~\ref{fig:warmup} shows the sequence of failures and corrections. Constituent-mean initialization reduced HellaSwag from 79.0\% to 71.5\%. The freeze-leak run reached 34.0\%. An Indic-only warmup recovered to 50.0\%, while the final anchored warmup returned to 79.0\%.

This result is sometimes summarized incorrectly as a consequence of tied embeddings. The model is untied. The observed effect is instead about calibration: introducing and updating thousands of output rows changes the softmax competition faced by inherited tokens. A small retention-language anchor prevents the adaptation objective from optimizing only the new-language distribution.

\section{Continual Pretraining}

\subsection{Data preparation and quality control}

The CPT corpus combines English skill-carrying data, including code, mathematics, reasoning, and high-quality web text, with Indic web, academic, reference, and translated material~\cite{indicllmsuite}. The available Indic pool used for planning contains approximately 691B tokens.

Data processing applies exact deduplication followed by inter-document fuzzy deduplication. The implementation follows the practical ordering used in DataComp-LM processing: remove byte- or content-identical records first, then identify near-duplicate documents using fuzzy signatures~\cite{dclm}. Doing exact deduplication first reduces both compute and the number of trivial collisions seen by the fuzzy stage.

Every Indic document receives a continuous quality score and a discrete high-, medium-, or low-quality tier. The scorer combines structural length, script purity, character validity, repetition, boilerplate, URL density, language-model perplexity, and safety signals. Calibrated thresholds on the soft score assign the high, medium, and low tiers. Independent hard gates can reject a document regardless of its soft score.

\begin{table}[htbp]
\centering
\caption{Quality-scoring signal groups.}
\begin{tabularx}{0.92\textwidth}{@{}L{3.5cm}Y L{2.5cm}@{}}
\toprule
\textbf{Group} & \textbf{Examples} & \textbf{Role} \\
\midrule
Length and structure & word count, mean word length, mean line length & soft score \\
Script quality & target-script ratio, Latin ratio, invalid-character ratio & soft score / gate \\
Repetition & repeated words, characters, and $n$-grams & soft score \\
Web noise & symbol ratio, URL density, boilerplate density & soft score \\
Language fit & per-language KenLM log perplexity & soft score \\
Safety & NSFW and unsafe-content ratios & hard gate \\
\bottomrule
\end{tabularx}
\end{table}

Synthetic translations and rewrites are down-weighted in broad CPT and excluded from the highest-quality annealing slice. The aim is not to avoid synthetic data entirely. It is to prevent repeated translation style from dominating the late-stage distribution.

\subsection{Choosing the learning rate}

A four-arm, 1B-token-per-arm sweep compared peak learning rates of $10^{-4}$, $5\times10^{-5}$, $10^{-5}$, and $5\times10^{-6}$. Validation perplexity preferred $10^{-4}$, but HellaSwag retention preferred $10^{-5}$.

The disagreement is explainable. Validation perplexity was computed on a holdout close to the 90\%-Indic adaptation mixture. It therefore rewards rapid Indic fit even when that movement costs English retention. The deployment objective was multilingual gain under a retention constraint, so $10^{-5}$ was selected.

\begin{figure}[htbp]
  \centering
  \includegraphics[width=0.88\textwidth]{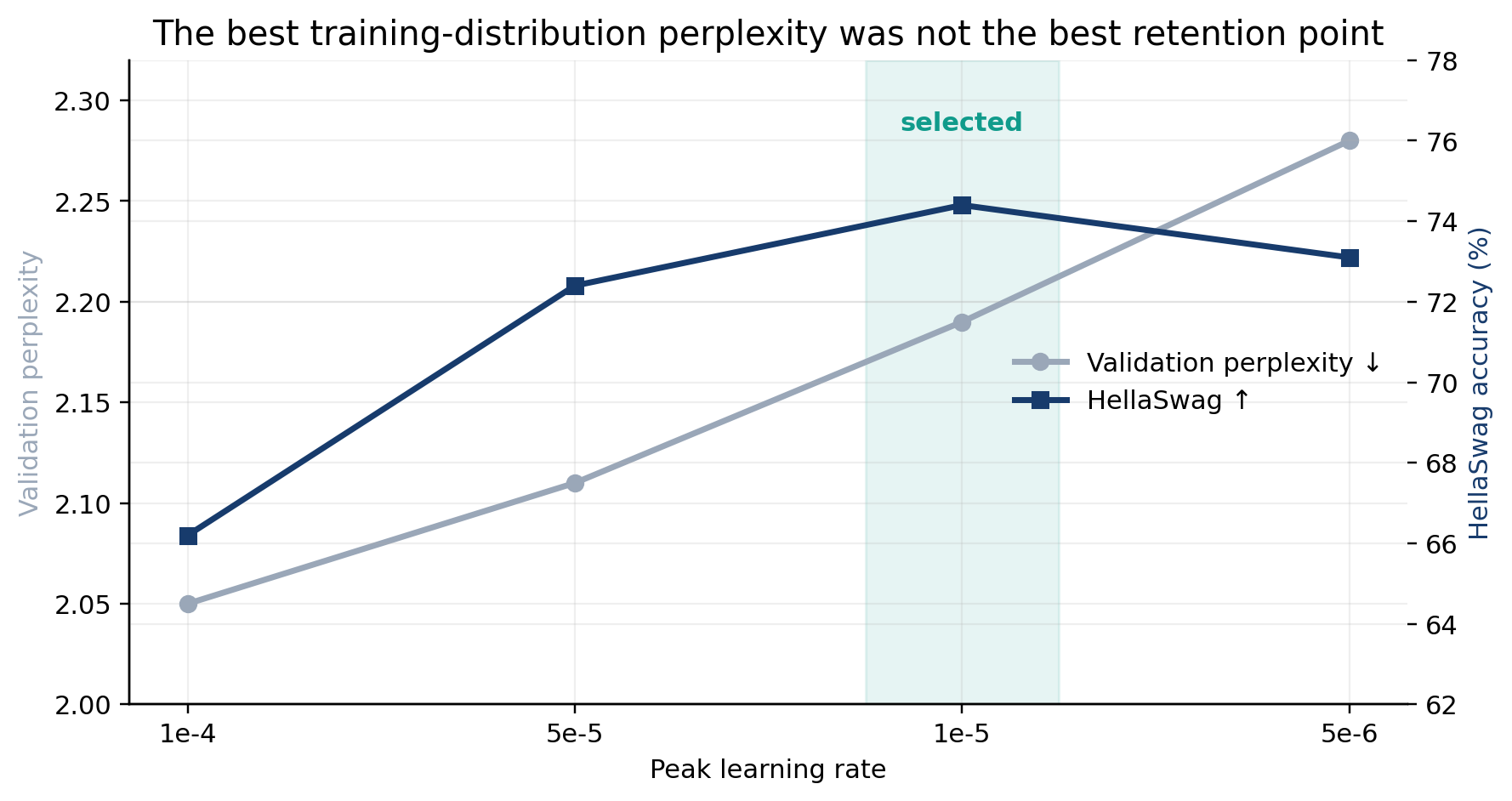}
  \caption{Learning-rate selection. Perplexity and downstream retention rank the candidates differently.}
  \label{fig:lr}
\end{figure}

\keypoint{Here, lower Indic-heavy validation perplexity did not imply a better multilingual model.}

The selected schedule uses a peak learning rate of $10^{-5}$, a minimum floor of $10^{-6}$, a linear warmup, and a Warmup--Stable--Decay schedule with a minus-square-root decay tail. Broad CPT uses a global batch of 2,048 sequences at length 8,192.

\subsection{Measured CPT performance ceiling}

The production topology uses 256 H200 GPUs. Expert parallelism keeps each expert group within one
eight-GPU node and its NVLink fabric. Table~\ref{tab:cptperformance} gives the stage-level settings
and measured throughput; Appendix~\ref{app:cptconfig} records the full configuration.

\begin{table}[htbp]
\centering
\caption{Measured steady-state performance across the three CPT stages. MFU uses the 989 TFLOP/s
BF16 peak.}
\label{tab:cptperformance}
\small
\begin{tabular}{@{}lrrrrr@{}}
\toprule
\textbf{Stage} & \textbf{s/iter} & \textbf{GFLOP/tok} & \textbf{tok/s/GPU} & \textbf{TFLOP/s/GPU} & \textbf{MFU} \\
\midrule
Broad CPT, 8k, CP=1 & 6.05 & 22.79 & 10{,}826 & 246.8 & 24.95\% \\
Annealing, 8k, CP=1 & 5.84 & 22.79 & 11{,}226 & 255.9 & 25.87\% \\
Long context, 128k, CP=8 & 159.77 & 40.91 & 6{,}563 & 268.5 & 27.15\% \\
\bottomrule
\end{tabular}
\end{table}

Aggregate throughput peaks at 2.87M tokens/s during annealing and falls to 1.68M during long-context
training. Utilization still rises because attention adds more work per token.

Appendix~\ref{app:throughputdiagnostics} reports timing and utilization, while Appendix~\ref{app:cptablations} records non-lineage checks.

\subsection{The realized training ladder}

The training ladder contains three distinct roles. The broad phase builds Indic capacity without abandoning English skills. Annealing concentrates the distribution on higher-quality sources. The last phase trains directly on long documents at 131,072 tokens.

\begin{figure}[htbp]
  \centering
  \includegraphics[width=0.78\textwidth]{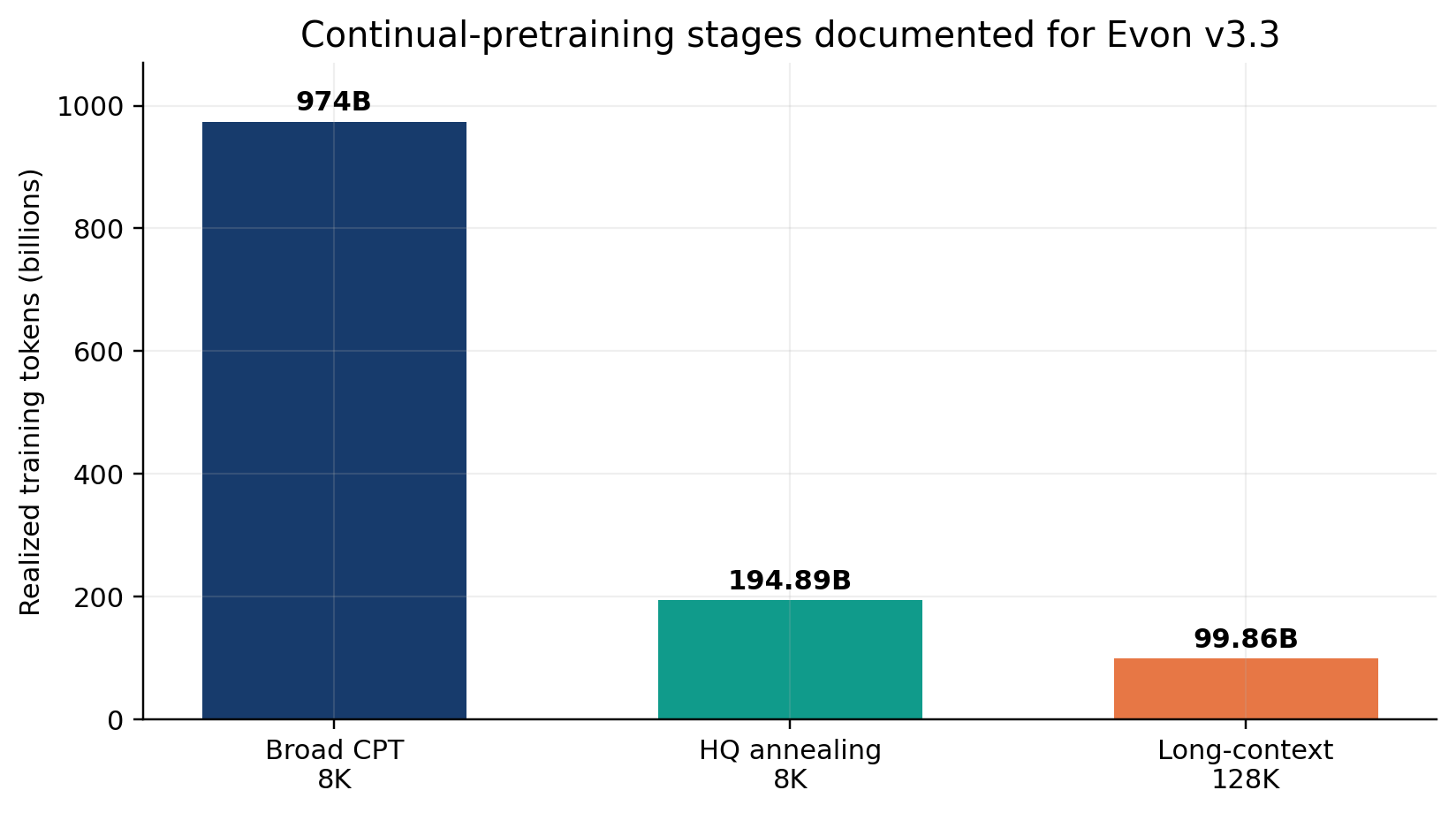}
  \caption{Realized production CPT token counts. The 128K extension consumed 99.86B of its 100B budget.}
  \label{fig:cptstages}
\end{figure}

\begin{table}[htbp]
\centering
\caption{Planned and realized continual-pretraining stages.}
\label{tab:cptstages}
\begin{tabularx}{\textwidth}{@{}L{2.8cm}L{2.4cm}L{2.7cm}Y@{}}
\toprule
\textbf{Stage} & \textbf{Context} & \textbf{Realized tokens} & \textbf{Role} \\
\midrule
Broad CPT & 8,192 & $\sim$974.49B & Diverse English--Indic skill and coverage phase. \\
HQ annealing & 8,192 & 194.89B & Higher-quality continuation initialized from the best broad-CPT validation checkpoint. \\
Long-context extension & 131,072 & $\sim$99.86B & Direct NoPE long-document training; 372 of $\sim$373 target iterations. \\
\bottomrule
\end{tabularx}
\end{table}

The long-context stage consumed 99.86B tokens and stopped one iteration short of its nominal 100B
budget. Table~\ref{tab:cptstages} gives the batch, length, and stage totals.

Across broad CPT, annealing, and long-context training, the documented realized total is 1.269T tokens.

Appendix~\ref{app:cpttraces} reproduces the available training panels for all three CPT stages,
including learning rate, gradient norm, language-model loss, and MoE load balancing. The phase-1
panel covers the exported interval, not every retained checkpoint.

\subsection{Checkpoint evaluation across CPT phase 1 and annealing}

\begin{figure}[htbp]
  \centering
  \includegraphics[width=0.98\textwidth]{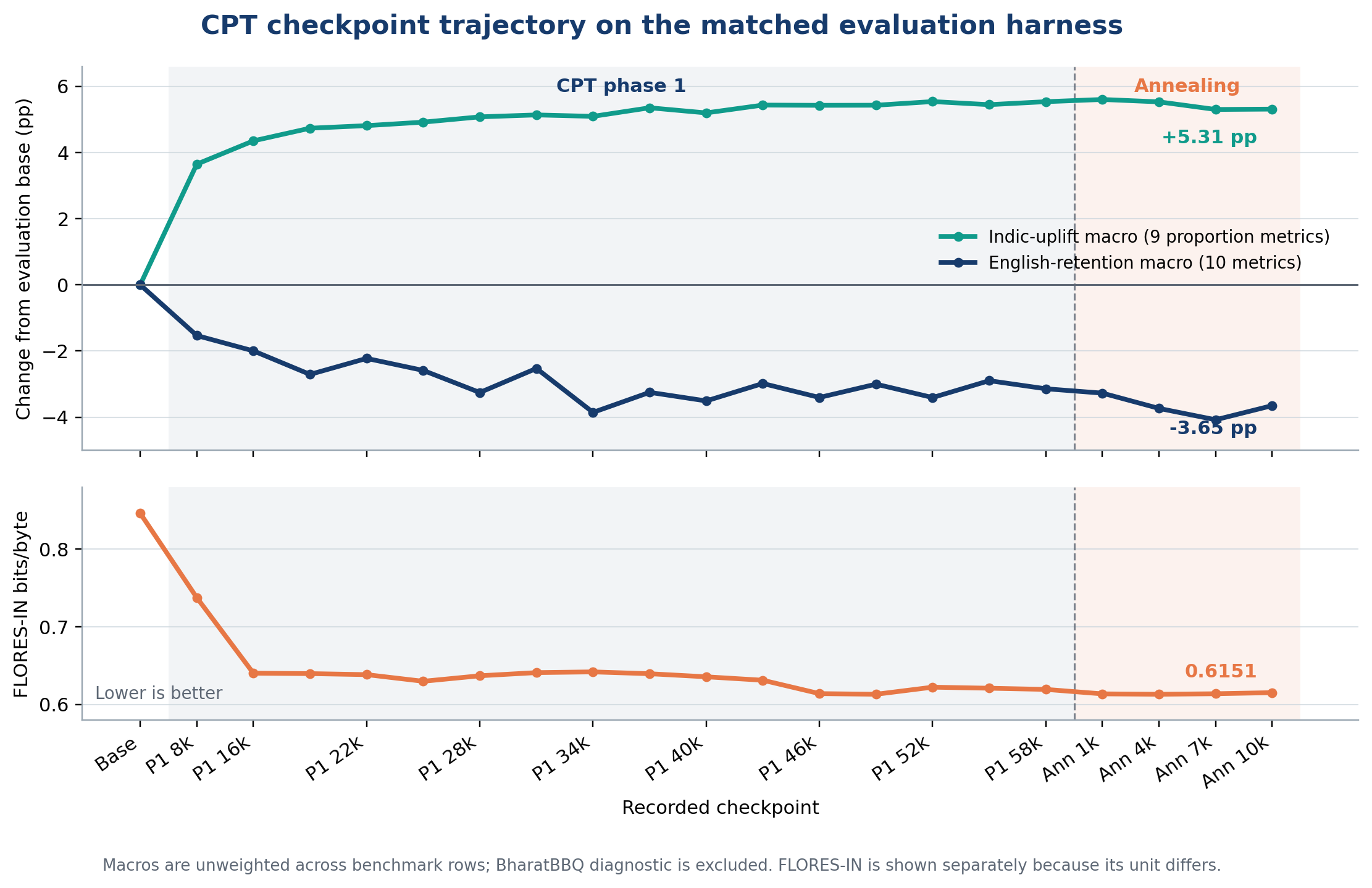}
\caption{Matched CPT checkpoint trajectory. The upper panel shows Indic uplift and English retention;
the lower panel shows FLORES-IN bits per byte. The divider marks the start of annealing.}
  \label{fig:cptcheckpointtrajectory}
\end{figure}

The matched checkpoint series shows where CPT changes the model. Figure~\ref{fig:cptcheckpointtrajectory}
traces the trajectory, Table~\ref{tab:cptcheckpointsummary} gives selected endpoints, and
Appendix~\ref{app:cptevals} reports the complete 21-benchmark series. Absolute scores from this
series should not be mixed with the earlier embedding-warmup evaluation.

\begin{table}[htbp]
\centering
\caption{Selected matched results across the CPT stages. The long-context column is \evonbase{}.}
\label{tab:cptcheckpointsummary}
\begin{tabularx}{\textwidth}{@{}Y L{1.7cm}R{1.4cm}R{1.5cm}R{2.0cm}R{2.0cm}@{}}
\toprule
\textbf{Benchmark} & \textbf{Metric} & \textbf{Base} & \textbf{Phase 1} & \textbf{Annealing} & \textbf{Long context} \\
\midrule
Indic accuracy macro & accuracy (\%) & 41.09 & 46.83 & 46.63 & 46.83 \\
MILU & accuracy (\%) & 40.63 & 53.54 & 53.52 & 53.24 \\
ARC-Challenge-Indic & accuracy (\%) & 35.23 & 48.20 & 48.20 & 47.09 \\
IndicGLUE-selected & score (\%) & 52.12 & 57.86 & 57.22 & 57.97 \\
IndicMMLU-Pro & accuracy (\%) & 14.44 & 20.03 & 20.02 & 19.44 \\
Indic BPB overall & bits/byte $\downarrow$ & 0.8453 & 0.6189 & 0.6132 & 0.6177 \\
English 9-bench mean & accuracy (\%) & 70.62 & 67.98 & 68.40 & 67.95 \\
MMLU & accuracy (\%) & 78.50 & 75.20 & 75.11 & 75.28 \\
HellaSwag & accuracy (\%) & 85.57 & 83.52 & 83.69 & 83.69 \\
\bottomrule
\end{tabularx}
\end{table}

Phase 1 produces most of the movement: broad Indic gains with a measurable English-retention cost.
Annealing preserves those gains and further improves FLORES-IN bits per byte, while English results
remain mixed. Because later SFT and RL evaluations use different protocols, only the trajectory
within this series should be compared.

\subsection{Ablation-guided language allocation}

A 120B-token ablation showed that the weakest starting languages gained most, while a more Indic-heavy
mixture did not improve the aggregate. These non-lineage studies informed later language allocation;
Appendix~\ref{app:cptablations} gives the results and planning weights.

\subsection{Long-context mixture}

The long-context manifest balances long Indic and English documents with mathematics, long-context SFT-style data, code, retention data, and a smaller multilingual component.

\begin{table}[htbp]
\centering
\caption{Composition of the 100B-token long-context pool. The selected run consumed $\sim$99.86B tokens.}
\label{tab:longctxpool}
\begin{tabular}{@{}lrrr@{}}
\toprule
\textbf{Bucket} & \textbf{Weight} & \textbf{Shards} & \textbf{Pool tokens} \\
\midrule
Indic long documents & 0.22 & 9 & 22.0B \\
Indic high-quality & 0.15 & 132 & 15.0B \\
English long documents & 0.20 & 7 & 20.0B \\
Mathematics and STEM & 0.14 & 1 & 14.0B \\
Long-context supervised data & 0.10 & 3 & 10.0B \\
Code & 0.08 & 1 & 8.0B \\
Base-retention data & 0.07 & 2 & 7.0B \\
Other multilingual & 0.04 & 2 & 4.0B \\
\bottomrule
\end{tabular}
\end{table}

We call the post-long-context checkpoint \textbf{\evonbase{}}, and all later training starts from it.
Appendix~\ref{app:cptevals} compares it with the earlier CPT checkpoints.

\section{Supervised Fine-Tuning}

\subsection{Role in the model story}

Continual pretraining leaves \evonbase{} fluent in the target languages and domains, but that does not by itself define a helpful assistant. SFT teaches the chat protocol: how to follow a request, produce a final answer, call tools, maintain a conversation, and use the requested language and script.

We call the public-facing SFT stage \textbf{\evonit{}}. Its retained path is \sftpone{} $\rightarrow$ \sftptwo{}
$\rightarrow$ \sftrthree{} $\rightarrow$ corrected epoch 2. A continuation after \sftptwo{} was discarded because of a
data-splitting error, and the curriculum was rebuilt as \sftrthree{}. Experimental runs outside this path are
not ancestors of the released RL model.

\begin{table}[htbp]
\centering
\caption{SFT lineage relevant to the released RL checkpoint.}
\label{tab:sftlineage}
\begin{tabularx}{0.92\textwidth}{@{}L{3.3cm}L{3.2cm}Y@{}}
\toprule
\textbf{Step} & \textbf{Disposition} & \textbf{Role} \\
\midrule
\sftpone{} & retained ancestor & first supervised curriculum stage \\
\sftptwo{} & retained ancestor & second supervised stage \\
intervening continuation & discarded & affected by data-point splitting; not in the released lineage \\
\sftrthree{} epoch 1 & retained branch point & first pass over the rebuilt \sftrthree{} corpus. Iteration 2,000 is the branch point for epoch 2 \\
\evonit{} (\sftrthree{} epoch 2) & selected checkpoint & epoch-2 continuation from \sftrthree{} epoch-1 iteration 2,000;\textsuperscript{\ensuremath{\dagger}} direct initialization for RL \\
\addlinespace
\multicolumn{3}{@{}p{0.90\textwidth}@{}}{\footnotesize \textsuperscript{\ensuremath{\dagger}}Epoch 2 restarts from \sftrthree{} epoch-1 iteration 2,000 with fresh optimizer state; later epoch-1 checkpoints showed elevated MoE load-balancing loss.} \\
\bottomrule
\end{tabularx}
\end{table}

The retained SFT stages also differ in corpus composition. Table~\ref{tab:sftstagesummary} gives the
row-level summary, and Appendix~\ref{app:sftmix} gives the language mix and approximate source-family shares.
The selected \evonit{} checkpoint uses the \sftrthree{} corpus, whose materialized rows are 64.0\% English.

\begin{table}[htbp]
\centering
\caption{Materialized corpus summary across the retained SFT lineage.}
\label{tab:sftstagesummary}
\begin{tabularx}{\textwidth}{@{}L{3.3cm}R{2.8cm}R{2.0cm}Y@{}}
\toprule
\textbf{Stage} & \textbf{Materialized rows} & \textbf{English share} & \textbf{Principal change} \\
\midrule
\sftpone{} & $\sim$12.5M & 50.7\% & Era-1 phase-1 pool; instruction following, mathematics, and code dominate. \\
\sftptwo{} & $\sim$14.7M & 54.6\% & Era-1 phase-2 pool; code, science, and agentic shares increase. \\
\sftrthree{} & $\sim$15.2M & 64.0\% & Era-2 rebuild with a larger English share. \\
\bottomrule
\end{tabularx}
\end{table}

\subsection{Selected \evonit{} checkpoint}

RL initializes from the iteration-2,400 export of the corrected \sftrthree{} epoch-2 run. It represents
80.53B scheduled sequence positions, including padding. A direct sample of the assistant-only loss
mask estimates roughly 15B loss-bearing positions. The run is stable at this point, with a recorded
language-model loss of 0.614. Appendix~\ref{app:sftcheckpoint} gives the full export, packing, and
schedule accounting.

\subsection{Data construction}
\label{sec:sftdata}

The SFT mixture covers English and eight Indic languages in native and romanized scripts~\cite{aya,crossgen,monoormulti}. It includes
question answering, reasoning, mathematics, summarization, instruction following, tool interaction,
and multi-turn dialogue. The final \sftrthree{} build is 64.0\% English and 36.0\% Indic by materialized rows.
Appendix~\ref{app:sftmix} gives the source families; these row shares do not measure token exposure.

Gnani-published datasets use the Apache-2.0 terms stated on the Evon repository. Upstream datasets
retain their original terms and attribution requirements.

Translation is structure-aware. User-facing prose can be translated, while code, mathematical spans,
tool schemas, function names, and machine-readable arguments are preserved byte-for-byte. Failed or
incomplete translations are quarantined rather than replaced with English text.

Conversations are normalized into a common schema with visible answer content separated from reasoning content. Deduplication is applied after normalization so that equivalent conversations stored in different source formats can match.

\subsection{Masking and long-sequence packing}

Training loss is applied only to assistant spans that should be learned. System and user history remain visible as context but are masked from the objective. Tool chains remain intact so that intermediate assistant actions and tool observations form one coherent training example.

Packing places multiple conversations into 131,072-token sequences while preserving their boundaries.
Table~\ref{tab:sftconfig} records the production topology and optimizer.

\begin{table}[htbp]
\centering
\caption{SFT configuration for the long-context production family.}
\label{tab:sftconfig}
\begin{tabularx}{0.88\textwidth}{@{}L{4.2cm}Y@{}}
\toprule
\textbf{Setting} & \textbf{Value} \\
\midrule
Maximum sequence length & 131,072 \\
Precision & BF16 \\
GPUs & 256 NVIDIA H200 \\
Parallelism & TP=1, PP=1, CP=4, EP=8, DP=64 \\
Global / micro batch & 256 / 1 \\
Optimizer & Adam, $\beta_2=0.95$, $\epsilon=10^{-8}$ \\
Weight decay / clipping & 0.1 / 1.0 \\
Packing objective & high utilization with assistant-only loss \\
Downstream role & initialization for both RL investigations \\
\bottomrule
\end{tabularx}
\end{table}

Appendix~\ref{app:sfttraces} reproduces the available Weights \& Biases panels for \sftpone{}, \sftptwo{}, and the final \sftrthree{} stage. The stage-wise loss levels should not be compared as if they came from one stationary data distribution: each curriculum rebuild changes the source mixture, packing, and assistant-token composition.

\subsection{What SFT establishes}

The SFT initialization begins RL with strong native-script output: measured in-script rates range from 93.5\% to 99.1\% in one multilingual harness, and a later matched IndicIFEval-ground evaluation records 99.9\%. It also establishes the response format and tool-call syntax expected by the RL parsers.

The purpose of reporting these pre-RL conventions is not cosmetic. RL optimizes whatever is rewarded by its environments. If a convention is present in SFT but absent from every RL reward, it can disappear quickly even while benchmark reasoning improves.

\section{Reinforcement Learning}
\label{sec:rl}

\subsection{GRPO in one equation}

The released RL stage uses group relative policy optimization (GRPO)~\cite{deepseekmath}. For each
prompt it samples $G=16$ completions, centers each reward against the other fifteen, and scales all
advantages by the whole group's sample standard deviation:

\begin{align}
  \mu_{-i} &= \frac{1}{G-1}\sum_{j\ne i} r_j, \\
  \sigma &= \sqrt{\frac{1}{G-1}\sum_{j}{(r_j-\overline{r})}^2}, \\
  A_i &= \frac{r_i-\mu_{-i}}{\sigma+\epsilon}, \qquad \epsilon = 10^{-6},
\end{align}

The division is applied only for $\sigma>0$. Equal-reward groups contain no relative signal and are
resampled. Appendix~\ref{app:rlconfig} records the estimator details and the later experimental floor.

\keypoint{The model learns which completion was better than the other completions for the same prompt. It does not require a separately trained critic to assign an absolute value to each response.}

Dynamic sampling discards and replaces groups in which every completion receives the same reward.
Consequently, a dataset's nominal row share is not its effective training share:

\begin{equation}
  s_{\mathrm{effective},e}
  \propto s_{\mathrm{nominal},e}\,p_{\mathrm{mixed},e},
\end{equation}

where $p_{\mathrm{mixed},e}$ is the fraction of prompt groups in environment $e$ that contain more than one reward value. Realized shares must therefore be measured from rollout logs.

\subsection{Released \evonrl{}}
\evonrl{} is the public BF16 checkpoint and the source of the planned FP8 artifact. It initializes
from \evonit{} and trains on four NeMo-Gym task environments:

\begin{itemize}
  \item \textbf{math with judge}: final-answer equivalence after boxed-answer extraction;
  \item \textbf{structured outputs}: JSON validity and schema conformance;
  \item \textbf{workplace assistant}: correctness of an iterative tool-call sequence;
  \item \textbf{MCQA}: exact match on a selected option.
\end{itemize}

No custom reward shaping is applied in the released run. Each environment returns its own binary or task-specific scalar. There is no language reward, length reward, or repetition penalty.

\begin{table}[htbp]
\centering
\caption{Released \evonrl{} training configuration.}
\label{tab:releasedrlconfig}
\begin{tabularx}{0.9\textwidth}{@{}L{4.0cm}Y@{}}
\toprule
\textbf{Setting} & \textbf{Value} \\
\midrule
Initialization & \evonit{} \\
Algorithm & GRPO with leave-one-out baseline and whole-group standard-deviation normalization \\
Resolved estimator flags & \texttt{use\_leave\_one\_out\_baseline=true}; \texttt{normalize\_rewards=true} \\
Generations per prompt & 16 \\
Dynamic sampling & enabled; zero-variance groups resampled \\
Sampling temperature & 1.0 \\
Inference placement & colocated vLLM and training workers \\
Tool-call parser & Qwen3 XML parser \\
Reasoning parser & Nemotron-v3 parser \\
Custom reward shaping & none \\
\bottomrule
\end{tabularx}
\end{table}

\begin{figure}[htbp]
  \centering
  \includegraphics[width=0.98\textwidth]{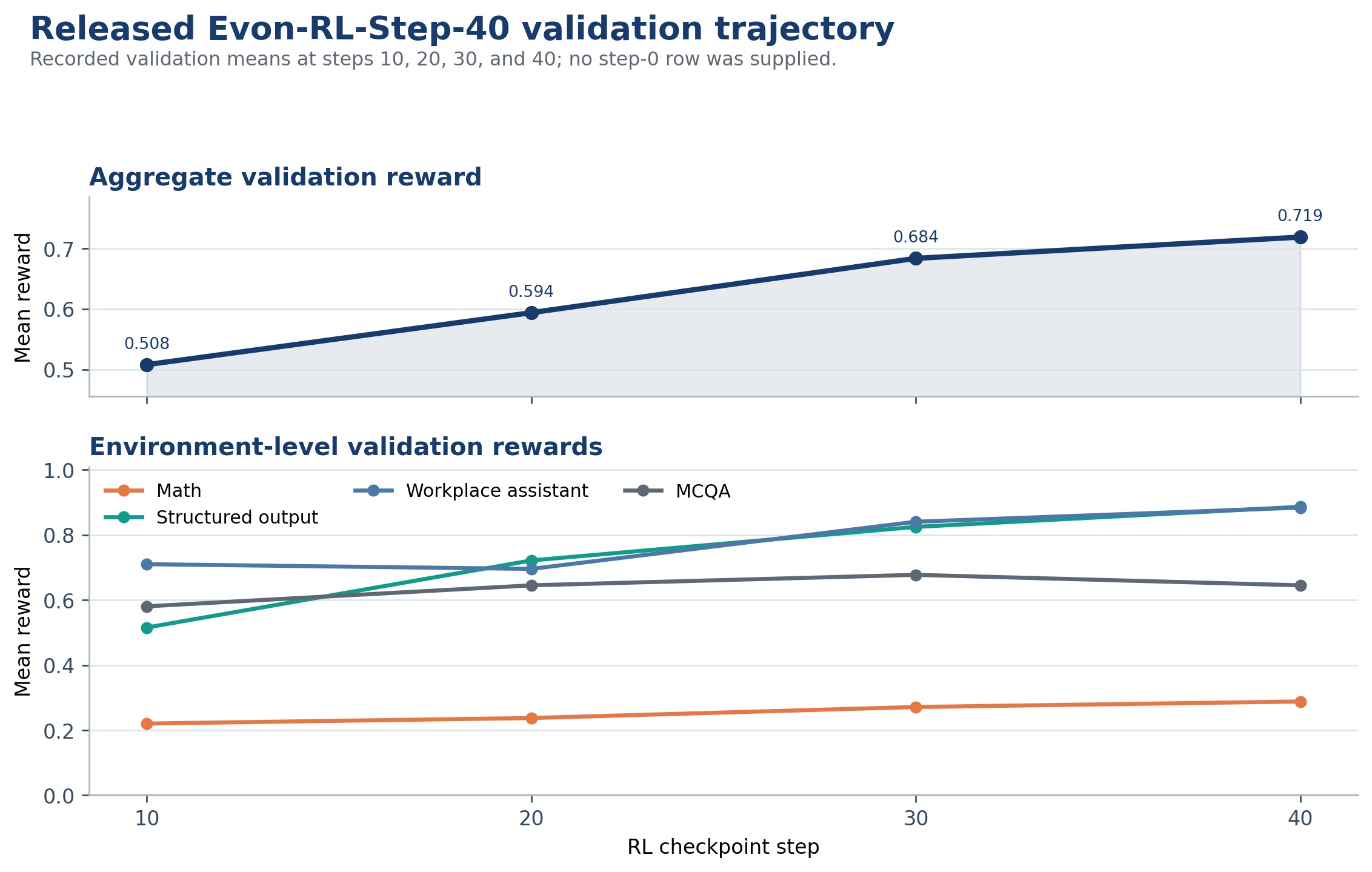}
\caption{Released-run validation rewards. Structured output and workplace assistance improve most;
the curves are optimization diagnostics, not benchmark accuracies.}
  \label{fig:releasedrlvalidation}
\end{figure}

The largest gains appear on tasks represented directly in the reward mixture: MATH rises 36.07
points and GPQA Diamond 26.84. Competition mathematics moves little because the SFT checkpoint is
already strong there. Table~\ref{tab:releasedrlresults} gives the matched results; extractive-QA
comparisons use F1 because the exact-match rule changed between recorded runs.

The regressions cluster in Indic extractive question answering even as Indic free generation improves.
On XQuAD-Indic, the share answered in English rises from 8.7\% to 91.7\%~\cite{langmixing}. Conditional accuracy in
the target script roughly doubles, although this is a self-selected subset rather than a controlled
estimate.

English outputs are classified before grounding, so the lower ungrounded count is not independent
evidence of better reading. The evidence points mainly to output language, although a smaller
capability loss remains possible (Appendix~\ref{app:qataxonomy}).

This behavior follows the reward boundary. The run contains no language reward, and English
completions receive 0.1681 more reward on average, almost entirely from structured-output and
workplace-assistant tasks. IndicIFEval is sensitive to evaluation state; the canonical release-protocol
value is 33.05, and Appendix~\ref{app:qataxonomy} records the other readings.

\subsection{Capability versus convention}
\label{sec:capconv}

The released RL study separates capability from emission convention. On 5,000 MATH problems,
accuracy rises from 51.00 to 85.80, while conditional accuracy on boxed answers barely changes.
Changing only the boxed-answer emission mix reproduces 94\% of the gain; Appendix~\ref{app:matched}
gives the decomposition.

The exact decomposition depends on where the response is partitioned, but the conclusion does not:
a benchmark swing can be carried largely by an output convention. Appendix~\ref{app:matched}
records the alternate partition and the unboxed-answer counts.

\keypoint{RL can destabilize a cheap output convention without erasing the capability underneath. Track in-script rate, boxed-answer rate, parse success, and tool-call count alongside benchmark accuracy.}

This observation motivates a separate experimental branch: every product-critical convention needs a non-zero, correctly scoped training signal.

\section{Release Evaluation}

\subsection{Comparison with open models}
\label{sec:external}

We evaluate \shortmodel{} and six open checkpoints on the same 35 benchmarks. The suite combines
English knowledge, reasoning, and code benchmarks~\cite{mmlu,mathbench,gsm8k,gpqa,humaneval,livecodebench,ifeval}
with Indic question answering, instruction following, and generation
benchmarks~\cite{xquad,indicifeval,indicgenbench,mgsm}. Reference models use
their published sampling settings, so the most reliable reading is movement within the Evon lineage,
not a controlled ranking across providers. Two reference checkpoints emit a non-standard whitespace
marker on code tasks; those code results are excluded.

\begin{figure}[htbp]
  \centering
  \includegraphics[width=0.92\textwidth]{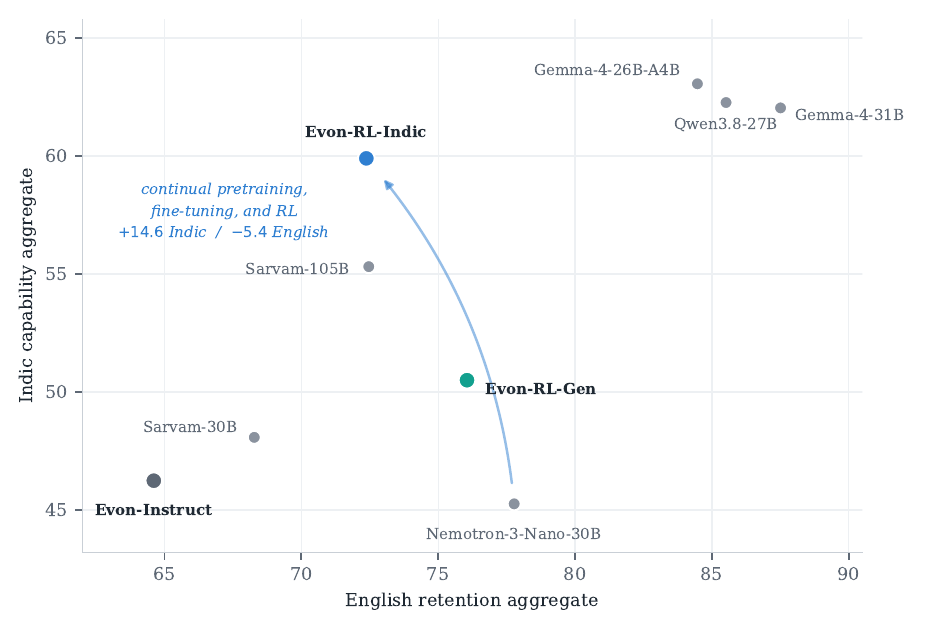}
\caption{English retention and Indic capability. The arrow shows movement from the
architecture-identical starting checkpoint; reference models use their published sampling settings.}
  \label{fig:evalpos}
\end{figure}

Figure~\ref{fig:evalpos} summarizes equal-weight English-retention and Indic-capability aggregates.
Relative to its architecture-identical starting point, the released model gains 5.2 Indic points while
giving up 1.7 English points. The unreleased Indic-aware branch gains 14.6 for 5.4. Larger open
models lead both aggregates; the figure is meant to show the movement produced by the pipeline.

\subsection{Safety evaluation}
\label{sec:safety}

IndicSafe evaluates 5,965 examples across 12 Indian languages. Overall accuracy is 86.50\%. The model is strongest on harmless controls and ambiguous-but-safe prompts. The hardest categories are political manipulation and gender/caste discrimination.

\begin{table}[htbp]
\centering
\caption{IndicSafe accuracy by harm category.}
\begin{tabular}{@{}lr@{}}
\toprule
\textbf{Category} & \textbf{Accuracy (\%)} \\
\midrule
Harmless control & 97.93 \\
Tricky / ambiguous & 97.77 \\
Offensive or hate speech & 89.80 \\
Harmful instructions & 85.62 \\
Misinformation / conspiracy & 83.50 \\
Health misinformation & 80.41 \\
Religious hate / provocation & 80.08 \\
Political manipulation & 72.49 \\
Gender and caste discrimination & 70.26 \\
\bottomrule
\end{tabular}
\end{table}

BBQ moves in opposite directions by context type. Disambiguated accuracy rises from 94.13 to 97.70
and bias falls from 0.99 to 0.60; ambiguous accuracy falls from 92.29 to 88.69 and bias rises from
4.33 to 7.39. The model appears more willing to commit, which helps when context resolves the answer
and increases stereotyped completions when it does not.

Safety scores are not a deployment guarantee. They measure one dataset and one decoding configuration. Applications should add domain-specific testing, moderation, logging, and human escalation.

\section{Experimental Research Branches (Not Released)}
\label{sec:branches}

Everything in this section belongs to research lineage, not the public release. The experiments are
included because their negative results transfer and clarify the limitation in
Section~\ref{sec:capconv}.

\subsection{A sibling Indic-aware branch}

\evonrlindic{} starts independently from the same \evonit{} export as the released run. It adds an
Indic instruction-following environment, a scoped language reward, and a repetition penalty. On
IndicIFEval-ground, the SFT score of 40.52 falls to 33.05 under the released mixture but rises to
60.00 under the experimental one. XQuAD-Indic F1 similarly moves from 36.51 to 17.19 and 50.44,
respectively.

Because several ingredients change together, these gains cannot be attributed to the language reward
alone. They do show that the released regression is a property of its reward mixture rather than an
inevitable effect of RL\@. The branch also learns a one-tool-call-per-turn policy, showing that a
targeted repair can create another convention. Appendix~\ref{app:phase2rl} gives the full analysis.

\begin{figure}[htbp]
  \centering
  \includegraphics[width=0.82\textwidth]{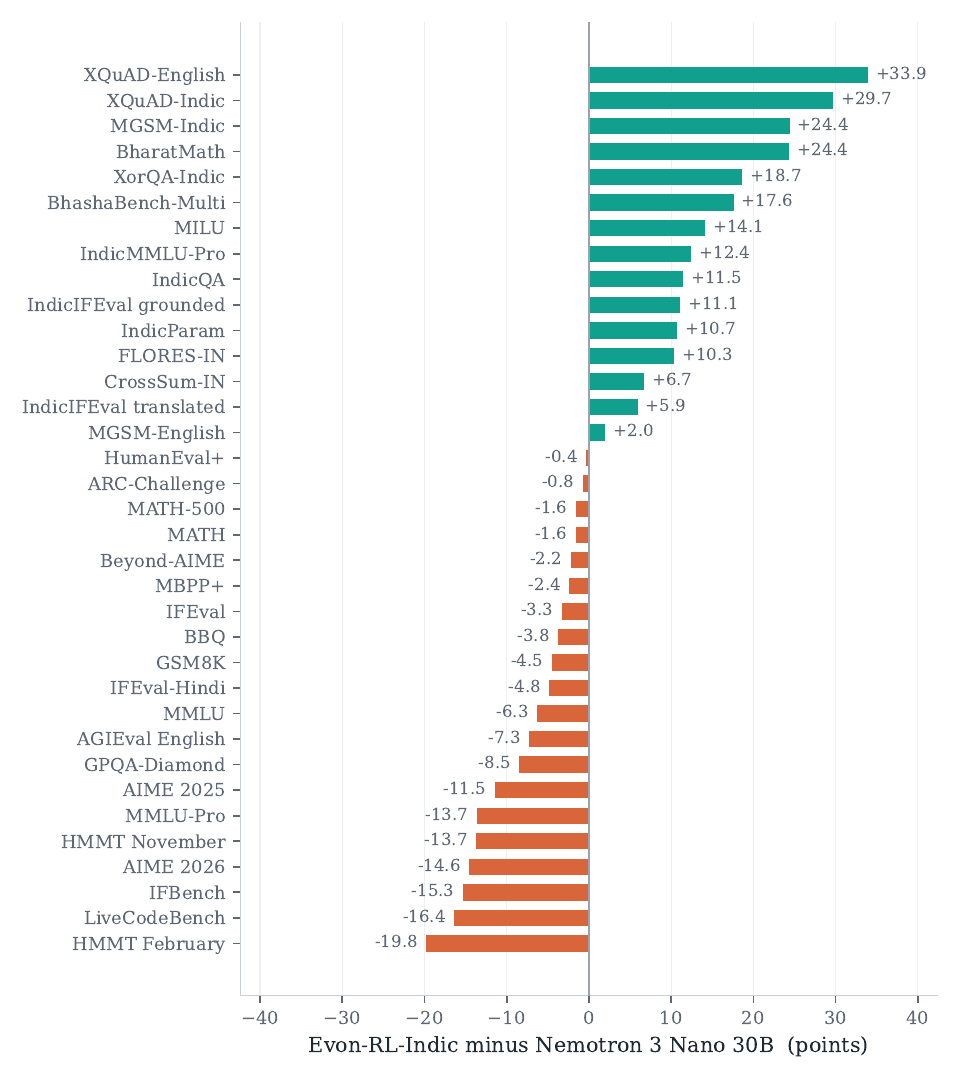}
  \caption{Per-benchmark change for the unreleased Indic-aware branch. Positive values favour the branch.}
  \label{fig:evaldelta}
\end{figure}

Thirteen of fourteen Indic benchmarks improve over the starting model, and eleven gain more than ten
points. The losses cluster in competition mathematics, code, and instruction following
(Figure~\ref{fig:evaldelta}).

\begin{figure}[htbp]
  \centering
  \includegraphics[width=0.96\textwidth]{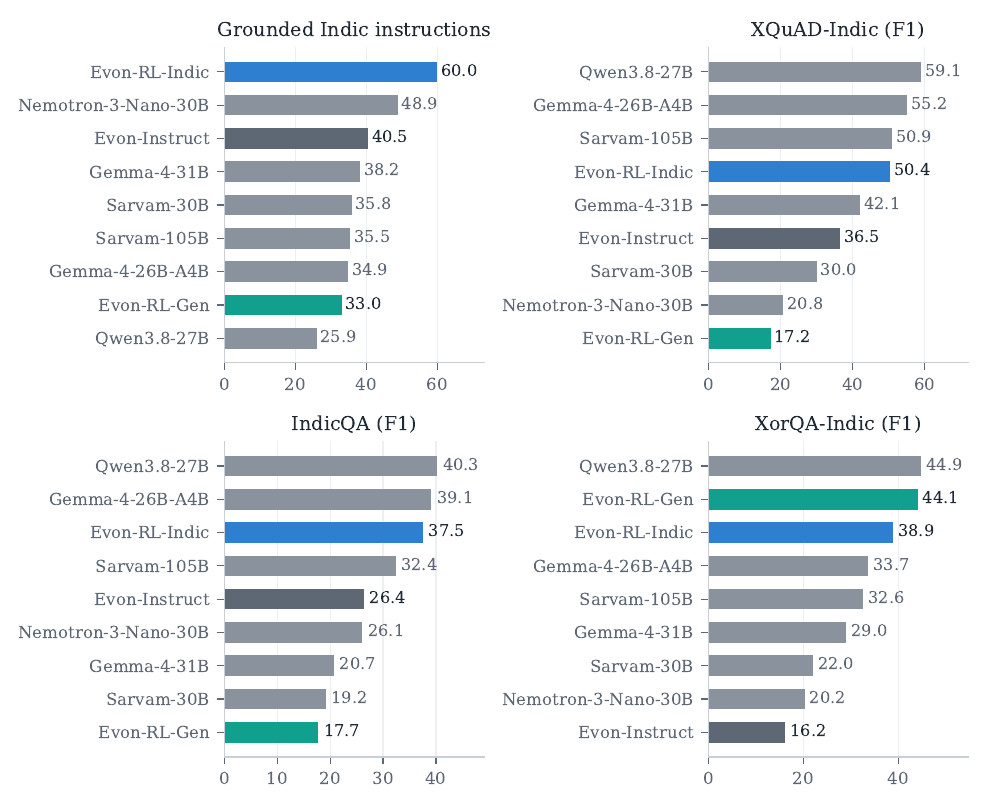}
  \caption{Grounded Indic instruction following and Indic extractive QA for the unreleased branch.}
  \label{fig:evalground}
\end{figure}

The branch reaches 60.0 on grounded Indic instruction following, compared with 48.9 for the strongest
open reference. It is competitive, rather than leading, on Indic extractive QA
(Figure~\ref{fig:evalground}).

\subsection{Learning from a judge}
\label{sec:judge}

Verifiable rewards only reach tasks with a checkable answer. Mathematics, multiple choice, and
schema conformance can be graded by a program. Open-ended helpfulness cannot. To extend
reinforcement learning past that boundary we trained a reward model, a judge, and then optimized
the policy against it. This section describes that judge, what training against it bought, and what
it cost. The branch is a research lineage continuing from \evonrlindic{}. It is not part of the
released model.

\paragraph{An Indic preference set had to be built first.}
We combine native prompts with translated HelpSteer3-style prompts in a 70:30 ratio. Two models
generate candidate responses and a third labels the preferred one. Of 98,744 pairs in Hindi, Telugu,
Tamil, and Malayalam, 98,327 receive labels; Appendix~\ref{app:judgeopd} gives the construction details.

\paragraph{The judge writes a verdict rather than emitting a scalar.}
Initialized from a 26B-A4B model, it reasons over two responses and returns calibrated helpfulness
scores plus a forced relative ranking. It scores 90.8 on RM-Bench against 87.3 for a 235B-parameter
judge, with most of the margin on the hard slice. The comparison is against model-generated labels,
however, and the judge has not been validated against human Indic preferences.

\paragraph{The instrument had the same disease as the patient.}
At a 4,096-token cap, JudgeBench falls from 88.3 to 83.7 because some responses never reach the
verdict tag. Thus 4.6 apparent quality points come from the judge's own emission convention;
Appendix~\ref{app:judgeopd} gives the completion-rate analysis.

\keypoint{A reward model is measured by benchmarks that carry the same emission-convention confound
as the policy's own. Check a judge's parse and completion rates before trusting its score, and
re-check them at the generation cap you actually train with.}

\paragraph{What training against the judge bought.}
The largest gains are on LiveCodeBench ($+13.2$) and MMLU-Pro ($+8.8$). Knowledge, reasoning,
grounded Indic instruction following, and mathematics also improve, while empty or unterminated
outputs fall on most gated benchmarks. Table~\ref{tab:judgepolicygains} gives the complete set of
headline gains; Appendix~\ref{app:judgeopd} also describes the ring comparison used for judging.

\noindent\textbf{What it cost: the answer, wrapped in a sentence.} Extractive question answering
loses 24--38 F1 points even though answer spans remain present at similar rates. The parent returns a
bare span; the judge-trained checkpoint wraps it in a sentence, adds a median of 9--17 tokens, and
reduces overlap precision. Conversational elaboration has transferred to a terse-answer task.

\noindent\textbf{What it cost: the requested language.} FLORES-IN falls 17 chrF++ points for a
different reason. English-target script compliance falls from 98.6\% to 24.5\%, and the model often
copies the Indic source instead. Monolingual extraction changes the form of a largely preserved
answer; cross-lingual tasks can lose the answer itself.

\paragraph{A length bonus drove reasoning to zero.}
Because the judge never sees hidden reasoning, reasoning length drifted. Rewarding the shortest trace
drove its mean from 6,020 characters to zero while the judge score improved. A two-sided target still
shortened reasoning by 60\% and cost more than 25 AIME points. The adopted design is neutral below a
generous budget and penalizes only traces above it.

\keypoint{Any bonus placed on the extreme of a property the model controls cheaply will be collected,
and a two-sided length target is an attractor. Constrain such properties with a one-sided bound
outside a permitted range rather than a reward inside it.}

\paragraph{Mixing verifiable rewards back in did not work.}
A 70:30 mix of judge and binary-verifier prompts also failed. All-pass and all-fail verifier groups
carry no relative signal, so dynamic sampling discarded 35--49\% of them while every judge group
survived. The intended mixture became roughly 80:20 at the gradient. The underlying pool contained
too many problems the model almost always solved or almost always missed;
Appendix~\ref{app:mixture} gives the full analysis.

\keypoint{When two reward sources are mixed, the realized mixture is the one that survives filtering,
not the one specified. Measure the share each source contributes at the gradient, and instrument
every environment rather than the one most likely to be healthy.}

\subsection{Recovering a convention by distillation}
\label{sec:opd}

The judge-trained checkpoint improves knowledge, code, and reasoning but loses useful conventions.
On-policy distillation recovers those conventions without another reward model or verifier.

\paragraph{The parent teaches the child.}
The student starts from the judge-trained checkpoint, while \evonrlindic{} serves as its teacher.
On the student's own responses, the teacher supplies a top-64 next-token distribution at every
position:
\begin{equation}
\mathcal{L}(\theta)=\mathbb{E}_{x,\;y\sim\pi_S}\Bigl[\tfrac{1}{|y|}\textstyle\sum_t
w\,\mathrm{KL}\bigl(p_T^{(t)}\,\|\,p_S^{(t)}\bigr)+(1-w)\,\mathrm{KL}\bigl(p_S^{(t)}\,\|\,p_T^{(t)}\bigr)\Bigr],
\qquad w=0.5.
\label{eq:opd}
\end{equation}
The two KL directions are weighted equally. One restores probability to forms retained by the
teacher; the other suppresses forms the teacher rejects. Teacher and student share an architecture
and tokenizer, and no reward is optimized during this stage.

\paragraph{Most of what was lost comes back, and most of what was gained stays.}
Across the six degraded benchmarks, distillation recovers 81\% of the loss; extractive QA recovers
74--99\%. Cross-lingual QA and translation return to their starting values within measurement noise.
The student also retains all of the MMLU-Pro gain, about three quarters of the code and mathematics
gains, at least half on MMLU, IndicMMLU-Pro, and AGIEval, and one third on GPQA Diamond.

\paragraph{The repair did not come from the repaired tasks.}
The 29,988 distillation prompts contain mathematics and multiple-choice questions, but no extraction,
translation, or code. The repair therefore comes from the teacher's token distribution rather than
task-specific examples. Unlike a sparse reward, that distribution supervises every position, even
when the desired form has become rare.

\paragraph{It is not free.}
Distillation moves the whole distribution, including behavior that did not need repair. IFEval falls
2.6 points, translated IndicIFEval 1.9, and Indic grade-school mathematics 4.8. Non-termination and
empty-output rates also worsen on some tasks.

\keypoint{A convention lost to reward optimization can be restored by distilling from the checkpoint
that still has it, without re-specifying the reward. The cost is that distillation moves everything,
including behavior that did not need repair.}

\subsection{Weight-space alternatives to distillation}
\label{sec:merge}

Two cheaper weight-space controls require no generation: equal interpolation with the parent and
extrapolation beyond the judge-trained checkpoint. Neither is a release candidate.

Table~\ref{tab:merge} reports how much of the extractive question-answering loss each method returns.
Distillation recovers 84\% of it across the four benchmarks, interpolation recovers 41\%, and
extrapolation does not recover any of it. Extrapolation instead moves those benchmarks further in the
direction the judge pushed them, ending well below the checkpoint it was meant to repair.

\begin{table}[htbp]
\centering
\small
\caption{Share of extractive-QA loss recovered by each repair. Negative values indicate further loss.}
\label{tab:merge}
\begin{tabular}{@{}lrrrr@{}}
\toprule
\textbf{Benchmark} & \textbf{Loss to repair} & \textbf{Distillation} & \textbf{Interpolation} & \textbf{Extrapolation} \\
\midrule
XQuAD-English & 38.25 & $+74.5\%$ & $+65.8\%$ & $-61.2\%$ \\
XQuAD-Indic   & 32.77 & $+89.9\%$ & $+32.1\%$ & $-25.2\%$ \\
IndicQA       & 24.45 & $+73.7\%$ & $+21.0\%$ & $-24.0\%$ \\
XorQA-Indic   & 24.46 & $+99.1\%$ & $+46.1\%$ & $-34.6\%$ \\
\midrule
\textbf{Macro} & & $\mathbf{+84.3\%}$ & $+41.3\%$ & $-36.2\%$ \\
\bottomrule
\end{tabular}
\end{table}

Interpolation preserves more of the judge's gains but repairs only half as much extractive QA as
distillation. Extrapolation makes the loss worse. Pushing farther along a reward direction therefore
accelerates damage to behavior that the reward ignores.

\section{FP8 Quantization and Serving}
\label{sec:fp8}

The FP8 checkpoint planned for release is produced from the frozen BF16 \evonrl{} export using post-training quantization, not quantization-aware training. The pipeline uses NVIDIA Model Optimizer 0.45 through Megatron-Bridge~\cite{modelopt}. No pruning, distillation, or INT4 weight-only compression is applied.

\subsection{Selective E4M3 W8A8 recipe}

Quantized sites use static E4M3 FP8 for both weights and inputs, with per-tensor symmetric scaling and no affine zero-point term. A max calibrator records the absolute maximum $a_{\max}$. Writing $F_{\max}$ for the largest finite E4M3 magnitude and $Q_{\mathrm{E4M3}}$ for rounding to the nearest representable E4M3 value with finite-range saturation, the conceptual quantize--dequantize operation is

\begin{equation}
  s = \frac{a_{\max}}{F_{\max}},
  \qquad
  q = Q_{\mathrm{E4M3}}\!\left(\frac{x}{s}\right),
  \qquad
  \widehat{x}=s q.
\end{equation}

Unlike INT8, E4M3 is a floating-point format with a non-uniform representable grid determined by its exponent and mantissa. Quantization is therefore an FP8 cast after scaling, not uniform rounding to an integer code. Zero is represented directly by the floating-point format rather than introduced through an affine quantization zero point.

\keypoint{Each quantized tensor uses one scale that maps its largest observed magnitude into the E4M3 range. The stored values use roughly half the bytes of BF16, while sensitive modules remain in BF16.}

The conservative recipe quantizes the routed and shared MoE expert input/output projections across all 23 MoE layers. Attention projections, attention matrix multiplications, Mamba projections and convolutions, routers, embeddings, final normalization, and the output head remain BF16.

\begin{table}[htbp]
\centering
\caption{Precision allocation in the FP8 artifact planned for release.}
\label{tab:fp8coverage}
\begin{tabularx}{\textwidth}{@{}L{4.1cm}L{3.7cm}Y@{}}
\toprule
\textbf{Component} & \textbf{Precision} & \textbf{Rationale} \\
\midrule
Routed expert FC1/FC2 & FP8 E4M3 W8A8 & dominant parameter and compute block \\
Shared expert FC1/FC2 & FP8 E4M3 W8A8 & same expert-matrix structure \\
Attention QKV/output and BMM & BF16 & excluded by conservative recipe \\
Mamba input/output projections and convolution & BF16 & protects recurrent sequence-mixing path \\
Routers / routing gates & BF16 & protects token-to-expert decisions \\
Embeddings, final norm, LM head & BF16 & protects vocabulary and output calibration \\
\bottomrule
\end{tabularx}
\end{table}

FP8 covers approximately 94\% of parameters and reduces the artifact from 63.5GB to 32GB, while
routers remain in BF16. Appendix~\ref{app:architectureaccounting} provides tensor-level accounting.

\subsection{Calibration and validation}

Calibration requested 1,024 multilingual examples but exhausted the source at 253 packed rows. All
artifact and paired-quality gates still pass, although a production recalibration should fill the
intended sample. Weight scales use weight maxima; activation scales come from the calibration pass.

Artifact gates verify FP8 expert weights, BF16 exclusions, scale tensors, the hybrid layer pattern,
tokenizer and EOS behavior, and the expected 32--34GB disk size. A no-op quantization that misses
fused three-dimensional expert tensors remains near 62GB and fails this gate.

\subsection{Serving performance}

\begin{figure}[htbp]
  \centering
  \includegraphics[width=0.86\textwidth]{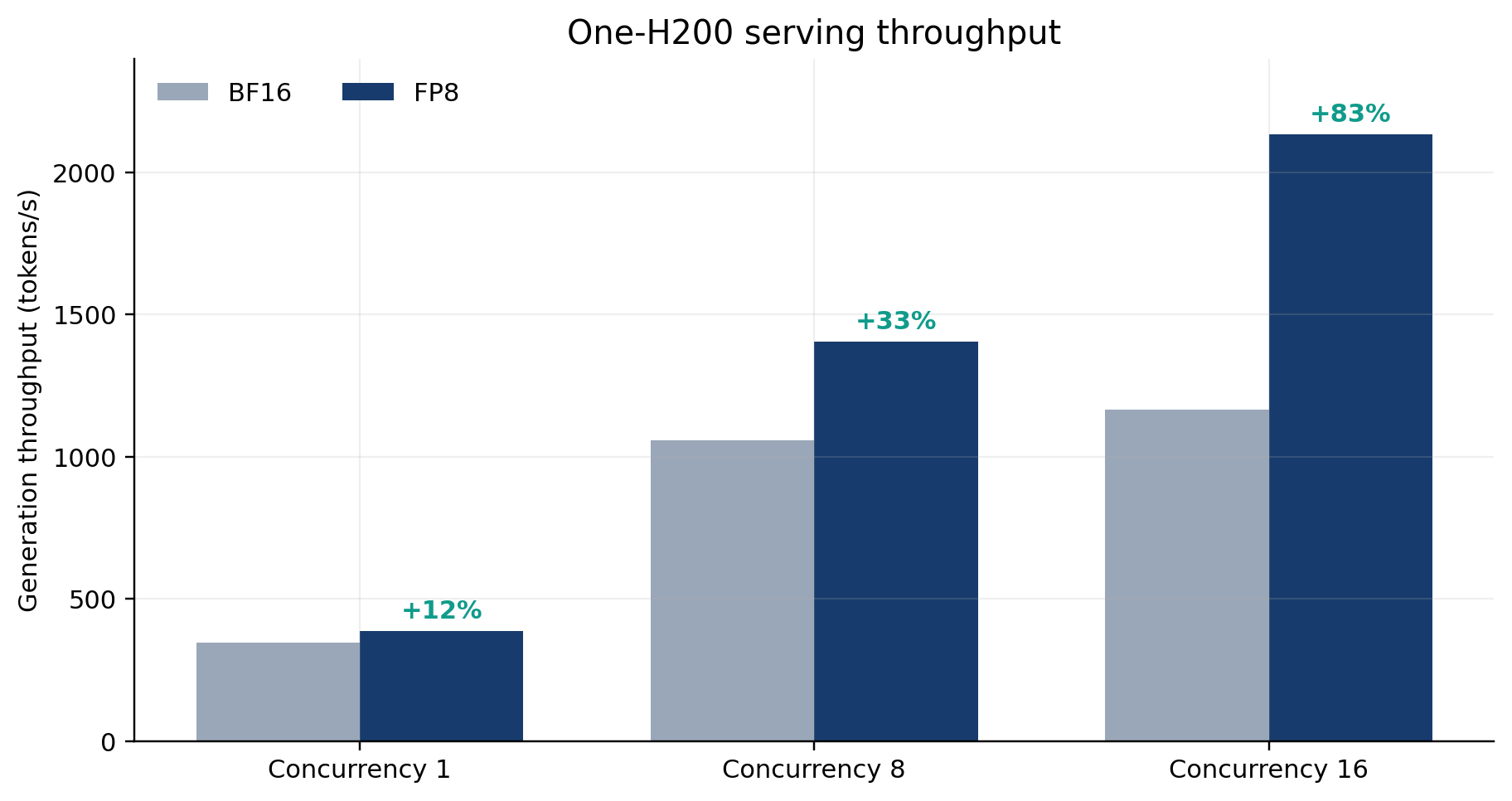}
  \caption{Paired BF16 and FP8 serving throughput on one NVIDIA H200 using the same serving stack.}
  \label{fig:fp8serving}
\end{figure}

BF16 and FP8 are served with the same stack on one H200 GPU\@. At concurrency 16, FP8 raises aggregate
throughput from 1,165 to 2,135 tokens per second and roughly doubles usable concurrency. The complete
latency and rate measurements appear in Table~\ref{tab:fp8serving} and Figure~\ref{fig:fp8serving}.

The gain differs from that of a dense transformer. Only six of the 52 layers contribute a growing
key-value cache; the Mamba layers carry context-independent state. FP8 saves resident weight memory,
increasing concurrency rather than usable context length.

\begin{table}[htbp]
\centering
\caption{Paired one-H200 serving measurements.}
\label{tab:fp8serving}
\begin{tabular}{@{}lrr@{}}
\toprule
\textbf{Metric} & \textbf{BF16} & \textbf{FP8} \\
\midrule
Warm TTFT, concurrency 1 & 348ms & 341ms \\
Warm generation rate, concurrency 1 & 345.7 tok/s & 387.3 tok/s \\
Generation-rate p50, concurrency 1 & 341.6 tok/s & 382.9 tok/s \\
Aggregate generation rate, concurrency 8 & 1,059 tok/s & 1,406 tok/s \\
Aggregate generation rate, concurrency 16 & 1,165 tok/s & 2,135 tok/s \\
Usable concurrency under latency/rate gate & $\sim$8 & $\sim$16 \\
Disk footprint & $\sim$63.5GB & 32GB \\
\bottomrule
\end{tabular}
\end{table}

\subsection{Matched BF16 versus FP8 quality}
\label{sec:fp8quality}

Across 32 paired accuracy benchmarks, the mean FP8-minus-BF16 delta is +0.0006, with seventeen
positive and fifteen negative results. Only three absolute deltas exceed 0.02, all on samples of at
most 541 items. IFBench is excluded because empty-response failures differ between the two arms.
Appendix~\ref{app:fp8q} lists every result. These are within-pair quantization deltas; their absolute
scores should not be mixed with the release-evaluation campaign.

\begin{figure}[htbp]
  \centering
  \includegraphics[width=0.86\textwidth]{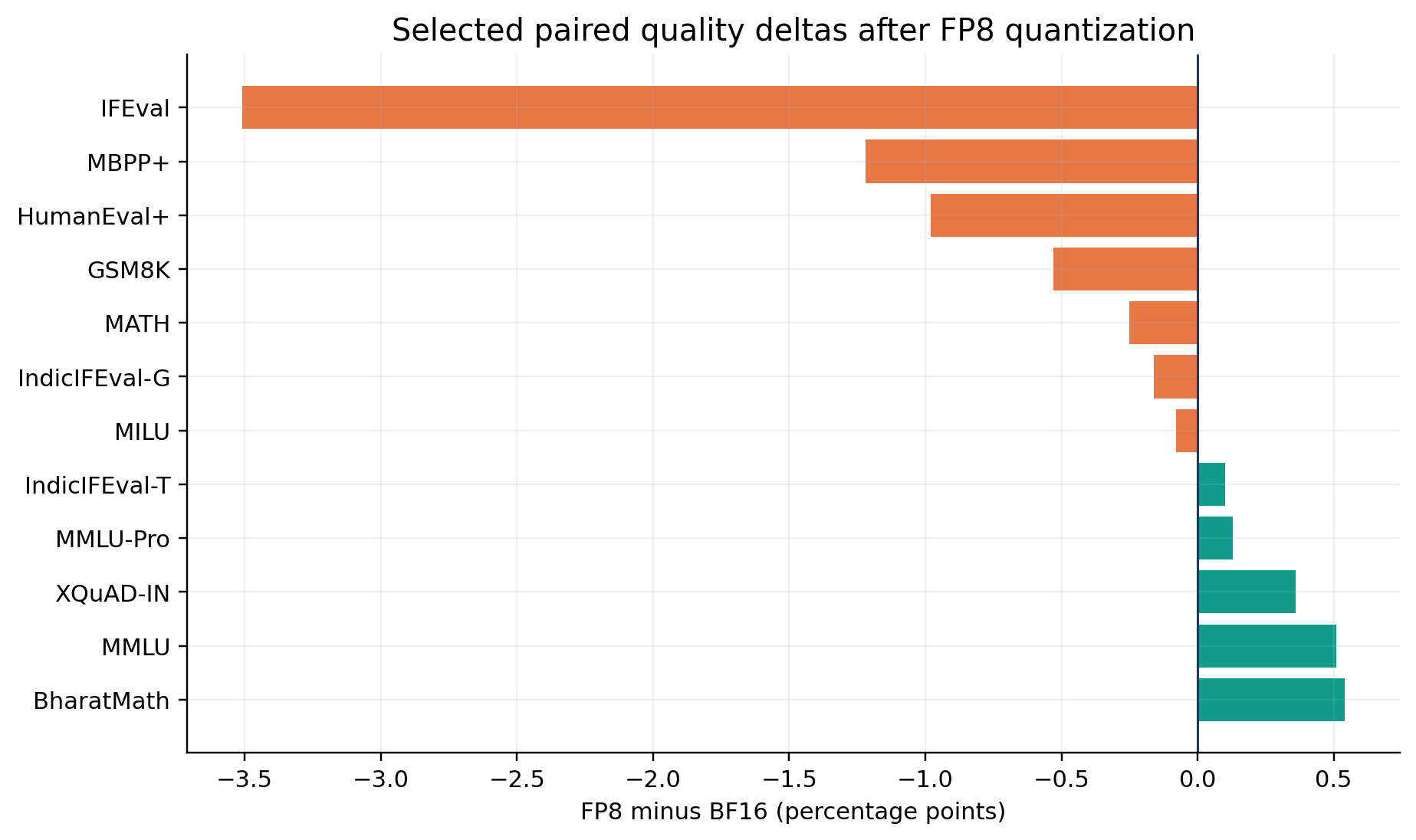}
  \caption{Selected paired accuracy deltas after FP8 PTQ\@. Small benchmark samples should not be over-interpreted.}
  \label{fig:fp8quality}
\end{figure}

FLORES-IN and CrossSum-IN move slightly downward, as do HumanEval+ and MBPP+; Indic MCQ and QA
are otherwise flat to slightly positive. On IFBench, 11.7\% of FP8 and 10.3\% of BF16 responses are
empty, on only partly overlapping items, so the difference mixes instruction following with
non-termination and is not included in the aggregate.

\section{Compute and Infrastructure}

Training uses NVIDIA H200 clusters, Megatron-Bridge for CPT and SFT, and NeMo-RL with NeMo-Gym for
reinforcement learning. Expert and context parallelism distribute sparse routing and long sequences.

Table~\ref{tab:compute} reports GPU-hours for the released lineage. It excludes tests, discarded
runs, retries, ablations, translation inference, and experimental branches. The SFT entry covers
the retained curriculum.

\begin{table}[htbp]
\centering
\caption{Recorded GPU-hours for relevant checkpoints in the released lineage.}
\label{tab:compute}
\begin{tabularx}{0.9\textwidth}{@{}Y R{3.2cm}@{}}
\toprule
\textbf{Released-lineage training scope} & \textbf{Recorded GPU-hours} \\
\midrule
Embedding warmup & 139 \\
CPT broad 974B phase & $\sim$25,147 \\
CPT annealing phase & $\sim$4,988 \\
Complete 128K CPT long-context stage, allocated cluster time & 4,977 \\
Entire retained SFT curriculum & $\sim$17,205 \\
Released \evonrl{} production run & 763 \\
FP8 quantization and calibration & 8 \\
\bottomrule
\end{tabularx}
\end{table}

The 128K ledger uses allocated cluster time, including startup and recovery. The RL value covers the
released step-40 run, and Appendix~\ref{app:throughputdiagnostics} gives detailed timing and MFU.

\section{Responsible Deployment}

\subsection{Intended use}

\shortmodel{} is intended for multilingual chat, long-context retrieval-augmented generation, English and Indic instruction following, mathematical reasoning, coding assistance, and tool-using agents. It can be deployed globally and self-hosted under the release license.

The model is not recommended as an autonomous authority for medical, legal, financial, employment, credit, or other high-stakes decisions. Deployers should evaluate the exact model artifact, quantization, prompt template, languages, and domain used in production.

\subsection{Risks}

The training corpus contains web and synthetic data and can reproduce factual errors, social bias, offensive material, or unsafe instructions. Multilingual safety is especially difficult because a safety behavior learned in one language may not transfer uniformly to another script, dialect, or code-mixed setting.

Reasoning traces can also expose unnecessary internal content, repeat sensitive prompt material, or create a false impression of certainty. Applications should decide explicitly whether reasoning is enabled and should not treat a generated chain of thought as an auditable proof.

\subsection{Recommended deployment controls}

\begin{itemize}
  \item run target-language and target-domain safety evaluations before launch;
  \item apply input/output moderation appropriate to the jurisdiction and application;
  \item verify requested output language and script downstream for strict workflows;
  \item sandbox tools, validate arguments, and require confirmation for consequential actions;
  \item log model version, quantization, template, and decoding settings for incident analysis;
  \item disclose to users when they are interacting with an AI system;
  \item retain human review for high-impact or irreversible decisions.
\end{itemize}

\section{Release and Reproducibility}

The public BF16 weights are hosted at
\url{https://huggingface.co/gnani/gnani-evon-v3.3-30B-A3B} under Apache License 2.0~\cite{evonrelease}. Inherited
components retain their original terms, documented in the release licence inventory. An FP8
derivative is planned; \evonbase{} and intermediate checkpoints are not.

The release bundle records checkpoint and tokenizer identifiers, the chat template, the evaluation
commit, and decoding settings. It also distinguishes report measurements from external results.

\clearpage
\section{Conclusion}

\shortmodel{} treats Indic capability as an end-to-end systems problem. The tokenizer reduces the
cost of Indian scripts; embedding warmup integrates the new vocabulary; continual pretraining adds
language capacity and direct 128K experience; SFT defines the assistant protocol; and the released
RL stage improves reasoning and agentic performance.

The central lesson is that an optimizer cannot reliably preserve behavior that no environment rewards.
The released mixture improves mathematics, structured output, multiple choice, and tool use, while
native-script compliance becomes less reliable. An unreleased Indic-aware branch reverses that trend,
but remains separate from claims about the public model.

The reward-model branch repeats the pattern: optimizing against a learned judge improves knowledge,
code, and reasoning but weakens bare-span answers and English-target translation. The judge itself
also loses 4.6 measured points when it fails to emit a verdict before the generation cap. Distillation
from an earlier checkpoint restores most of the lost conventions without task-specific repair data.

Remaining work includes router-health traces for the shipped expert count, the exact NVIDIA
source-repository identifier, checkpoint-specific SFT token accounting, the FP8 artifact identifier,
and serving measurements beyond one device.

\appendix

\section{Architecture and Artifact Accounting}
\label{app:architectureaccounting}

The released model contains 31,749,975,232 parameters. Under NVIDIA's convention, with six routed
experts active and input embeddings excluded, 3,313,773,760 are active per token. Including the input
embedding gives 3,752,111,296.

The vocabulary extension adds 32,000 rows to each of two untied matrices at model dimension 2,688:

\begin{equation}
  \Delta P_{\mathrm{vocab}}
  = 2 \times 32{,}000 \times 2{,}688
  = 172{,}032{,}000.
\end{equation}

Each MoE layer contains 128 routed experts of width 1,856 and one shared expert of width 3,712.
With a \texttt{relu2} MLP, this gives 258 expert tensors per layer, or 5,934 across 23 layers.
Routers remain separate and in high precision.

Ignoring biases, a routed expert contains 9,977,856 parameters and the shared module contains
19,955,712. The expert weights therefore contribute

\begin{align}
  P_{\mathrm{MoE}}
  &= \left(128\times9{,}977{,}856 + 19{,}955{,}712\right)\times23
   = 29{,}833{,}789{,}440, \\
  P_{\mathrm{MoE,active}}
  &= \left(6\times9{,}977{,}856 + 19{,}955{,}712\right)\times23
   = 1{,}835{,}925{,}504.
\end{align}

The embeddings, output projection, Mamba, attention, normalization, and routing parameters reconcile
these figures with the measured 31.75B total and NVIDIA's 3.31B active-parameter convention.

\section{Exploratory 64K Vocabulary Study}
\label{app:mixv5}

mixv5 is the strongest completed fertility experiment. It uses an approximately 8B-token mixture,
balanced across the eight target languages and spanning native, romanized, and transliterated text.
It is a parallel study and does not alter the production CPT lineage.

The 64K extension reduces mean Samanantar fertility by 40.8\% relative to the upstream tokenizer.
It improves further on the selected 32K tokenizer overall, but Malayalam is slightly worse. The
per-language results are reported in Table~\ref{tab:mixv5fertility}.

\begin{figure}[htbp]
  \centering
  \includegraphics[width=0.94\textwidth]{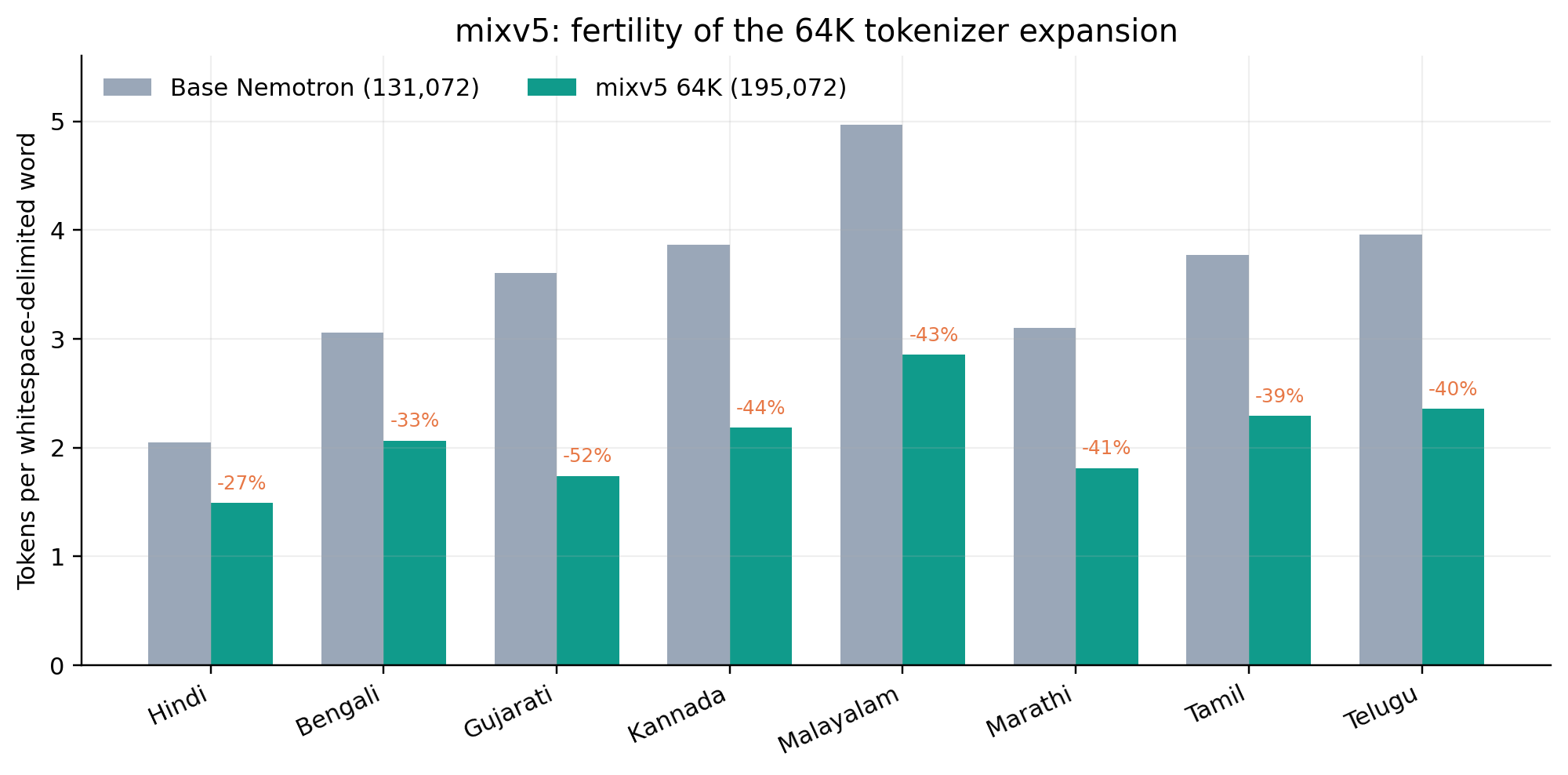}
  \caption{Fertility of the exploratory 64K mixv5 tokenizer against the upstream tokenizer. This tokenizer was not used by the production model.}
  \label{fig:mixv5fertility}
\end{figure}

\begin{table}[htbp]
\centering
\caption{Per-language fertility for mixv5, evaluated on 2,000 Samanantar samples per language with seed 42.}
\label{tab:mixv5fertility}
\begin{tabular}{@{}lrrr@{}}
\toprule
\textbf{Language} & \textbf{Base Nemotron} & \textbf{mixv5 64K} & \textbf{Change} \\
\midrule
Hindi & 2.0513 & 1.4924 & $-27.2\%$ \\
Bengali & 3.0598 & 2.0637 & $-32.6\%$ \\
Gujarati & 3.6086 & 1.7402 & $-51.8\%$ \\
Kannada & 3.8643 & 2.1830 & $-43.5\%$ \\
Malayalam & 4.9654 & 2.8548 & $-42.5\%$ \\
Marathi & 3.0981 & 1.8135 & $-41.5\%$ \\
Tamil & 3.7705 & 2.2906 & $-39.2\%$ \\
Telugu & 3.9595 & 2.3585 & $-40.4\%$ \\
\midrule
Mean & 3.5472 & 2.0996 & $-40.8\%$ \\
\bottomrule
\end{tabular}
\end{table}

The normalization variants are nearly indistinguishable by fertility. The shipped regex variant
therefore reflects segmentation quality, not a measurable sequence-length advantage.

\subsection{Production tokenizer chain of custody}

The production tokenizer descends from AdaptMix Round 09. Of its 32,000 extension entries, 31,999
remain byte-identical in the staged CPT tokenizer. The only intentional exception is ID 144891.

Model preparation removes one degenerate merge and assigns ID 144891 to an unreachable placeholder.
Before SFT export, the slot becomes \texttt{<|file\_sep|>}; only its label changes.

\keypoint{Production therefore uses the 163,072-entry AdaptMix Round-09 lineage. The 195,072-entry mixv5 result is a later parallel study and should never be described as the released tokenizer.}

An independent loader and sampler reproduces the native-script macro fertility within 0.006 of the
reported result. A smaller off-protocol evaluation is retained only as a robustness check.

\section{Training-Time and Throughput Diagnostics}
\label{app:throughputdiagnostics}

Iteration time was logged every tenth optimizer step. Table~\ref{tab:iterationtimebystage} reports
the steady median after step 50 and the full logged range, including startup and recovery costs.

The CPT phase-1 figures combine both segments of that stage. Its step time drifts upward across training, so the median reflects the settled value rather than a stationary rate; the other stages are stationary.

\begin{table}[H]
\centering
\caption{Optimizer-step timing across the production CPT and SFT stages.}
\label{tab:iterationtimebystage}
\begin{tabular}{@{}lrrrr@{}}
\toprule
\textbf{Stage} & \textbf{Logged $n$} & \textbf{Median} & \textbf{p5--p95} & \textbf{Max / steady} \\
\midrule
CPT phase 1 & 5,808 & 6.05 & 5.63--6.18 & 16.73 / 6.05 \\
CPT annealing & 1,161 & 5.84 & 5.81--5.97 & 20.80 / 5.84 \\
CPT long context & 37 & 159.89 & 157.73--162.65 & 173.96 / 159.91 \\
\sftpone{} & 207 & 16.71 & 15.43--18.09 & 52.85 / 16.70 \\
\sftptwo{} & 383 & 17.40 & 16.55--18.31 & 49.99 / 17.39 \\
\evonit{} & 266 & 17.58 & 16.74--18.42 & 60.07 / 17.57 \\
\bottomrule
\end{tabular}
\end{table}

These medians supply the throughput and utilization values in Table~\ref{tab:cptperformance}.

Table~\ref{tab:sftthroughputbystage} reports medians after step 50 from the recorded SFT telemetry.
MFU is relative to the H200 BF16 peak of 989 TFLOP/s per GPU\@.

\begin{table}[H]
\centering
\footnotesize
\caption{Measured steady SFT throughput by retained stage.}
\label{tab:sftthroughputbystage}
\begin{tabular}{@{}lrrrrrr@{}}
\toprule
\textbf{Stage} & \textbf{TFLOP/s/GPU} & \shortstack{\textbf{Aggregate}\\\textbf{TFLOP/s}} & \textbf{Tokens/s} & \shortstack{\textbf{Tokens/s}\\\textbf{per GPU}} & \textbf{Seq./s} & \textbf{MFU} \\
\midrule
\sftpone{} & 321.2 & 82.2k & 1.938M & 7,570 & 14.8 & 32.5\% \\
\sftptwo{} & 308.4 & 79.0k & 1.894M & 7,398 & 14.4 & 31.2\% \\
\evonit{} & 305.2 & 78.1k & 1.870M & 7,304 & 14.3 & 30.9\% \\
\bottomrule
\end{tabular}
\end{table}

\section{CPT Ablations and Non-lineage Validation Runs}
\label{app:cptablations}

The experiments in this section informed configuration and mixture choices but are not checkpoints in the released model's ancestry.

\subsection{Operational validation runs}

A separate 5B-token smoke run validated 128K memory fit, NoPE execution, and cache construction.
Smaller regression runs checked the topology. Neither is included in production quality or compute
totals.

\subsection{The 120B mixing ablation}

\begin{figure}[H]
  \centering
  \includegraphics[width=0.94\textwidth]{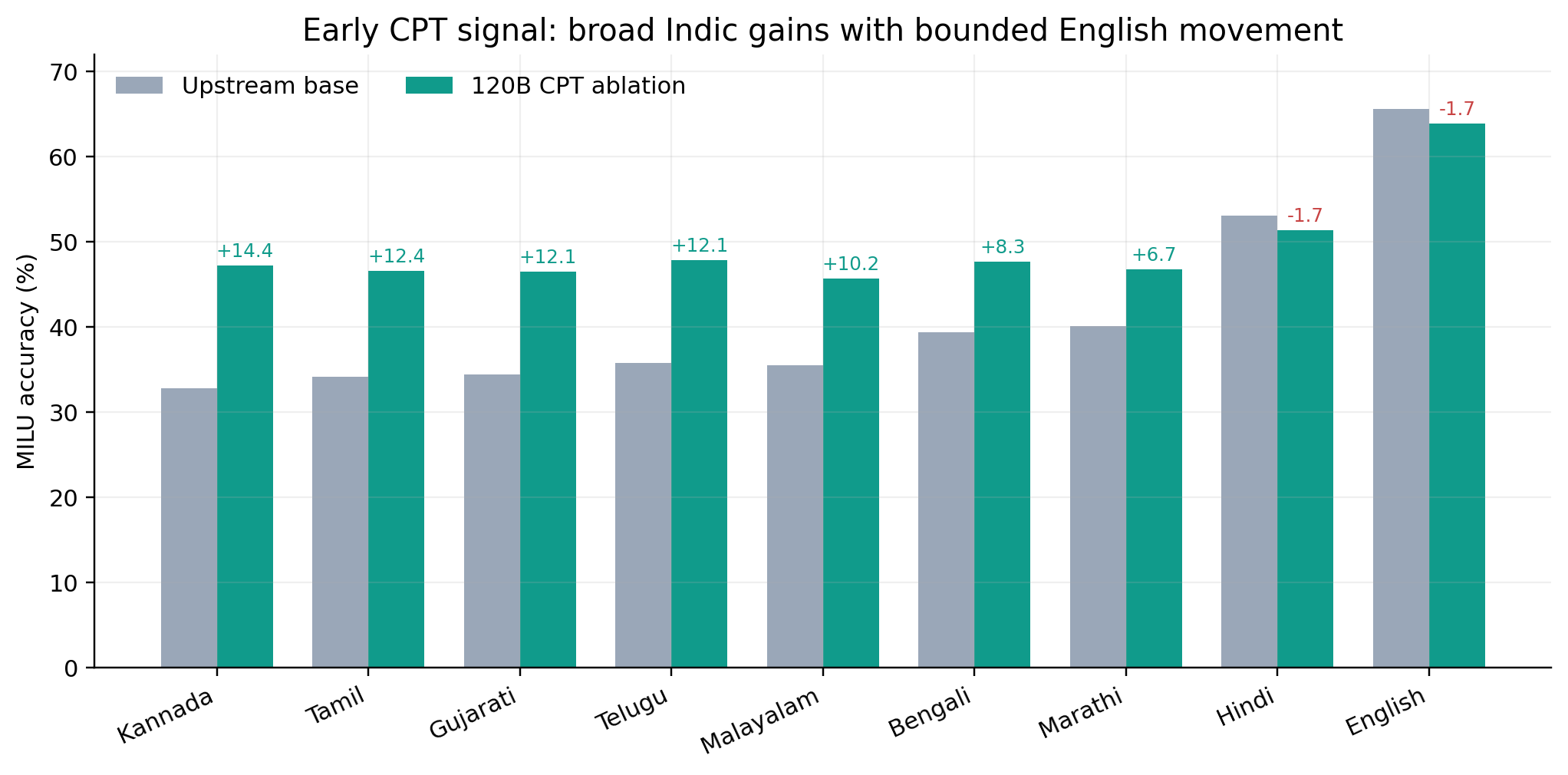}
  \caption{MILU accuracy before and after the 55:45 English--Indic 120B CPT ablation. Labels show percentage-point change. This run is not part of the released checkpoint ancestry.}
  \label{fig:miluablation}
\end{figure}

The 55:45 English--Indic ablation showed that marginal value did not track data abundance. Kannada
gained 14.4 points, while Hindi, the strongest starting language, declined slightly. A second,
more Indic-heavy mixture did not improve the aggregate. Later planning therefore combined measured
uplift with source quality and repetition constraints; Figure~\ref{fig:miluablation} and
Table~\ref{tab:miluablation} give the full language breakdown.

\begin{table}[H]
\centering
\caption{Per-language MILU results from the 120B ablation. Displayed endpoints are rounded to one decimal place; deltas are computed from the underlying higher-precision scores.}
\label{tab:miluablation}
\begin{tabular}{@{}lrrr@{}}
\toprule
\textbf{Language} & \textbf{Base} & \textbf{120B CPT} & \textbf{$\Delta$} \\
\midrule
Kannada & 32.8 & 47.2 & +14.4 \\
Tamil & 34.2 & 46.6 & +12.4 \\
Gujarati & 34.4 & 46.5 & +12.2 \\
Telugu & 35.8 & 47.9 & +12.1 \\
Malayalam & 35.5 & 45.7 & +10.1 \\
Bengali & 39.4 & 47.7 & +8.4 \\
Marathi & 40.1 & 46.8 & +6.7 \\
Hindi & 53.1 & 51.4 & $-1.8$ \\
English & 65.6 & 63.9 & $-1.7$ \\
\bottomrule
\end{tabular}
\end{table}

\subsection{Uplift-weighted planning table}

Table~\ref{tab:cptplanningweights} records the planning weights derived from measured uplift and
high-quality data availability.

\begin{table}[H]
\centering
\caption{Language weights derived from 120B-ablation uplift and high-quality availability. MILU uplifts use the underlying higher-precision scores.}
\label{tab:cptplanningweights}
\begin{tabular}{@{}lrrrr@{}}
\toprule
\textbf{Language} & \textbf{MILU uplift} & \textbf{HQ availability} & \textbf{Weight} & \textbf{Est.\ epochs} \\
\midrule
Telugu & +12.1 & 20B & 16\% & 1.03 \\
Tamil & +12.4 & 12B & 15\% & 1.60 \\
Kannada & +14.4 & 6B & 13\% & 2.78 \\
Gujarati & +12.2 & 8B & 12\% & 1.92 \\
Malayalam & +10.1 & 17B & 12\% & 0.90 \\
Bengali & +8.4 & 19B & 12\% & 0.81 \\
Hindi & $-1.8$ & 35B & 12\% & 0.44 \\
Marathi & +6.7 & 9B & 8\% & 1.14 \\
\bottomrule
\end{tabular}
\end{table}

\section{Detailed CPT Configuration}
\label{app:cptconfig}

\begin{table}[H]
\centering
\small
\caption{Selected broad-CPT configuration fields.}
\label{tab:cptconfig}
\begin{tabularx}{\textwidth}{@{}L{4.7cm}Y@{}}
\toprule
\textbf{Field} & \textbf{Value} \\
\midrule
Initialization & embedding-warmed 163,072-vocabulary checkpoint \\
Precision & BF16 \\
Sequence length & 8,192 for broad CPT and annealing \\
Broad-CPT global / micro batch & 2,048 / 1 \\
Parallelism & TP=1, EP=8, PP=1; data parallelism fills remaining ranks \\
Optimizer & distributed AdamW \\
Peak / minimum learning rate & $1.0\times10^{-5}$ / $1.0\times10^{-6}$ \\
Schedule & Warmup--Stable--Decay with minus-square-root decay \\
Warmup & 200 iterations in the locked production recipe \\
Stable fraction & 0.8 \\
MoE dispatcher & flex + HybridEP \\
Expert capacity & capacity factor 1.25 with pad-to-capacity \\
Kernels & grouped GEMM, selective recomputation, CUDA graph capture for attention and Mamba \\
Tracking & Weights \& Biases; supplied production panels reproduced below \\
\bottomrule
\end{tabularx}
\end{table}

\subsection{Production CPT training traces}
\label{app:cpttraces}

The panels below reproduce the production training series under the public stage names. The sequence
load-balancing curve is the MoE auxiliary objective, not the language-model loss.

\begin{figure}[H]
  \centering
  \includegraphics[width=0.98\textwidth]{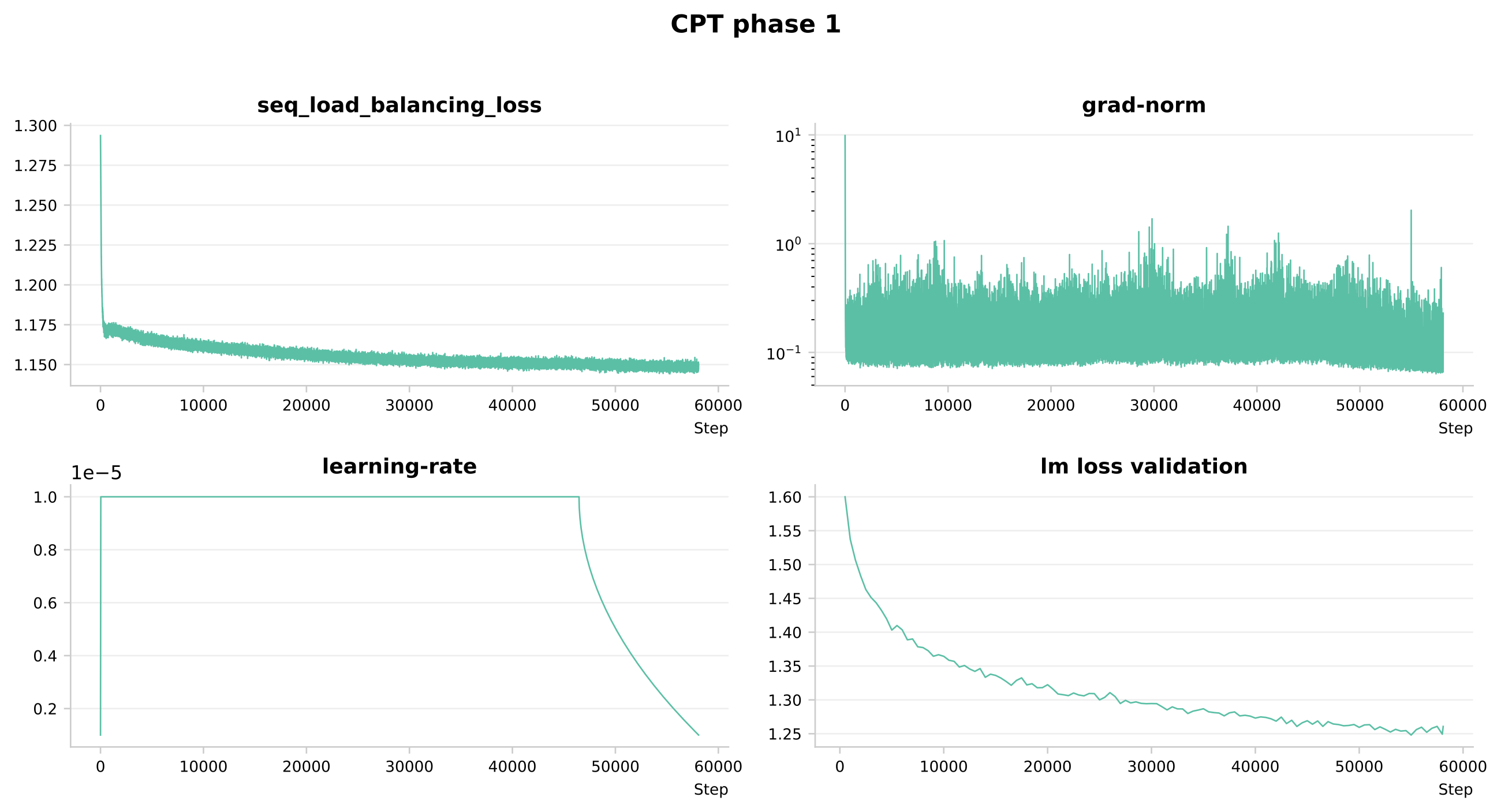}
  \caption{CPT phase-1 training traces through iteration 58,084.}
  \label{fig:cptphase1wandb}
\end{figure}

CPT phase 1 ran in two segments with a 123-iteration overlap. Their language-model losses agree to
$3\times10^{-5}$ in that interval; Figure~\ref{fig:cptphase1wandb} shows the combined trace.

\begin{figure}[H]
  \centering
  \includegraphics[width=0.98\textwidth]{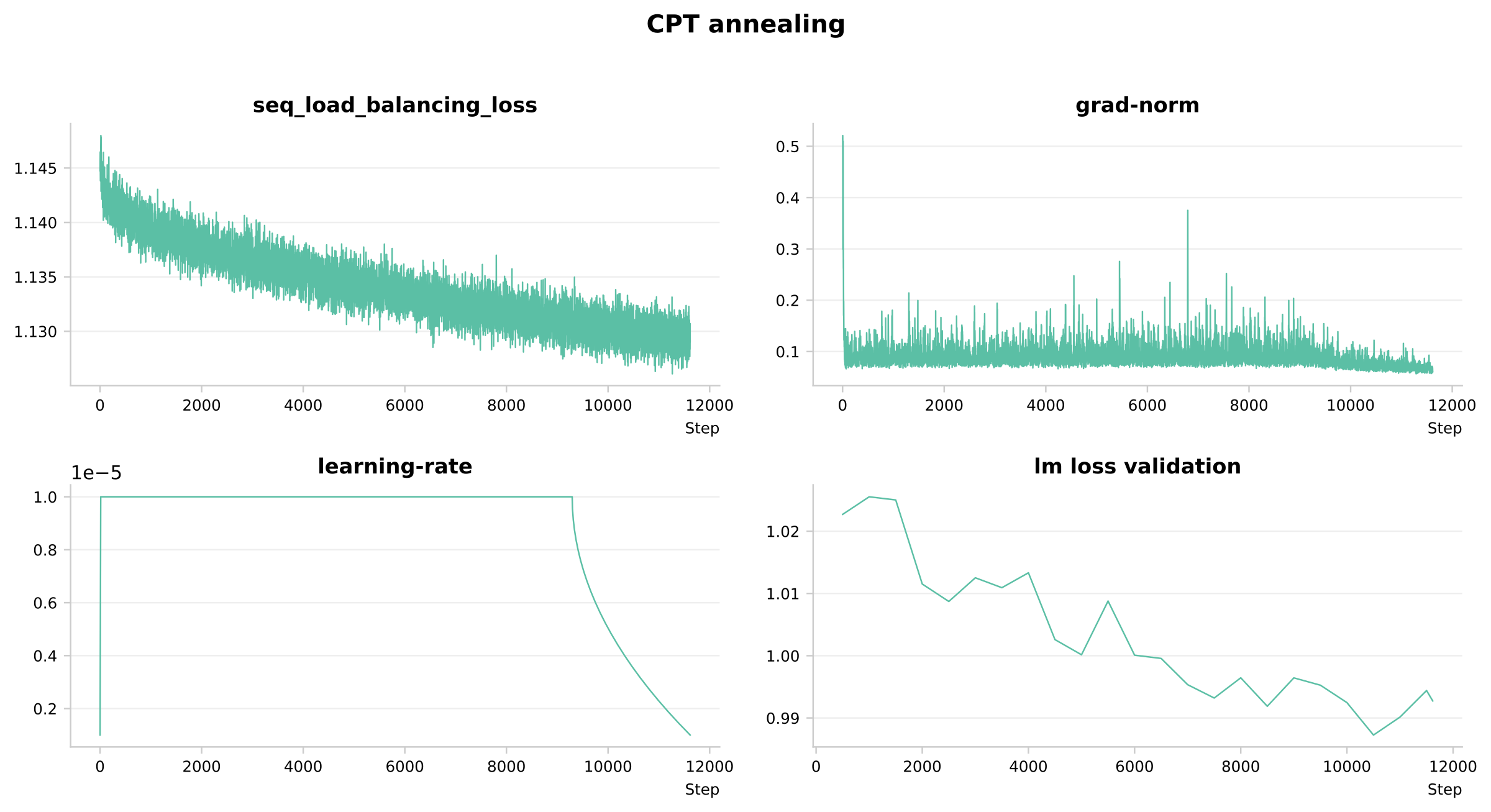}
  \caption{CPT annealing traces. Learning-rate decay begins late while validation loss remains near 1.0.}
  \label{fig:cptannealwandb}
\end{figure}

\begin{figure}[H]
  \centering
  \includegraphics[width=0.98\textwidth]{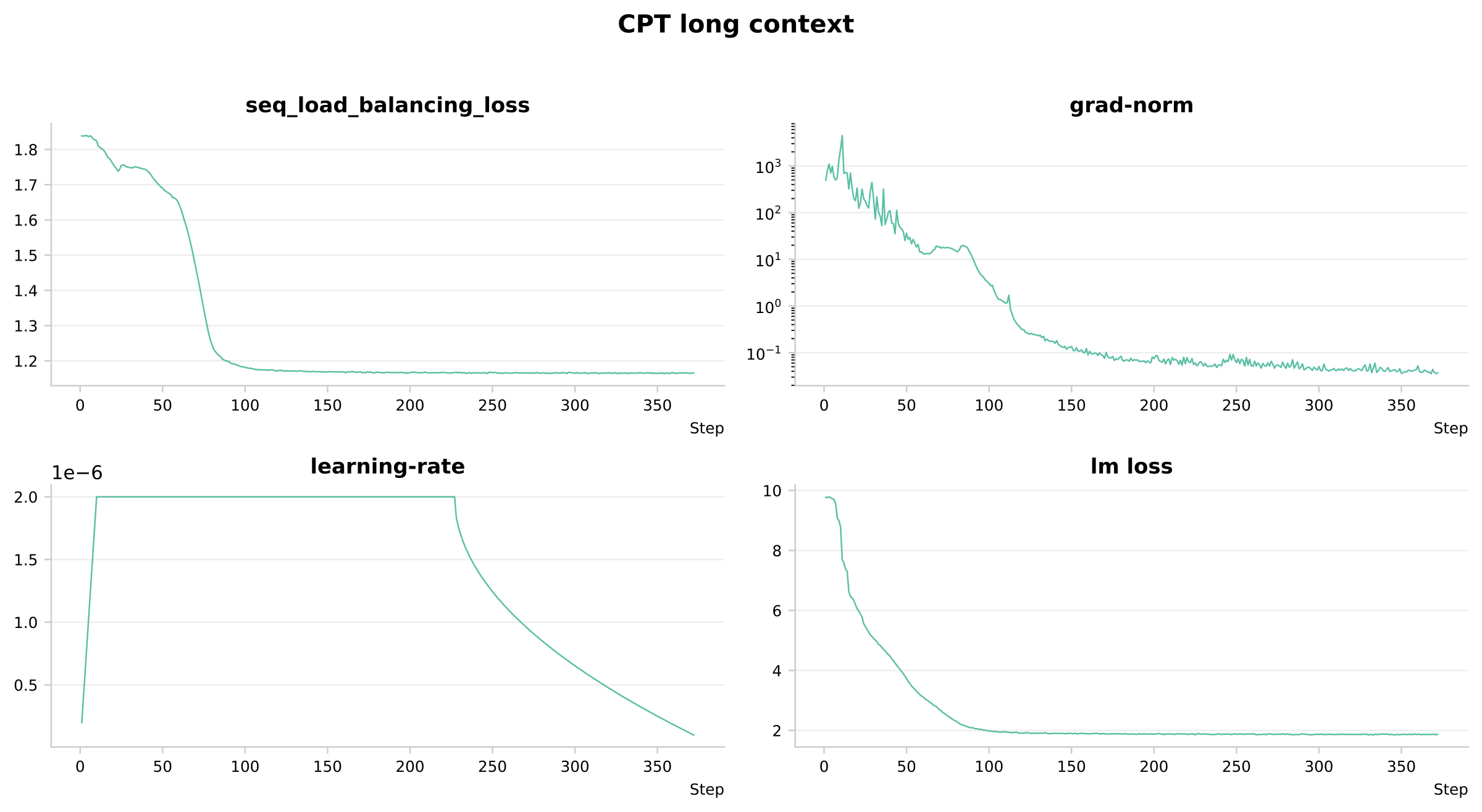}
  \caption{CPT long-context traces through iteration 372; the early transition settles into stable training.}
  \label{fig:cptlongwandb}
\end{figure}

\input{cpt_data_composition.tex}

\section{CPT Checkpoint Evaluation Series}
\label{app:cptevals}

The matched series covers the base and each CPT endpoint. Table~\ref{tab:cptevalendpoints} reports
all benchmarks, with percentage and bits-per-byte metrics kept separate. Deltas compare long context
with the base and should be read within rows; $\ddag$ marks a diagnostic, and $\circ$ an unevaluated slice.

\clearpage
\input{table28_cpt_trajectory.tex}

BharatMath uses precision-corrected accuracy because boxed-answer emission changes across the
training ladder. Phase~1 and long-context accuracy are nearly identical, at 41.06 and 40.95, while
the raw numeric-match score and response length move with answer format. This is the same
capability-versus-convention pattern discussed in Section~\ref{sec:capconv}.

Eight of nine Indic proportion metrics improve from the base to the final long-context checkpoint;
Sanskriti is the exception. The English mean falls 2.67 points, almost entirely during phase~1, with
MMLU-Pro contributing the largest decline. Long-context training adds little further English loss.
HumanEval finishes above its base score, while several other tasks recover during annealing before
easing again.

\subsection{HELMET long-context evaluation}
\label{app:helmet}

HELMET uses base-model completion prompts, greedy decoding, and seed 42; values are on a 0--100
scale. Recall aggregates the harder retrieval slices and should not be compared directly with the
broader RULER mean. Citation and re-ranking are omitted because their structured output format is
not elicited by completion prompting. Indic slices were not evaluated at 128K.

\input{tables/table_helmet.tex}

\section{SFT Corpus Composition Across Retained Stages}
\label{app:sftmix}

The SFT curriculum was rebuilt across three retained stages. This appendix reports materialized
source rows, while Table~\ref{tab:sftselected} separately records packed and scheduled exposure.

\subsection{Selected checkpoint accounting}
\label{app:sftcheckpoint}

The RL initialization is the iteration-2,400 export recorded in the export metadata. At global
batch size 256 and sequence length 131,072, it represents

\begin{equation}
  2{,}400 \times 256 \times 131{,}072
  = 80{,}530{,}636{,}800
\end{equation}

scheduled sequence positions, including padding. A direct mask measurement estimates that 18.2\%
are trainable, or roughly 15B loss-bearing tokens. Table~\ref{tab:sftselected} gives the exact sample
and packing measurements.

The prepared epoch contains 702,824 packed sequences. Sequential bucket boundaries produce 2,752
configured steps, with masked padding in each final partial batch and no wrap-around.

The job reached iteration 2,665; RL uses the iteration-2,400 export. Table~\ref{tab:sftselected}
records its loss, schedule, endpoint, and checkpoint accounting.

\begin{table}[H]
\centering
\caption{Training record for the SFT checkpoint used to initialize RL.}
\label{tab:sftselected}
\small
\renewcommand{\arraystretch}{1.3}
\setlength{\tabcolsep}{12pt}
\begin{tabularx}{\textwidth}{@{}L{7.0cm}Y@{}}
\toprule
\textbf{Setting} & \textbf{Value} \\
\midrule
Public stage & \evonit{} \\
Export source & selected export, verified from export metadata \\
Scheduled sequence positions through export & 80.53B, including padding \\
Approximate loss-bearing positions & $\sim$15B (18.2\%; assistant-only loss mask) \\
Trainable-rate measurement & 95,515 / 524,288 positions over four packed-parquet rows \\
Intra-pack efficiency & 99.98\% \\
Prepared packed-sequence capacity & 92.12B positions across 702,824 sequences \\
Fully scheduled capacity & 92.34B positions across 2,752 global batches \\
Packed rows / scheduled documents & 702,824 / 15,160,245 \\
Iterations: configured / recorded / RL export & 2,752 / 2,665 / 2,400 \\
Last job-log line / last checkpoint written & 2,660 / 2,600 \\
Epoch traversal & sequential bucket blocks; no repeat or wrap-around \\
Learning-rate schedule & cosine, $5\times10^{-7}$ peak to $1\times10^{-7}$ minimum \\
Optimizer & Adam, $\beta_2=0.95$, $\epsilon=10^{-8}$, weight decay 0.1 \\
Materialized-row language balance & 64.0\% English and 36.0\% Indic \\
RL-init loss & 0.614 \\
\bottomrule
\end{tabularx}
\end{table}

\begin{table}[H]
\centering
\small
\caption{Stage totals for the retained SFT corpus chain.}
\label{tab:sftstagetotals}
\begin{tabularx}{\textwidth}{@{}L{3.3cm}R{2.5cm}R{1.5cm}R{1.5cm}Y@{}}
\toprule
\textbf{Stage} & \textbf{Materialized rows} & \textbf{Buckets} & \textbf{English} & \textbf{Key character} \\
\midrule
\sftpone{} & $\sim$12.5M & 12 & 50.7\% & Era-1 phase-1 pool; instruction following, mathematics, and code dominate. \\
\sftptwo{} & $\sim$14.7M & 14 & 54.6\% & Era-1 phase-2 pool; code, science, and agentic shares increase. \\
\sftrthree{} & $\sim$15.2M & 15 & 64.0\% & Era-2 rebuild with a larger English share. \\
\bottomrule
\end{tabularx}
\end{table}

\clearpage
\begin{table}[H]
\centering
\caption{Materialized-row language mix across the retained SFT chain.}
\label{tab:sftlanguagemix}
\begin{tabular}{@{}lrrr@{}}
\toprule
\textbf{Language} & \textbf{\sftpone{}} & \textbf{\sftptwo{}} & \textbf{\sftrthree{}} \\
\midrule
English & 50.7\% & 54.6\% & 64.0\% \\
Hindi & 16.5\% & 15.2\% & 11.1\% \\
Tamil & 6.5\% & 6.0\% & 4.8\% \\
Telugu & 6.5\% & 6.0\% & 4.8\% \\
Kannada & 6.5\% & 6.0\% & 4.8\% \\
Malayalam & 6.5\% & 6.0\% & 4.8\% \\
Bengali & 3.4\% & 3.1\% & 2.7\% \\
Gujarati & 1.7\% & 1.6\% & 1.5\% \\
Marathi & 1.7\% & 1.6\% & 1.5\% \\
\bottomrule
\end{tabular}
\end{table}

\subsection{Source-family evolution}

Table~\ref{tab:sftpersource} groups the SFT sources into broad families and reports approximate
within-stage shares of materialized rows, rounded to the nearest percentage point. English sources are
drawn primarily from NVIDIA's publicly released Nemotron post-training collections~\cite{nemotron3},
alongside Gnani-built Indic dialogue and agentic-trace data. Indic rows in each family include
structure-aware translations of the corresponding English sources (Section~\ref{sec:sftdata}).

\begin{table}[H]
\centering
\small
\caption{Approximate source-family composition across the retained SFT corpus chain (share of materialized rows).}
\label{tab:sftpersource}
\begin{tabular}{@{}lrrr@{}}
\toprule
\textbf{Source family} & \textbf{\sftpone{}} & \textbf{\sftptwo{}} & \textbf{\sftrthree{}} \\
\midrule
Mathematics (incl.\ proofs) & $\sim$29\% & $\sim$28\% & $\sim$15\% \\
Code (incl.\ competitive programming) & $\sim$24\% & $\sim$29\% & $\sim$13\% \\
Instruction following and chat & $\sim$28\% & $\sim$15\% & $\sim$19\% \\
Science and reasoning & $\sim$17\% & $\sim$19\% & $\sim$9\% \\
Agentic, tool use, and software engineering & $\sim$2\% & $\sim$9\% & $\sim$12\% \\
Indic dialogue & -- & -- & $\sim$8\% \\
Safety and domain-specialized & $<$1\% & $\sim$1\% & $\sim$1\% \\
Other curriculum buckets & -- & -- & $\sim$23\% \\
\bottomrule
\end{tabular}
\end{table}

The final \sftrthree{} build is 64.0\% English and 36.0\% Indic at the materialized-row level. These are not
token-level sampling or assistant-loss shares, which may differ from the row distribution.

\section{SFT Training Traces}
\label{app:sfttraces}

The \sftpone{}, \sftptwo{}, and \sftrthree{} panels below show the retained SFT training stages. Step-local changes in language-model loss reflect changes in curriculum buckets as well as optimization; they should not be interpreted as a single stationary loss curve across the three stages.

\begin{figure}[H]
  \centering
  \includegraphics[width=0.98\textwidth]{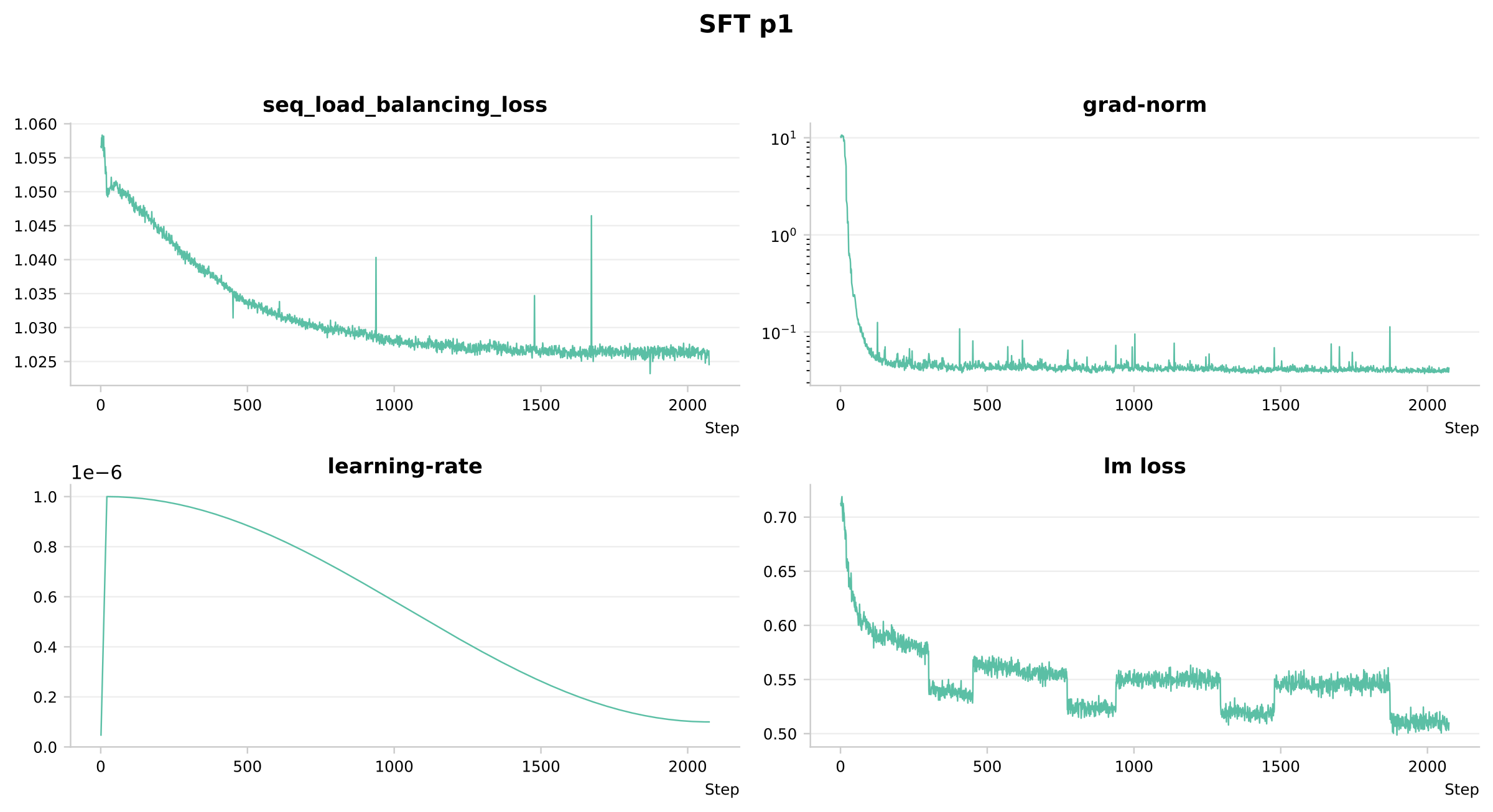}
  \caption{Weights \& Biases traces for \sftpone{}. The panels report the MoE sequence load-balancing loss, gradient norm, cosine learning-rate schedule, and assistant-masked language-model loss. Loss plateaus and transitions align with changes in the sequential curriculum buckets.}
  \label{fig:sftp1wandb}
\end{figure}

\begin{figure}[H]
  \centering
  \includegraphics[width=0.98\textwidth]{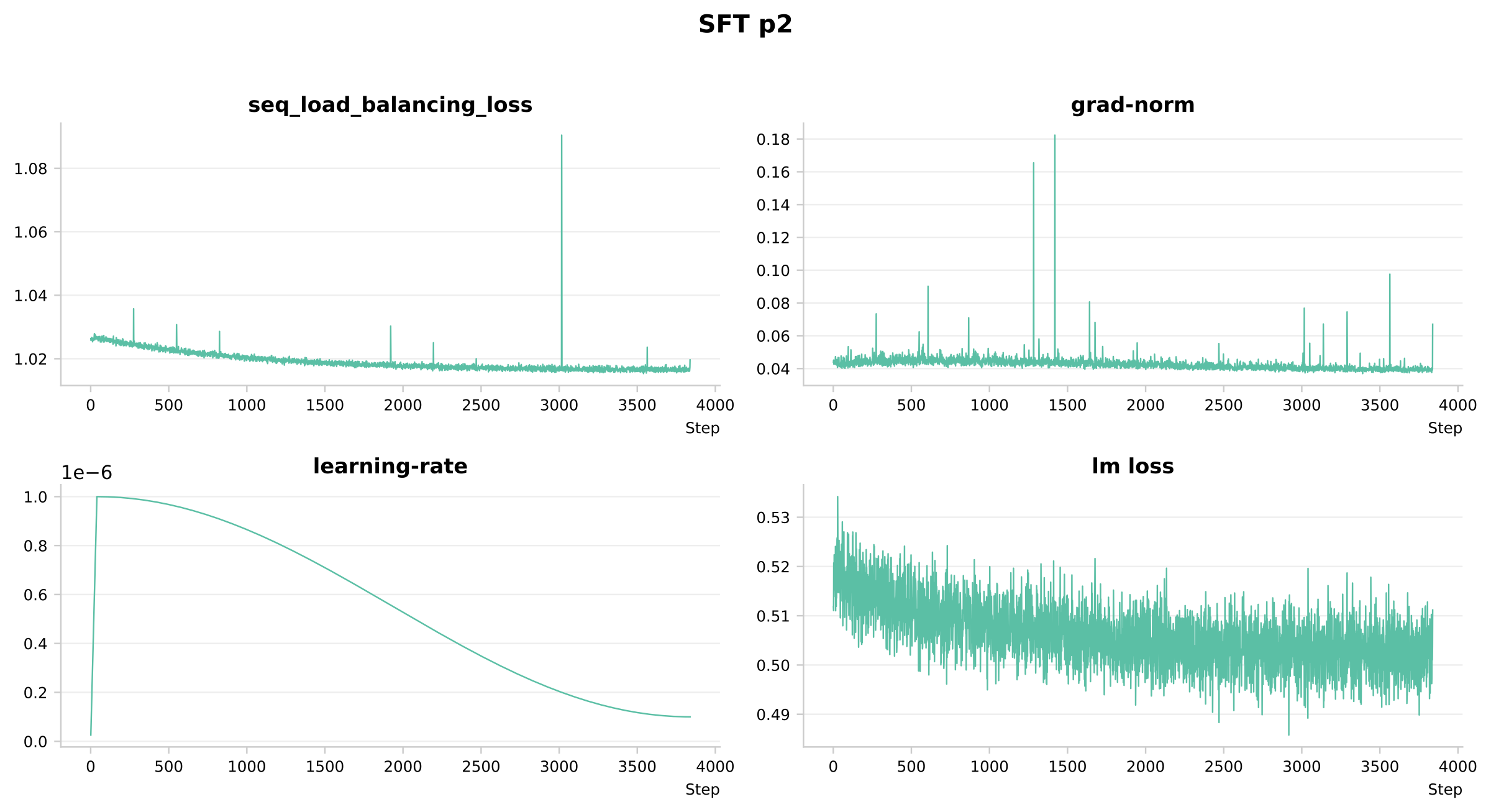}
  \caption{Weights \& Biases traces for \sftptwo{}. The assistant-masked loss trends gradually downward under the cosine learning-rate schedule, while isolated gradient-norm and load-balancing spikes remain transient.}
  \label{fig:sftp2wandb}
\end{figure}

\begin{figure}[H]
  \centering
  \includegraphics[width=0.98\textwidth]{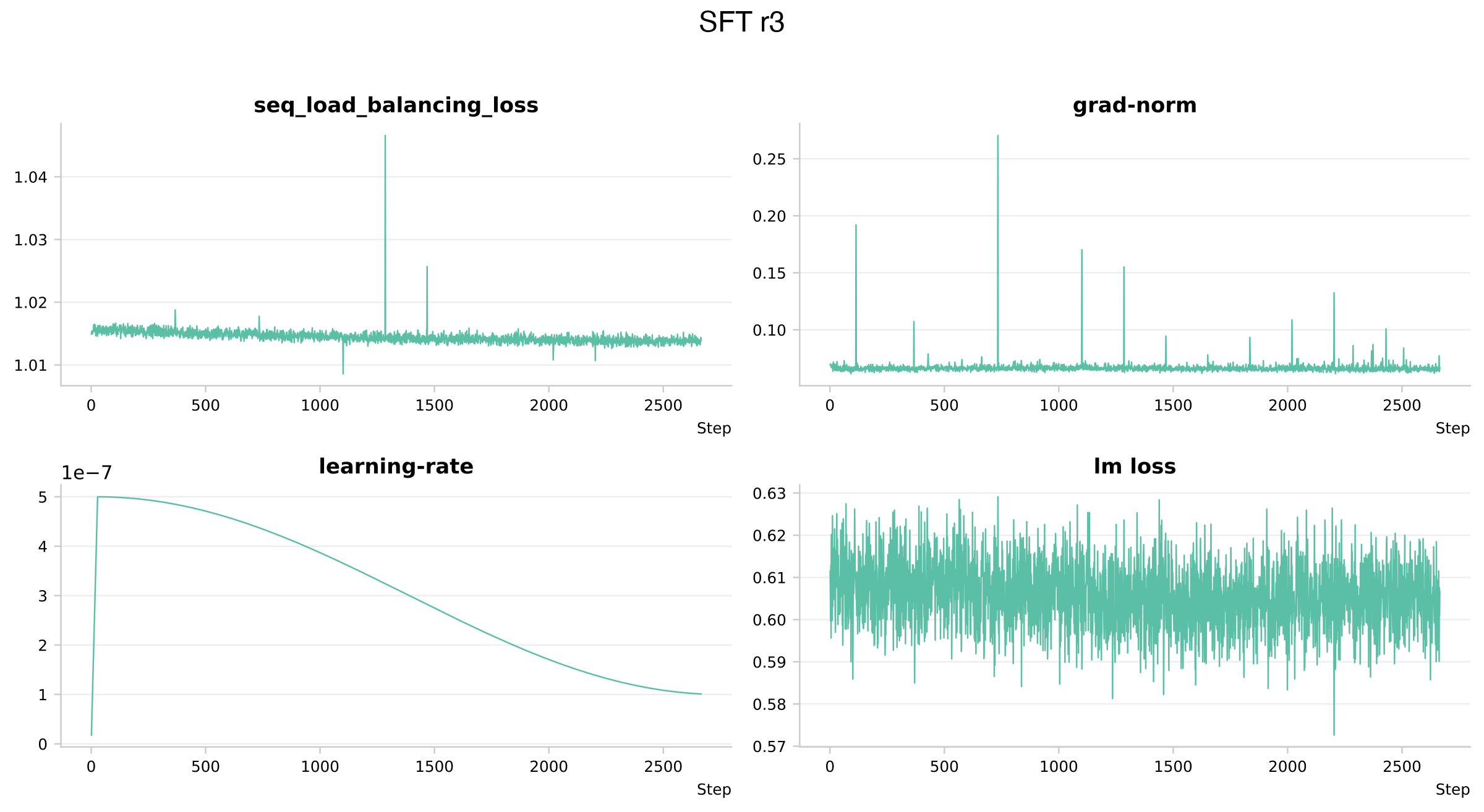}
  \caption{Final \sftrthree{} SFT traces through iteration 2,665; \evonit{} lies within this run.}
  \label{fig:sftr3wandb}
\end{figure}

\clearpage
\section{Matched Evaluation of the Released Checkpoint}
\label{app:matched}

\input{tables/table18_matched.tex}

\paragraph{MATH emission decomposition.}
The main text partitions responses after reasoning is removed. Partitioning the raw response instead
finds 150 additional boxed SFT outputs without changing the 2,516 correct boxed answers. The resulting
denominators describe different conventions and must not be mixed.

\section{Indic Extractive-QA Failure Taxonomy}
\label{app:qataxonomy}

\input{appendix_qa_taxonomy.tex}

\section{Sibling Indic-Aware RL Study (Not Released)}
\label{app:phase2rl}

\input{appendix_phase2.tex}

\section{Reinforcement-Learning Configuration, Side by Side}
\label{app:rlconfig}

This appendix sets the released run beside the Indic-aware branch of
Section~\ref{sec:branches}. The two share an initialization, an optimizer, a length budget, and every
KL and clipping setting. They differ in how prompts are batched, and in what the reward mixture
contains.

\begin{table}[htbp]
\centering\small
\caption{Training configuration for the released run and the Indic-aware branch, both read at
step~40. ``Same'' means the two runs carry an identical setting.}
\label{tab:rlconfigcompare}
\begin{tabular}{@{}L{6.0cm}L{4.6cm}L{3.4cm}@{}}
\toprule
\textbf{Setting} & \textbf{\evonrl{} (released)} & \textbf{\evonrlindic{} (branch)} \\
\midrule
\multicolumn{3}{@{}l}{\textit{Optimizer and learning rate}} \\
Optimizer & Adam, distributed, BF16, weight decay 0.01 & Same \\
Peak learning rate & $5\times10^{-6}$ & Same \\
Minimum learning rate & $5\times10^{-7}$ & Same \\
Schedule & constant, 13-iteration warmup from the floor & Same \\
Gradient clipping & 1.0 & Same \\
Learning rate logged at step 40 & $5\times10^{-6}$ & $5\times10^{-6}$ \\
\addlinespace[3pt]
\multicolumn{3}{@{}l}{\textit{Batching}} \\
Prompts per step & 128 & 255 \\
Generations per prompt & 16 & 8 \\
Nominal rollouts per step & 2,048 & 2,040 \\
Training global batch & 2,048 & 2,040 \\
Training micro-batch & 4 & 4 \\
Generation batch & 32 & 32 \\
Retained samples at step 40 & 2,048, with 238 discarded & 2,040, with 610 discarded \\
Dynamic sampling & on, up to 10 extra generation batches & Same \\
Cluster in the saved configuration & 32 nodes, 8 GPUs each & 30 nodes, 8 GPUs each \\
\addlinespace[3pt]
\multicolumn{3}{@{}l}{\textit{KL, clipping, and advantage}} \\
PPO ratio clip & 0.2 either side & Same \\
KL penalty in the loss & 0 & Same \\
KL penalty in the reward & off & Same \\
KL estimator & $k_3$, input clamp 20, output clamp 10 & Same \\
Advantage & GRPO, leave-one-out baseline, normalized & Same \\
KL penalty logged at step 40 & 0.0 & 0.0 \\
\addlinespace[3pt]
\multicolumn{3}{@{}l}{\textit{Lengths}} \\
Maximum total sequence & 49,152 & Same \\
Maximum new tokens per generation & 16,384 & Same \\
Rollout-engine context limit & 49,152 & Same \\
Sequence-packing tile & 98,304 tokens & Same \\
Generated tokens at step 40, median & 2,527 & 2,220 \\
Generated tokens at step 40, 95th percentile & 10,276 & 11,781 \\
\bottomrule
\end{tabular}
\end{table}

The saved branch configuration names 30 nodes, while the run itself used 16. It also carries three
corrections absent from the released run. First, it clamps the advantage-normalization divisor at
$\max(\sigma,0.05)+10^{-6}$. Without this floor, tiny reward differences can become full-sized
updates.

Second, multi-turn rows are reduced from 37\% to 11\% of the Indic instruction pool. Third, empty
content is treated as malformed rather than merely unrewarded.

\begin{figure}[htbp]
  \centering
  \includegraphics[width=0.94\textwidth]{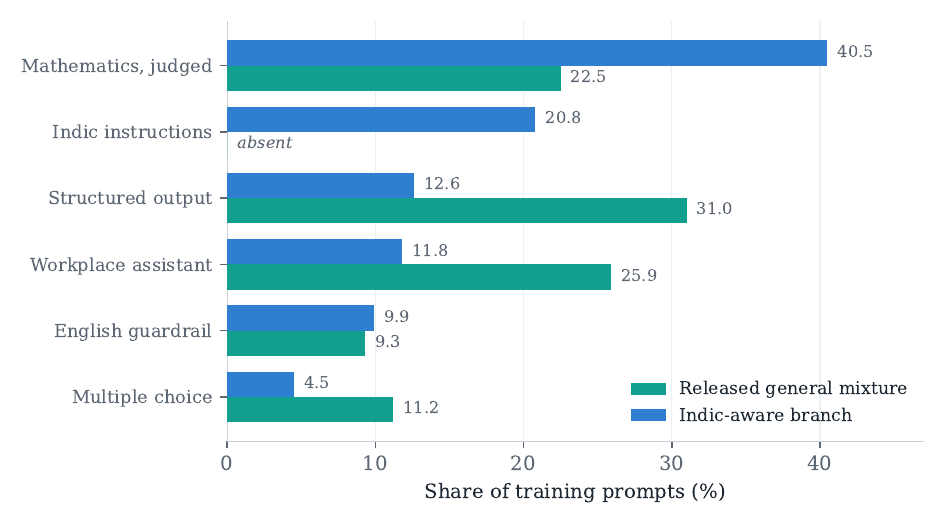}
  \caption{Nominal environment shares before dynamic sampling for the released and Indic-aware runs.}
  \label{fig:rlenvmix}
\end{figure}

\section{Reward Model and Distillation Study (Not Released)}
\label{app:judgeopd}

\input{appendix_judge_opd.tex}

\section{Matched BF16 and FP8 Benchmark Results}
\label{app:fp8q}

\begin{table}[htbp]
\centering\small
\caption{All 32 paired BF16 and FP8 accuracy results. Deltas are FP8 minus BF16.}
\label{tab:fp8qfull}
\begin{tabular}{@{}lrrr@{}}
\toprule
\textbf{Benchmark} & \textbf{BF16} & \textbf{FP8} & \textbf{$\Delta$} \\
\midrule
AGIEval English & 0.7347 & 0.7440 & $+0.0093$ \\
AIME 2025 & 0.7438 & 0.7500 & $+0.0062$ \\
AIME 2026 & 0.7458 & 0.7729 & $+0.0271$ \\
ARC-Challenge & 0.9590 & 0.9582 & $-0.0008$ \\
BBQ & 0.8869 & 0.8819 & $-0.0050$ \\
Beyond-AIME & 0.5356 & 0.5337 & $-0.0019$ \\
BharatMath & 0.7317 & 0.7370 & $+0.0053$ \\
BhashaBench-Multi & 0.4922 & 0.4916 & $-0.0006$ \\
GPQA-Diamond & 0.6503 & 0.6566 & $+0.0063$ \\
GSM8K & 0.9431 & 0.9378 & $-0.0053$ \\
HMMT February & 0.6083 & 0.6438 & $+0.0355$ \\
HMMT November & 0.6333 & 0.6354 & $+0.0021$ \\
HumanEval+ & 0.8854 & 0.8756 & $-0.0098$ \\
IFEval & 0.8281 & 0.7930 & $-0.0351$ \\
IFEval-Hindi & 0.5059 & 0.4965 & $-0.0094$ \\
IndicIFEval grounded & 0.3305 & 0.3290 & $-0.0015$ \\
IndicIFEval translated & 0.4987 & 0.4997 & $+0.0010$ \\
IndicMMLU-Pro & 0.6888 & 0.6882 & $-0.0006$ \\
IndicParam & 0.3928 & 0.3939 & $+0.0011$ \\
IndicQA & 0.1222 & 0.1320 & $+0.0098$ \\
LiveCodeBench v6 & 0.5774 & 0.5793 & $+0.0019$ \\
MATH & 0.8476 & 0.8451 & $-0.0025$ \\
MATH-500 & 0.8400 & 0.8380 & $-0.0020$ \\
MBPP+ & 0.7878 & 0.7757 & $-0.0121$ \\
MGSM-English & 0.9840 & 0.9680 & $-0.0160$ \\
MGSM-Indic & 0.8460 & 0.8520 & $+0.0060$ \\
MILU & 0.7833 & 0.7825 & $-0.0008$ \\
MMLU & 0.8668 & 0.8719 & $+0.0051$ \\
MMLU-Pro & 0.7726 & 0.7739 & $+0.0013$ \\
XorQA-Indic & 0.3524 & 0.3528 & $+0.0004$ \\
XQuAD-English & 0.6370 & 0.6378 & $+0.0008$ \\
XQuAD-Indic & 0.1286 & 0.1321 & $+0.0035$ \\
\bottomrule
\end{tabular}
\end{table}

\clearpage
\section{Glossary}

\begin{description}
  \item[BPE] Byte-pair encoding, an iterative method that learns frequently occurring token merges.
  \item[CPT] Continual pretraining, further next-token training of an existing base model.
  \item[CP] Context parallelism, which partitions a long sequence across devices.
  \item[EP] Expert parallelism, which partitions MoE experts across devices.
  \item[Fertility] Average tokenizer tokens produced per whitespace-delimited word.
  \item[Constituent-mean initialization] Initialization of a new token row from the old-token pieces that exactly compose it.
  \item[GQA] Grouped-query attention, in which multiple query heads share fewer key/value heads.
  \item[GRPO] Group relative policy optimization, which estimates relative advantages from several completions of the same prompt.
  \item[In-script rate] Fraction of outputs written in the target writing system.
  \item[MoE] Mixture of experts, a sparse layer that routes each token through a subset of expert networks.
  \item[NoPE] No explicit positional embedding.
  \item[SFT] Supervised fine-tuning on prompt/response or conversational demonstrations.
  \item[TP] Tensor parallelism, which partitions individual tensors and matrix operations across devices.
  \item[WSD] Warmup--Stable--Decay learning-rate schedule.
\end{description}

\end{document}

%% file: cpt_data_composition.tex
\subsection{CPT data composition}
\label{sec:cptdatacomposition}

The three continual-pretraining stages draw on separate blend manifests. Broad CPT uses a 974B
English-majority mixture and annealing a 196B high-quality mixture, while the long-context stage
uses the pool already reported in Table~\ref{tab:longctxpool}. The broad-CPT manifest retains 184
shard prefixes; 90 shards from its source manifest were dropped for absent per-shard caches and
the remaining weights renormalized to unity.

Two share conventions appear in the manifest audit block and are easy to conflate. A
\emph{target} share is a fraction within the English or Indic budget; a \emph{sampling} share is
a fraction of the whole mix. The same code allocation reads 53.2\% one way and 29.1\% the other.
The figures below are realized sampling shares, computed as weight $\times$ 974.5B from the
per-source accounting, and are therefore comparable across the whole table. Blend weights are sampling probabilities rather than guarantees: realized per-source consumption also depends on shard exhaustion, and several sources in the annealing manifest are scheduled for less than one epoch.

\begin{table}[htbp]
\centering\small
\caption{Broad-CPT composition from the 184 retained shard weights.}
\label{tab:cptbroadsplit}
\begin{tabular}{lrrrr}
\toprule
Group & Target & Realized & Share & $\Delta$ \\
\midrule
English              & 536.0B & 532.7B & 54.66\% & $-3.3$B \\
Indic (12 languages) & 438.5B & 436.6B & 44.80\% & $-1.9$B \\
Other multilingual   & --     &   5.3B &  0.54\% & --      \\
\midrule
Total                & 974.5B & 974.5B & 100\%   &         \\
\bottomrule
\end{tabular}
\end{table}

\begin{table}[htbp]
\centering\small
\caption{Realized Indic allocation in broad CPT, by language. Only the first eight languages
received dedicated vocabulary expansion (Section~\ref{sec:tokenizer}); Odia, Punjabi, Urdu and
Nepali receive pretraining data on a small budget without tokenizer support.}
\label{tab:cptbroadindic}
\begin{tabular}{lrr@{\hspace{2em}}lrr}
\toprule
Language & Tokens & Share & Language & Tokens & Share \\
\midrule
Hindi     & 132.4B & 13.59\% & Marathi  & 17.1B & 1.76\% \\
Tamil     &  69.3B &  7.11\% & Odia     &  4.4B & 0.45\% \\
Telugu    &  66.6B &  6.84\% & Punjabi  &  1.6B & 0.17\% \\
Bengali   &  53.8B &  5.52\% & Urdu     &  1.4B & 0.14\% \\
Malayalam &  33.5B &  3.43\% & Nepali   &  1.3B & 0.14\% \\
Kannada   &  28.7B &  2.95\% &          &       &        \\
Gujarati  &  26.3B &  2.70\% &          &       &        \\
\bottomrule
\end{tabular}
\end{table}

The largest shard is English code at 12.21\% of the mix, followed by synthetic code question
answering at 6.90\%. No Indic shard exceeds 2.5\%.

\begin{table}[htbp]
\centering\small
\caption{Annealing composition by bucket from the manifest's per-dataset allocations.}
\label{tab:cptannealbuckets}
\begin{tabular}{lrr}
\toprule
Bucket & Allocated & Share \\
\midrule
Reasoning            & 62.70B & 32.2\% \\
Knowledge retention  & 34.20B & 17.5\% \\
Mathematics and STEM & 31.52B & 16.2\% \\
Code                 & 25.65B & 13.2\% \\
Indic high-quality   & 17.10B &  8.8\% \\
Instruction          & 10.10B &  5.2\% \\
Multilingual         &  5.70B &  2.9\% \\
Web                  &  5.70B &  2.9\% \\
\bottomrule
\end{tabular}
\end{table}

The listed allocations are 1.14\% below the stage budget because two dropped shards were
redistributed across the survivors. Annealing contains 11.7\% Indic data, down from 44.8\% in broad
CPT\@; its long-document complement supplied the long-context stage.

That reduction should not be read as the cause of the further MMLU decline recorded for this
stage in Table~\ref{tab:cptevalendpoints}. The English majority in annealing is skill-shaped
rather than knowledge-shaped: reasoning, mathematics and code together account for 61.5\% of the
stage against 17.5\% of explicit knowledge retention. A stage that is 88\% English can still cost
English knowledge when little of that English is knowledge-dense, which is the more plausible
mechanism for the observed change.

%% file: table28_cpt_trajectory.tex
\begin{table}[H]
\centering\small
\caption{Per-benchmark matched trajectory across the three CPT stages.}
\label{tab:cptevalendpoints}
\begin{tabular}{@{}llrrrrr@{}}
\toprule
\textbf{Benchmark} & \textbf{Metric} & \textbf{Base} & \textbf{Phase 1} & \textbf{Annealing} & \textbf{Long ctx.} & \textbf{$\Delta$} \\
\midrule
\multicolumn{7}{@{}l}{\textit{Indic capability}} \\
Indic macro (excl.\ BharatMath) & accuracy & 41.09 & 46.83 & 46.63 & 46.83 & $+5.74$ \\
Indic macro (incl.\ BharatMath) & accuracy & 39.54 & 46.03 & 45.66 & 46.01 & $+6.47$ \\
ARC-Challenge-Indic & accuracy & 35.23 & 48.20 & 48.20 & 47.09 & $+11.86$ \\
Belebele & accuracy & 37.49 & 42.17 & 42.27 & 42.54 & $+5.05$ \\
BharatMath & accuracy & 29.99 & 41.06 & 39.70 & 40.95 & $+10.96$ \\
Global-MMLU-Lite & accuracy & 32.50 & 36.17 & 36.00 & 35.75 & $+3.25$ \\
IndicGLUE-selected & score & 52.12 & 57.86 & 57.22 & 57.97 & $+5.85$ \\
IndicMMLU-Pro & accuracy & 14.44 & 20.03 & 20.02 & 19.44 & $+5.00$ \\
IndicParam & accuracy & 28.08 & 31.59 & 31.55 & 31.58 & $+3.50$ \\
MILU & accuracy & 40.63 & 53.54 & 53.52 & 53.24 & $+12.61$ \\
Sanskriti & accuracy & 84.15 & 80.60 & 80.82 & 80.85 & $-3.30$ \\
FLORES-IN BPB, uncond. $\downarrow$ & bits/byte & 1.0945 & 0.8272 & 0.8140 & 0.8172 & $-0.2773$ \\
FLORES-IN BPB, cond. $\downarrow$ & bits/byte & 0.5962 & 0.4107 & 0.4124 & 0.4181 & $-0.1781$ \\
Indic BPB overall $\downarrow$ & bits/byte & 0.8453 & 0.6189 & 0.6132 & 0.6177 & $-0.2276$ \\
\addlinespace
\multicolumn{7}{@{}l}{\textit{English retention}} \\
Overall, nine-bench mean & mean & 70.62 & 67.98 & 68.40 & 67.95 & $-2.67$ \\
ARC-Challenge & likelihood acc & 67.66 & 67.49 & 67.92 & 67.15 & $-0.51$ \\
GSM8K, strict & exact match & 86.13 & 84.46 & 84.46 & 84.00 & $-2.13$ \\
HellaSwag & likelihood acc & 85.57 & 83.52 & 83.69 & 83.69 & $-1.88$ \\
MMLU & likelihood acc & 78.50 & 75.20 & 75.11 & 75.28 & $-3.22$ \\
MMLU-Pro & CoT exact match & 64.83 & 54.78 & 55.43 & 54.76 & $-10.07$ \\
OpenBookQA & likelihood acc & 46.60 & 43.60 & 44.80 & 44.20 & $-2.40$ \\
PIQA & likelihood acc & 84.44 & 83.03 & 83.46 & 82.75 & $-1.69$ \\
RACE & likelihood acc & 42.39 & 41.72 & 42.39 & 41.53 & $-0.86$ \\
WinoGrande & likelihood acc & 79.48 & 78.06 & 78.37 & 78.22 & $-1.26$ \\
HumanEval pass@1 & pass@1 & 31.71 & 29.88 & 34.15 & 33.54 & $+1.83$ \\
\addlinespace
\multicolumn{7}{@{}l}{\textit{Diagnostic}} \\
BharatBBQ acc\_norm & acc\_norm & 35.73 & 35.04 & 39.22\ddag & $\circ$ & -- \\
\bottomrule
\end{tabular}
\end{table}

%% file: tables/table_helmet.tex
\begin{table}[H]
\centering\small
\caption{HELMET evaluation of \evonbase{}, the checkpoint that concludes long-context continual pretraining.}
\label{tab:helmet}
\begin{tabular}{lrrrrr}
\toprule
Category & 8K & 16K & 32K & 64K & 128K \\
\midrule
Recall & 95.9 & 81.4 & 57.0 & 37.3 & 14.0 \\
RAG & 62.7 & 60.2 & 56.6 & 49.4 & 37.2 \\
ICL & 59.2 & 71.7 & 78.6 & 81.8 & 81.7 \\
LongQA & 28.8 & 38.5 & 36.5 & 40.6 & 26.8 \\
\addlinespace
Indic-RAG & 57.1 & 59.1 & 54.2 & 46.8 & -- \\
Indic-Reader & 17.9 & 17.0 & 16.0 & 15.2 & -- \\
Indic-Abstention & 99.7 & 99.6 & 99.0 & 95.6 & -- \\
\bottomrule
\end{tabular}
\end{table}

%% file: tables/table18_matched.tex
\begingroup
\small
Both checkpoints use one matched 35-benchmark campaign. Four extractive-QA rows were rescored with
strict exact match after a harness update; F1 was unaffected and is used in the narrative. Deltas are
RL minus SFT\@. Accuracy metrics are percentages, while translation metrics use native sacreBLEU
units. BharatMath is omitted pending a harness revision; $\downarrow$ means lower is better.

\begin{longtable}{@{}L{3.1cm}L{2.55cm}R{2.15cm}R{2.15cm}R{2.15cm}@{}}
\caption{Matched evaluation of \evonit{} and the released \evonrl{} checkpoint.}
\label{tab:releasedrlresults}\\
\toprule
\textbf{Benchmark} & \textbf{Metric} & \textbf{Final SFT ckpt} & \textbf{RL ckpt step 40} & \textbf{$\Delta$ (RL $-$ SFT)} \\
\midrule
\endfirsthead
\multicolumn{5}{c}{\tablename\ \thetable{}: continued} \\
\toprule
\textbf{Benchmark} & \textbf{Metric} & \textbf{Final SFT ckpt} & \textbf{RL ckpt step 40} & \textbf{$\Delta$ (RL $-$ SFT)} \\
\midrule
\endhead
\midrule
\multicolumn{5}{r}{\footnotesize Continued on next page} \\
\endfoot
\bottomrule
\endlastfoot
\addlinespace[2pt]
\multicolumn{5}{@{}l}{\textit{Math \& Reasoning}} \\
MATH & exact match & 48.69 & 84.76 & $+36.07$ \\
MATH-500 & exact match & 48.80 & 84.00 & $+35.20$ \\
GSM8K & exact match & 85.75 & 94.31 & $+8.56$ \\
MGSM-English & exact match & 91.60 & 98.40 & $+6.80$ \\
MGSM-Indic & exact match & 79.60 & 84.60 & $+5.00$ \\
AIME 2025 & pass@1 & 72.92 & 74.38 & $+1.46$ \\
AIME 2026 & pass@1 & 72.08 & 74.58 & $+2.50$ \\
Beyond AIME & pass@1 & 48.50 & 53.56 & $+5.06$ \\
HMMT February 2025 & pass@1 & 59.58 & 60.83 & $+1.25$ \\
HMMT November 2025 & pass@1 & 58.54 & 63.33 & $+4.79$ \\
GPQA Diamond & pass@1 & 38.19 & 65.03 & $+26.84$ \\
\addlinespace[2pt]
\multicolumn{5}{@{}l}{\textit{Code}} \\
LiveCodeBench & pass@1 & 51.35 & 61.27 & $+9.92$ \\
MBPP+ & pass@1 & 75.24 & 78.78 & $+3.54$ \\
HumanEval+ & pass@1 & 88.66 & 88.54 & $-0.12$ \\
\addlinespace[2pt]
\multicolumn{5}{@{}l}{\textit{Knowledge and MCQ}} \\
MMLU (exact match)\textsuperscript{\ensuremath{\dagger}} & exact match & 77.21 & 86.68 & $+9.47$ \\
MMLU-Pro & accuracy & 66.67 & 77.26 & $+10.59$ \\
IndicMMLU-Pro & accuracy & 47.07 & 68.88 & $+21.81$ \\
MILU & accuracy & 73.16 & 78.33 & $+5.17$ \\
ARC-Challenge & exact match & 93.26 & 95.90 & $+2.64$ \\
AGIEval English & exact match & 60.02 & 73.47 & $+13.45$ \\
BhashaBench-Multi & accuracy & 42.06 & 49.22 & $+7.16$ \\
\addlinespace[2pt]
\multicolumn{5}{@{}l}{\textit{Instruction Following}} \\
IFEval & prompt strict & 66.36 & 82.81 & $+16.45$ \\
IFEval & prompt loose & 67.65 & 83.73 & $+16.08$ \\
IFEval & instruction strict & 73.26 & 87.17 & $+13.91$ \\
IFEval & instruction loose & 74.22 & 87.77 & $+13.55$ \\
IFEval-Hindi & prompt strict & 52.71 & 50.59 & $-2.12$ \\
IFEval-Hindi & prompt loose & 53.42 & 51.30 & $-2.12$ \\
IFEval-Hindi & instruction strict & 60.51 & 59.64 & $-0.87$ \\
IFEval-Hindi & instruction loose & 61.62 & 60.27 & $-1.35$ \\
IndicIFEval-translated & prompt strict & 37.51 & 49.87 & $+12.36$ \\
IndicIFEval-translated & prompt loose & 38.88 & 50.87 & $+11.99$ \\
IndicIFEval-translated & instruction strict & 47.68 & 61.65 & $+13.97$ \\
IndicIFEval-translated & instruction loose & 49.04 & 62.43 & $+13.39$ \\
IndicIFEval-ground & prompt/\allowbreak instruction strict & 40.52 & 33.05 & $-7.47$ \\
IndicIFEval-ground & prompt/\allowbreak instruction loose & 43.59 & 33.43 & $-10.16$ \\
IFBench & prompt strict & 21.67 & 44.33 & $+22.66$ \\
IndicParam & accuracy & 32.94 & 39.28 & $+6.34$ \\
\addlinespace[2pt]
\multicolumn{5}{@{}l}{\textit{QA and Translation}} \\
XQuAD-English (EM) & exact match & 46.97 & 63.70 & $+16.73$ \\
XQuAD-English (F1) & F1 & 67.15 & 78.86 & $+11.71$ \\
XQuAD-Indic (EM) & exact match & 18.84 & 12.86 & $-5.98$ \\
XQuAD-Indic (F1) & F1 & 36.51 & 17.19 & $-19.32$ \\
XorQA-Indic (EM) & exact match & 6.67 & 35.24 & $+28.57$ \\
XorQA-Indic (F1) & F1 & 16.16 & 44.09 & $+27.93$ \\
IndicQA (EM) & exact match & 12.77 & 12.22 & $-0.55$ \\
IndicQA (F1) & F1 & 26.36 & 17.72 & $-8.64$ \\
FLORES-IN & chrF++ & 14.55 & 36.33 & $+21.78$ \\
FLORES-IN & BLEU & 3.64 & 14.94 & $+11.30$ \\
CrossSum-IN (chrF) & chrF & 7.00 & 12.81 & $+5.81$ \\
\addlinespace[2pt]
\multicolumn{5}{@{}l}{\textit{Bias diagnostic}} \\
BBQ & accuracy (disamb.) & 94.13 & 97.70 & $+3.57$ \\
BBQ & accuracy (amb.) & 92.29 & 88.69 & $-3.60$ \\
BBQ & bias score (disamb.) $\downarrow$ & 0.99 & 0.60 & $-0.39$ \\
BBQ & bias score (amb.) $\downarrow$ & 4.33 & 7.39 & $+3.06$ \\
\addlinespace
\multicolumn{5}{@{}L{15.8cm}@{}}{\footnotesize \textsuperscript{\ensuremath{\dagger}}MMLU here is generative exact match and is not comparable to the loglikelihood accuracy reported in Table~\ref{tab:cptevalendpoints}.} \\
\multicolumn{5}{@{}L{15.8cm}@{}}{\footnotesize The MATH gain combines emission and conditional accuracy. In raw responses, boxed-answer emission rises from 2{,}911 to 4{,}612 of 5{,}000 problems, while boxed conditional accuracy rises from 86.4\% to 92.2\%. Emission explains 81--83\% of the gain, depending on decomposition order. Section~\ref{sec:capconv} instead partitions after reasoning is stripped, matching the live grader. These are two definitions applied to the same campaign and should not be mixed.} \\
\end{longtable}
\endgroup

%% file: appendix_qa_taxonomy.tex
Section~\ref{sec:rl} attributes the Indic extractive-QA regression to an unrewarded output
convention rather than to a loss of reading ability. This appendix records the item-level evidence.

Each response is assigned to one bin, in order: \textbf{B0}, no answer; \textbf{B1b}, English where
the gold span is Indic; \textbf{B2}, a gold-equivalent answer recovered only by prose-tolerant
regrading; \textbf{B3}, a target-script answer that is neither gold-equivalent nor a verbatim passage
span; \textbf{B4}, a wrong passage span; and \textbf{OK} otherwise.

Table~\ref{tab:qataxonomy} reports the four largest bins, so rows do not sum to 100\%. No response is
empty, and the omitted bins contribute at most three percent per row. Shares are equal-weight macros
over eight languages and should not be pooled across languages. This slice differs from the
eleven-leaf scorecard macro in Table~\ref{tab:releasedrlresults}.

\begin{table}[htbp]
\centering
\caption{XQuAD-Indic failure taxonomy, equal-weight macro over eight languages.}
\label{tab:qataxonomy}
\begin{tabular}{@{}lrrrrr@{}}
\toprule
\textbf{Checkpoint} & \textbf{Macro F1} & \textbf{B1b} & \textbf{B2} & \textbf{B3} & \textbf{OK} \\
\midrule
\evonit{} (SFT initialization) & 40.38 & 8.7\% & 5.8\% & 62.6\% & 21.6\% \\
Released \evonrl{} & 16.59 & \textbf{91.7\%} & 0.1\% & \textbf{2.5\%} & 2.8\% \\
Indic-aware sibling (Sec.~\ref{app:phase2rl}) & 53.43 & 15.3\% & 6.0\% & 45.4\% & 31.9\% \\
\bottomrule
\end{tabular}
\end{table}

IndicIFEval also varies across evaluation states. The FP8-paired BF16 row below is the canonical
release-protocol reading used in the main text; the earlier rows are retained as sensitivity checks.

\begin{table}[htbp]
\centering\small
\caption{Recorded IndicIFEval sensitivity across evaluation states.}
\label{tab:rlmeasurement}
\begin{tabular}{@{}lrrr@{}}
\toprule
\textbf{Measurement} & \textbf{Ground strict} & \textbf{Translated strict} & \textbf{Ground in-script} \\
\midrule
Initial matched evaluation & 28.66 & 47.45 & 5.4\% \\
Dedicated IndicIFEval evaluation\textsuperscript{\ensuremath{\ast}} & 63.88 & -- & 54.2\% \\
FP8-paired BF16 baseline & 33.05 & 49.87 & -- \\
\bottomrule
\end{tabular}

{\footnotesize \textsuperscript{\ensuremath{\ast}}The loose-criterion score is 65.62.}
\end{table}

IndicQA reproduces the language shift: B1b rises from 9.5\% to 73.0\%. A prose-tolerant regrade moves
macro F1 by at most 0.24 points, so extraction alone does not explain the loss.

The taxonomy has two limits. B3 is neither gold-equivalent nor verbatim, but it is not independently
classified as ungrounded because incorrect Indic answers rarely reproduce a passage span exactly. Likewise,
OK means that the response contains the gold span, not necessarily that it is only the gold span.
These distinctions matter most for the sibling branch, whose answers are usually in the target script.

Conditioning on target-script output separates emission rate from answer quality. Table~\ref{tab:qaconditional}
shows that the script rate collapses while conditional EM and F1 rise on both benchmarks. This is
consistent with preserved reading ability and a changed output convention, although it is not a
controlled estimate.

\begin{table}[htbp]
\centering\small
\caption{Target-script rate and conditional accuracy before and after released RL.}
\label{tab:qaconditional}
\begin{tabular}{@{}lrrrrrr@{}}
\toprule
& \multicolumn{2}{c}{Target-script rate} & \multicolumn{2}{c}{Conditional EM} & \multicolumn{2}{c}{Conditional F1} \\
\cmidrule(lr){2-3}\cmidrule(lr){4-5}\cmidrule(lr){6-7}
\textbf{Benchmark} & \textbf{SFT} & \textbf{RL} & \textbf{SFT} & \textbf{RL} & \textbf{SFT} & \textbf{RL} \\
\midrule
XQuAD-Indic & 89.9 & 5.4 & 23.22 & 51.50 & 37.69 & 61.19 \\
IndicQA & 86.9 & 24.3 & 20.24 & 45.23 & 27.00 & 50.54 \\
\bottomrule
\end{tabular}
\end{table}

Target-script answers form a self-selected subset, so the conditional rise may reflect selection.
We use it as evidence against broad reading collapse, not as a capability estimate.

The sibling study's Indic-aware checkpoint restores F1 above the initialization while its own B1b
share, at 15.3\%, remains above the initialization's 8.7\%. Rewarding target-script output recovered
the task score without fully restoring the convention.

%% file: appendix_phase2.tex
\subsection{Branch definition and reward}

\evonrlindic{} is a sibling research branch, not the public Evon v3.3 checkpoint and not a continuation of the released RL run. It initializes independently from the same \evonit{} iteration-2,400 checkpoint and adds a fifth environment, \texttt{indic\_if}, designed for Indic instruction following. Its reward is

\begin{equation}
  R = R_{\mathrm{task}}
      + 0.3R_{\mathrm{lang}}
      - 0.0R_{\mathrm{len}}
      - 0.3R_{\mathrm{rep}},
\end{equation}

Malformed or abandoned answers score $-0.25$. The branch also floors the advantage-normalization
denominator at 0.05, preventing tiny reward differences from becoming large updates.

$R_{\mathrm{task}}$ is the fraction of constraints satisfied. $R_{\mathrm{lang}}$ measures the share of answer words written in the row's target script. $R_{\mathrm{rep}}$ detects repeated $n$-gram bursts and character runs across the full generation. The length term is disabled in the experimental run.

The scopes are deliberate. Language compliance is computed only over the answer after the final reasoning block, because English internal reasoning with an Indic user-visible answer is acceptable. Repetition is computed over the entire generation, because looping reasoning is undesirable in any language.

The reward is target-aware rather than ``Indic is always good''. English examples target Latin script; Hindi targets Devanagari; Bengali targets Bengali script; and so on. A language-identification gate helps distinguish languages that share a script. Unknown language labels cause the language reward to be skipped rather than guessed.

\subsubsection{Why the experimental mix retains an English guardrail pool}

The final mix diverts 1,509 English rows, or 9.9\% of the input, from the existing task environments
into a shared guardrail pool. These rows are not an additional English corpus.

This guardrail is essential. An Indic-only reward taught the model to answer English prompts in Indic
scripts. The English rows instead make ``use the requested language'' learnable.

Directive ablations show that the literal language name is the active cue: explicit Hindi directives
rise from 38.4\% to 90.5\% in-script, while implicit forms remain near zero. The training prompts
therefore vary wording and placement but retain explicit language supervision.

\subsubsection{Experimental mixture and configuration}

The experimental run reinitializes from \evonit{} and trains for 40 steps over 15,300 prompts.
Table~\ref{tab:rlinputmix} and the configuration table give the mixture and rollout details.

The input proportions are cap-driven rather than specified as a target weight vector. Each rated environment is first filtered to a difficulty band and then capped; the final proportions are the row counts that survive those operations. Mathematics dominates because it has the largest eligible pool. \texttt{indic\_if} and MCQA retain unrated rows and are therefore pool-size limited rather than difficulty selected. Logic environments are excluded from this experimental mix.

\begin{figure}[H]
  \centering
  \includegraphics[width=0.78\textwidth]{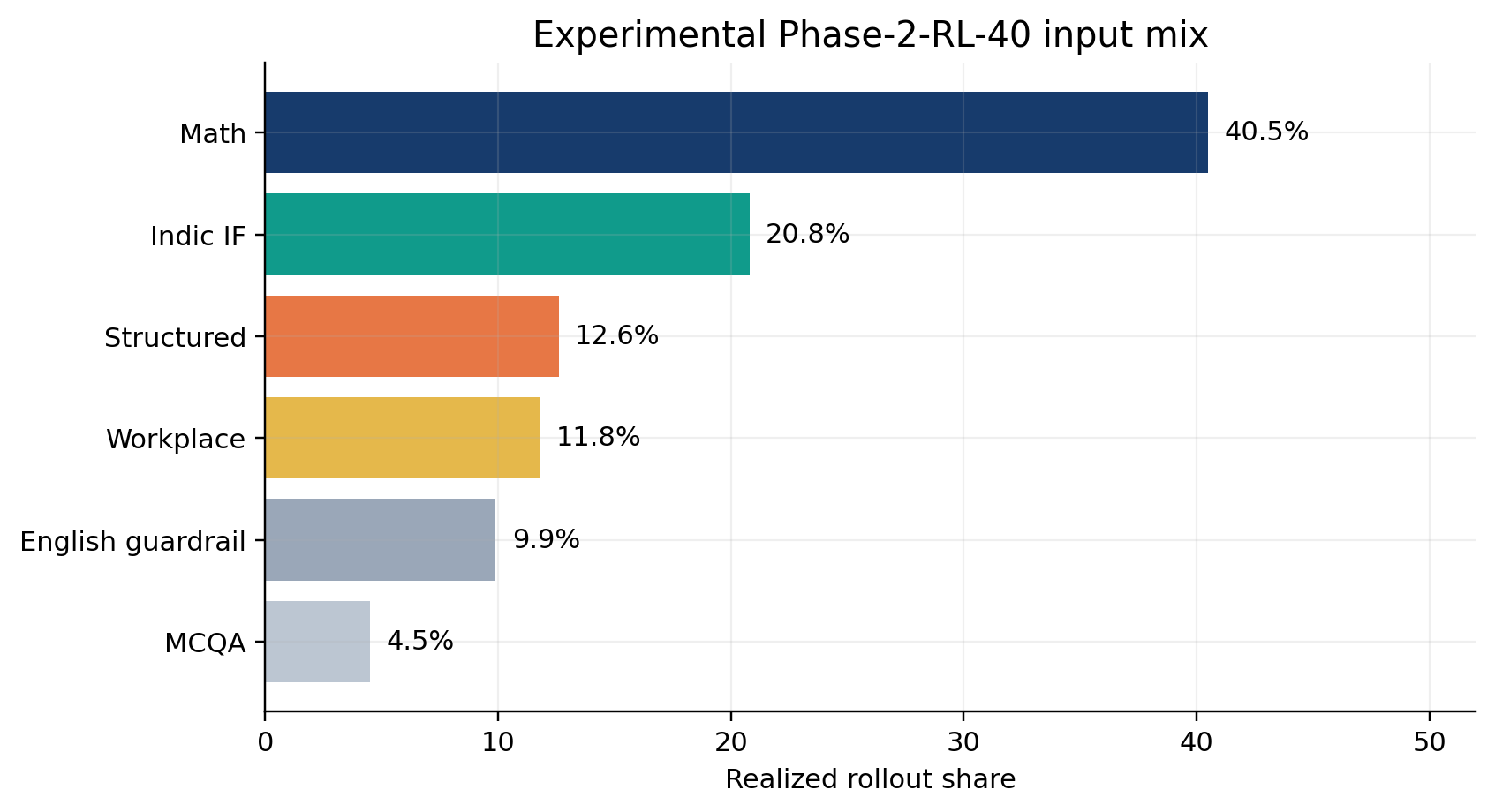}
  \caption{Experimental \evonrlindic{} training-row composition after difficulty filtering, environment caps, and English-row diversion. This mixture is not used by the released model.}
  \label{fig:rlmix}
\end{figure}

\begin{table}[H]
\centering
\caption{Cap-driven composition of the experimental \evonrlindic{} input mix. Component shares are rounded to one decimal place and therefore sum to 100.1\%; row counts sum exactly to 15,300.}
\label{tab:rlinputmix}
\begin{tabular}{@{}lrr@{}}
\toprule
\textbf{Pool} & \textbf{Rows} & \textbf{Final share} \\
\midrule
Mathematics & 6,192 & 40.5\% \\
Indic instruction following & 3,179 & 20.8\% \\
Structured outputs & 1,932 & 12.6\% \\
Workplace assistant & 1,804 & 11.8\% \\
English guardrail & 1,509 & 9.9\% \\
MCQA & 684 & 4.5\% \\
\midrule
Total & 15,300 & 100\% \\
\bottomrule
\end{tabular}
\end{table}

These row shares are not effective gradient shares: dynamic sampling removes all-pass and all-fail
groups, so environment contributions must be measured from the surviving rollouts.

\begin{table}[H]
\centering
\caption{Experimental \evonrlindic{} configuration.}
\begin{tabularx}{0.9\textwidth}{@{}L{4.1cm}Y@{}}
\toprule
\textbf{Setting} & \textbf{Value} \\
\midrule
Algorithm & GRPO with leave-one-out baseline and whole-group standard-deviation normalization \\
Inherited estimator flags & \texttt{use\_leave\_one\_out\_baseline=true}; \texttt{normalize\_rewards=true} \\
Generations per prompt & 8 \\
Dynamic sampling & enabled \\
Sampling temperature & 1.0 \\
Training steps & 40 \\
Training rows / validation rows & 15,300 / 256 \\
Recorded rollouts & 251,906 \\
Maximum generated tokens & 16,384 \\
Maximum RL sequence length & 49,152 \\
Model context capability & remains 131,072 \\
Advantage standard-deviation floor & 0.05 \\
Checkpoint cadence & every 5 steps \\
\bottomrule
\end{tabularx}
\end{table}

The 49,152-token RL cap is an optimization limit for rollout collection; it does not alter the model's trained or exported 131,072-token context capability.

Aggregate validation reward rises from 0.320 at initialization to 0.649 at step 40. Indic
instruction following and structured output contribute most of the gain; mathematics peaks at step
30 before easing. These branch-specific rewards are not directly comparable with the released run.

\begin{figure}[H]
  \centering
  \includegraphics[width=0.98\textwidth]{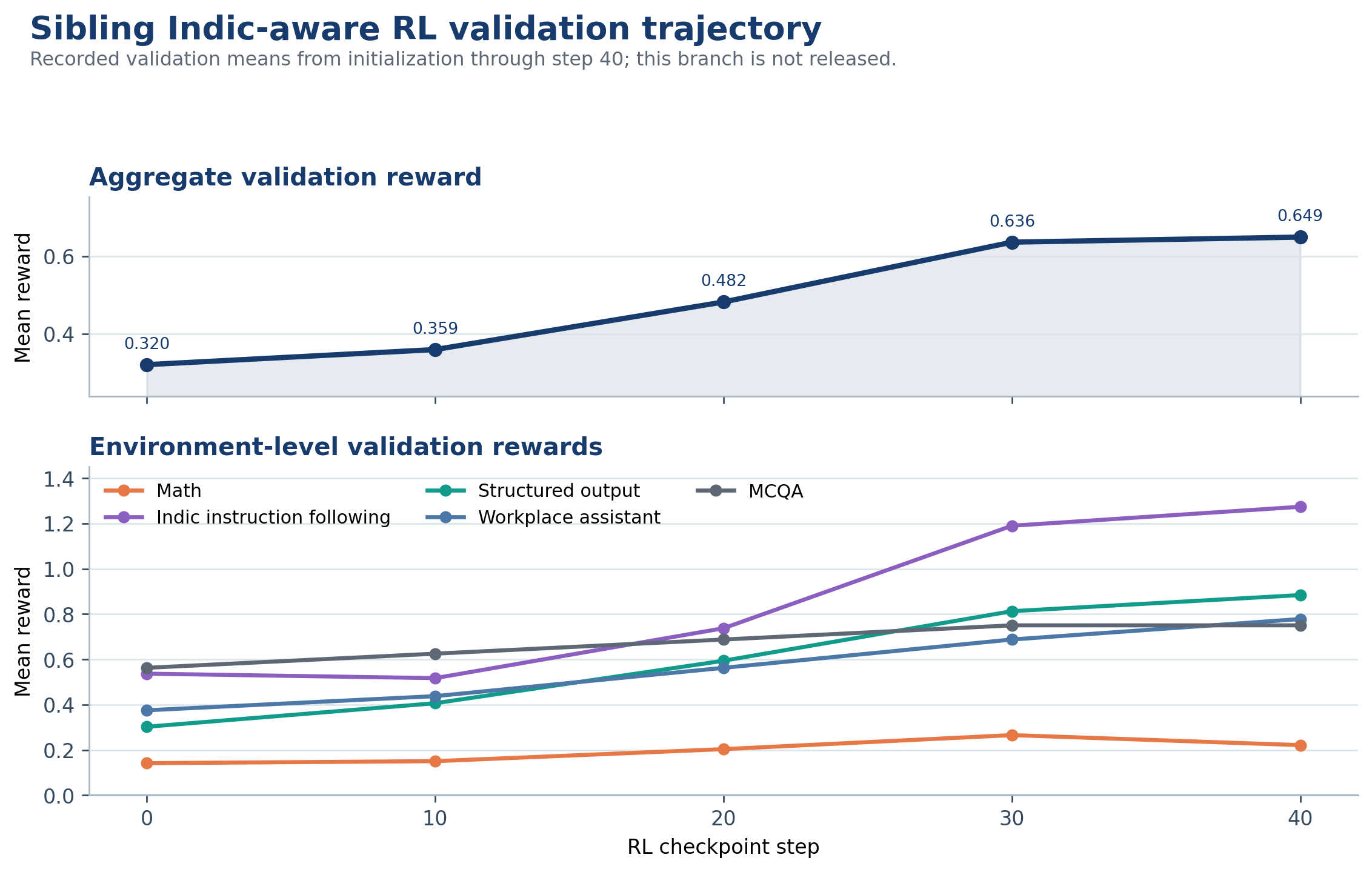}
  \caption{Validation rewards for the Indic-aware branch. These optimization diagnostics are not
  benchmark accuracies.}
  \label{fig:phase2rlvalidation}
\end{figure}

\subsubsection{Experimental matched results}

\evonrlindic{} is the first experimental artifact in this programme to improve reasoning and Indic instruction following simultaneously. It is reported as evidence about reward design, not as a result for the released Evon v3.3 checkpoint.

Both columns come from the campaign in Table~\ref{tab:releasedrlresults}, allowing direct comparison.
FLORES-IN chrF++ uses native sacreBLEU units, not percentages.

\begin{figure}[H]
  \centering
  \includegraphics[width=0.93\textwidth]{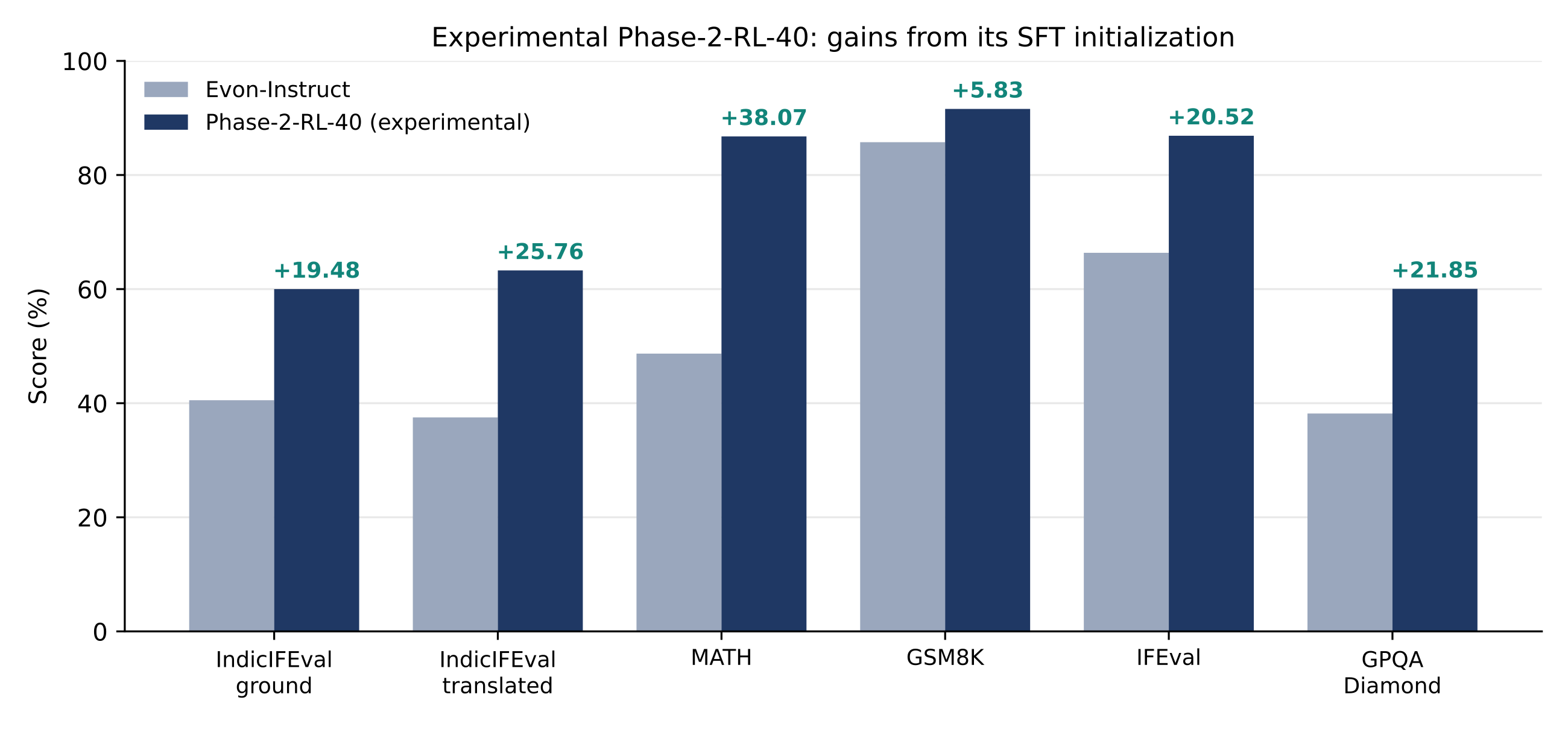}
  \caption{Selected \evonit{} to experimental \evonrlindic{} results. Deltas are percentage points; the checkpoint is not released.}
  \label{fig:rlresults}
\end{figure}

\begin{table}[H]
\centering
\small
\caption{Matched evaluation of the unreleased Indic-aware checkpoint against its SFT initialization.}
\label{tab:experimentalrlresults}
\begin{tabular}{@{}lrrr@{}}
\toprule
\textbf{Benchmark} & \textbf{\evonit{}} & \textbf{\evonrlindic{}} & \textbf{$\Delta$} \\
\midrule
IndicIFEval-ground, prompt strict & 40.52 & 60.00 & $+19.48$ \\
IndicIFEval-translated, prompt strict & 37.51 & 63.27 & $+25.76$ \\
XQuAD-Indic, F1 & 36.51 & 50.44 & $+13.93$ \\
MATH & 48.69 & 86.76 & $+38.07$ \\
GSM8K & 85.75 & 91.58 & $+5.83$ \\
IFEval, prompt strict & 66.36 & 86.88 & $+20.52$ \\
GPQA Diamond & 38.19 & 60.04 & $+21.85$ \\
FLORES-IN, chrF++ & 14.55 & 39.37 & $+24.82$ \\
\addlinespace
Native-Indic in-script rate\textsuperscript{\ensuremath{\dagger}} & 99.9 & 98.7 & $-1.2$ \\
\bottomrule
\end{tabular}

\footnotesize \textsuperscript{\ensuremath{\dagger}}In-script rate is a separate measurement and is not part of the benchmark campaign above.
\end{table}

Relative to the released checkpoint, the largest gains are on IndicIFEval-ground ($+26.95$ prompt-strict) and XQuAD-Indic ($+33.25$ F1). The table above measures the same benchmark against \evonit{} rather than against the released checkpoint, which is why its figure differs. Figure~\ref{fig:phase2tradeoffs} shows the remaining trade-offs.

Native-script retention is not uniform across prompt types. On the translated IndicIFEval split, in-script output falls from 95.1\% for \evonit{} to 84.6\% for the experimental checkpoint, even as task accuracy improves. This branch is therefore evidence for the reward-design mechanism, not a claim that every language-control problem is solved.

Step 40 is selected because it retains the step-30 mathematics gain while improving several other
tracked tasks. Mathematics is well defended because it has the largest share of the mixture.

\subsubsection{Direct comparison with the released checkpoint}

\begin{figure}[H]
  \centering
  \includegraphics[width=0.88\textwidth]{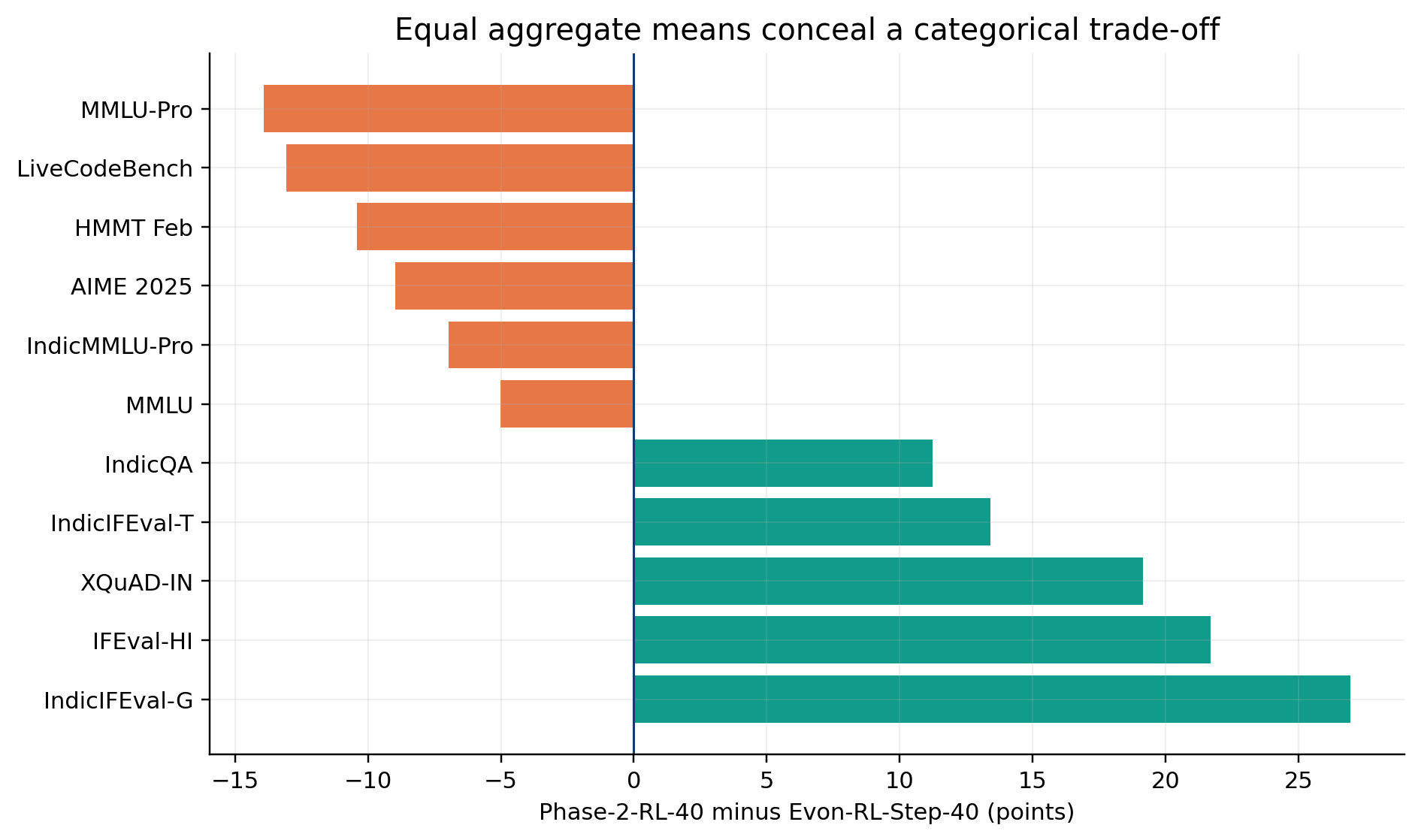}
  \caption{Selected matched differences between experimental \evonrlindic{} and released \evonrl{}. Positive values favor the experimental checkpoint.}
  \label{fig:phase2tradeoffs}
\end{figure}

The two branches are compared under one matched configuration, with strict exact match on both sides.
Their 33-metric means are effectively equal, but the equality hides opposite trade-offs. The
Indic-aware branch makes large gains on the instruction-following rows it targets, while the released
checkpoint has more numerous, smaller advantages elsewhere. Removing the five targeted
instruction-following rows gives the released checkpoint a 0.0257 lead. Figure~\ref{fig:phase2tradeoffs}
shows the largest differences.

\evonrlindic{} leads on instruction following, translation, and several Indic tasks. The released
checkpoint remains stronger on reasoning, code, and English knowledge.

The same trade-off appears in agentic evaluation: the branch improves the TauBench tasks but falls on
SWE-bench Verified (Table~\ref{tab:phase2agentic}).

\begin{table}[H]
\centering
\caption{Additional released-to-experimental agentic comparison. Deltas are percentage points.}
\label{tab:phase2agentic}
\begin{tabular}{@{}lrrr@{}}
\toprule
\textbf{Benchmark} & \textbf{\evonrl{}} & \textbf{\evonrlindic{}} & \textbf{$\Delta$} \\
\midrule
MultiChallenge & 0.4925 & 0.5000 & +0.75 \\
TauBench airline & 0.3733 & 0.4400 & +6.67 \\
TauBench retail & 0.1520 & 0.2807 & +12.87 \\
TauBench telecom & 0.0936 & 0.3626 & +26.90 \\
SWE-bench Verified & 0.2280 & 0.1220 & $-10.60$ \\
\bottomrule
\end{tabular}
\end{table}

\subsubsection{Experimental tool-use decomposition}

On BFCL v4, the overall 17-category score changes from 53.81 for SFT to 51.16 after RL\@. By itself,
that suggests a general tool-use regression, but the category breakdown reveals a narrower issue.

\begin{table}[H]
\centering
\caption{BFCL v4 decomposition.}
\begin{tabular}{@{}lrr@{}}
\toprule
\textbf{Category} & \textbf{\evonit{}} & \textbf{\evonrlindic{}} \\
\midrule
Headline, renormalized over 17 categories & 53.81 & 51.16 \\
Parallel / parallel-multiple & 69.5 / 80.0 & 0.5 / 0.5 \\
Live parallel / live parallel-multiple & 75.0 / 75.0 & 0.0 / 0.0 \\
Simple Python / multiple & 86.8 / 81.0 & 94.5 / 93.0 \\
Multi-turn base & 34.0 & 58.0 \\
\bottomrule
\end{tabular}
\end{table}

The experimental policy usually emits one tool call when the benchmark expects two in the same
assistant message. Its syntax parses correctly, and it issues the missing call on a second turn in 95
of 96 examined cases. The model retains the functional ability but has learned a one-action-per-turn
policy from an environment that always follows a call with an observation.

%% file: appendix_judge_opd.tex
This appendix records the reward-model, distillation, and abandoned-run configurations behind
Sections~\ref{sec:judge} and~\ref{sec:opd}. None belongs to the released lineage.

\subsection{Preference data}

Prompts combine native Indic instruction data with HelpSteer3-style English prompts translated into
eight Indian languages, in a 70:30 translated-to-native ratio. Two models generate one candidate
response each, and Gemini Flash labels the preference on a seven-point scale. HelpSteer3 contributes
the format and protocol, not its data. Of 98,744 pairs in Hindi, Telugu, Tamil, and Malayalam, 99.6\%
were labeled.

\subsection{The judge's reward}
\label{app:judge}

Initialized from Gemma~4 26B-A4B with its vision tower frozen, the judge emits two absolute
helpfulness scores and a forced relative ranking. GRPO rewards the correct side, calibration, and
agreement between the outputs; Table~\ref{tab:judgereward} gives the terms.

\begin{table}[htbp]
\centering\small
\caption{Reward terms used to train the generative judge.}
\label{tab:judgereward}
\begin{tabularx}{0.94\textwidth}{@{}L{3.4cm}Y@{}}
\toprule
\textbf{Term} & \textbf{Reward} \\
\midrule
Correct relative ranking & $+4$ \\
Calibration & $-1.2H_{1.5}(d_r)-0.6[H_1(d_{h_1})+H_1(d_{h_2})]$ \\
Absolute/relative inconsistency & $-0.5$ \\
Unparseable verdict & $-8$ \\
Parseable range & clipped to $[-6,4]$, then adjusted for batch-level rank balance \\
\bottomrule
\end{tabularx}
\end{table}

Here $H_\delta$ is Huber loss, $d_{h_k}=\hat h_k-h_k$, and $d_r=\hat r-r$.

The format penalty lies below the clip floor, so every parseable verdict beats every unparseable one.
Calibration also preserves a gradient when all samples choose the same side, which occurs in 78\% of
groups. Advantages are therefore centered but not divided by group standard deviation, avoiding the
amplification of small calibration differences. Training uses 262,834 examples, with each pair shown
in both orders; 72.4\% come from the Indic set.

Calibration, rather than accuracy, continues to improve. The rank-balance term reduces total variation
distance between predicted and target rank histograms from 0.327 to 0.083 without lowering accuracy.
Step 100 is exported; later checkpoints regress on both measures.

RM-Bench uses pairwise comparisons, so its scores are not directly comparable with scalar
reward-model scores. JudgeBench is a smaller capped-generation subset. The largest margin below is
on the hard slice, where style and substance point to different responses.

\begin{table}[htbp]
\centering\small
\caption{The 26B-A4B judge against a 235B-parameter generative reward model.}
\label{tab:genrmquality}
\setlength{\tabcolsep}{4pt}
\begin{tabular}{@{}lrrrrrrrrr@{}}\toprule
& \multicolumn{8}{c}{RM-Bench} & \\\cmidrule(lr){2-9}
Model & Chat & Math & Code & Safety & Easy & Normal & Hard & Overall & JudgeBench \\\midrule
Qwen3-Nemotron-235B-A22B-GenRM & 76.5 & \textbf{96.9} & 81.4 & 94.4 & 94.0 & 90.5 & 77.4 & 87.3 & 87.4 \\
Ours, 26B-A4B & 76.1 & 93.9 & \textbf{84.4} & \textbf{95.7} & 94.1 & \textbf{92.0} & \textbf{86.4} & \textbf{90.8} & \textbf{88.3} \\
\bottomrule
\end{tabular}
\end{table}

\subsection{Reinforcement learning against the judge}

Each prompt samples $G=16$ completions. Comparing neighbors around a ring requires 16 judge calls
instead of all 120 pairs. Every completion appears once in each position, cancelling position bias,
and the forced ranking prevents zero-variance groups. Reasoning is hidden from the judge. The policy
reward also includes script consistency, repetition control, and a stochastic reasoning-length term.

\begin{table}[htbp]
\centering\small
\caption{Selected gains from policy optimization against the judge, relative to \evonrlindic{}.}
\label{tab:judgepolicygains}
\begin{tabular}{@{}lr@{}}
\toprule
\textbf{Benchmark} & \textbf{Change} \\
\midrule
LiveCodeBench & $+13.2$ \\
MMLU-Pro & $+8.8$ \\
IndicMMLU-Pro & $+6.3$ \\
MMLU & $+5.8$ \\
GPQA Diamond & $+5.8$ \\
AGIEval & $+5.5$ \\
Grounded Indic instruction following & $+4.4$ \\
Mathematics aggregate & $+2.7$ \\
\bottomrule
\end{tabular}
\end{table}

Evaluation and distillation use step 20 to limit drift. Step 20 is not the highest-scoring judge
checkpoint, and the same regressions remain visible later, so this choice does not create them.

\subsection{Length control}

Three designs were tested. Rewarding the shortest trace drove reasoning to zero, while a two-sided
ramp reduced it by 60\% and cost more than 25 AIME points. The adopted one-sided penalty preserves
judge quality without attracting the policy to an artificial length target.

\subsection{The 70:30 mixture that failed}
\label{app:mixture}

The recipe was retrained with 30\% of prompts scored by binary mathematics and multiple-choice
verifiers. For a problem solved with probability $p$, a group of $G=16$ is degenerate with probability
$z(p)=p^{16}+{(1-p)}^{16}$. Dynamic sampling discards such all-pass and all-fail groups. The measured
degenerate rate was high enough to trigger a second generation batch on 23 of 29 steps, making each
step 2.27 times slower than the judge-only baseline.

All judge groups survived, while 35--49\% of verifier groups were discarded. The specified 70:30
split therefore became roughly 80:20 at the gradient, and the run was stopped after thirty steps.
About 40\% of the verifier pool contained problems the model almost always solved or almost always
missed. Restricting the pool to intermediate pass rates would reduce the discard probability by more
than an order of magnitude. The original monitor missed this because it inspected only the judge
environment, where groups cannot degenerate.

\subsection{Distillation configuration}

The teacher is \evonrlindic{}, and the student starts from judge-training step 20. At each student
token, supervision uses the teacher's top-64 next-token distribution under Equation~\ref{eq:opd},
with $w=0.5$. The 29,988 mathematics and multiple-choice prompts cover English and eight Indian
languages, with source-problem splits preventing translated leakage into validation. Training runs
for 100 steps at a constant learning rate of $10^{-5}$; only step 20 is evaluated.